\documentclass[twoside]{article}

\usepackage[accepted]{aistats2023}
\usepackage[round]{natbib}

\usepackage{xr}
\makeatletter
\newcommand*{\addFileDependency}[1]{
  \typeout{(#1)}
  \@addtofilelist{#1}
  \IfFileExists{#1}{}{\typeout{No file #1.}}
}
\makeatother

\usepackage[utf8]{inputenc} 
\usepackage[T1]{fontenc}    
\usepackage[hidelinks]{hyperref}       
\usepackage{url}            
\usepackage{booktabs}       
\usepackage{amsfonts}       
\usepackage{nicefrac}       
\usepackage{microtype}      
\usepackage{xcolor}         
\usepackage{lipsum}
\usepackage{enumitem}
\usepackage{overpic}
\usepackage{subcaption}
\usepackage{amsthm,amsmath}
\usepackage{amsfonts,mathtools,amssymb,float}
\usepackage{bm}
\usepackage{hyperref}
\numberwithin{equation}{section}
\theoremstyle{plain}
\newtheorem{theorem}{Theorem}

\newtheorem{lemma}{Lemma}[section]

\newtheorem*{definition*}{Definition}

\numberwithin{equation}{section}
\theoremstyle{plain}

\DeclareMathOperator{\E}{\mathbf{E} }

\makeatletter
\newcommand*{\rom}[1]{\expandafter\@slowromancap\romannumeral #1@}

\newcommand{\var}{\textup{var}}

\newcommand{\ttop}{^{\top}}

\renewcommand{\i}{\mathbf{i}}

\newcommand{\D}{\mathcal{D}}
\newcommand{\R}{\mathbb{R}}

\renewcommand{\P}{\mathbf{P}}

\newcommand{\RR}{\mathsf{R}}
\newcommand{\Q}{\mathsf{Q}}
\newcommand{\Z}{\mathsf{Z}}

\newcommand{\K}{\mathsf{K}}
\newcommand{\tK}{\widetilde{\mathsf{K}}}
\newcommand{\ts}{\textstyle}
\newcommand{\ve}{\varepsilon}

\newcommand{\tve}{\tilde{\ve}_{1-\alpha}}
\newcommand{\vealph}{\ve_{1-\alpha}}
\newcommand{\dealph}{\delta_{1-\alpha}}
\newcommand{\tdealph}{\tilde{\delta}_{1-\alpha}}

\newcommand{\MMD}{\textup{MMD}}
\newcommand{\MMK}{\textup{MMK}}

\makeatletter
\def\namedlabel#1#2{\begingroup
    #2%
    \def\@currentlabel{#2}%
    \phantomsection\label{#1}\endgroup
}
\makeatother

\begin{document}

%

%

\runningauthor{Yao, Erichson, Lopes}

\twocolumn[

\aistatstitle{Error Estimation for Random Fourier Features}

\aistatsauthor{Junwen Yao
\And N. Benjamin Erichson
\And  Miles E. Lopes
}

\aistatsaddress{ University of California, Davis \And   Lawrence Berkeley National Laboratory \And University of California, Davis } ]

\begin{abstract}
Random Fourier Features (RFF) is among the most popular and broadly applicable approaches for scaling up kernel methods. In essence, RFF allows the user to avoid costly computations on a large kernel matrix via a fast randomized approximation. However, a pervasive difficulty in applying RFF is that the user \emph{does not know the actual error} of the approximation, or how this error will propagate into downstream learning tasks. Up to now, the RFF literature has primarily dealt with these uncertainties using theoretical error bounds, but from a user's standpoint, such results are typically impractical---either because they are highly conservative or involve unknown quantities. To tackle these general issues in a data-driven way, this paper develops a bootstrap approach to \emph{numerically estimate} the errors of RFF approximations. Three key advantages of this approach are: (1) The error estimates are specific to the problem at hand, avoiding the pessimism of worst-case bounds. (2) The approach is flexible with respect to different uses of RFF, and can even estimate errors in downstream learning tasks. (3) The approach enables adaptive computation, so that the user can quickly inspect the error of a rough initial kernel approximation and then predict how much extra work is needed. Lastly, in exchange for all of these benefits, the error estimates can be obtained at a modest computational cost.

\end{abstract}

\section{INTRODUCTION} \label{sec:intro}

Although kernel methods are fundamental to many types of machine learning systems, they have an Achilles heel, insofar as they have limited scalability when they are applied to large datasets in a direct manner~\citep{Scholkopf2002,Shawe2004}. The basic source of this issue is that an $n\times n$ kernel matrix derived from $n$ data points typically incurs an $\mathcal{O}(n^2)$ storage cost, and an $\mathcal{O}(n^3)$ processing cost for common learning tasks. Due to such bottlenecks, techniques for accelerating kernel methods have been studied extensively, and over the years, the approach of \emph{Random Fourier Features} (RFF) has become well-established as one of the most popular and effective ways to scale up kernel methods in a plethora of applications~\citep{Rahimi2007,Le2013,dai2014double,zhao2015fastmmd,Avron2017,zhang2019low,Liu2020,giannakis2021learning,kiessling2021wind}.

The core idea of RFF is to avoid direct computations on a large kernel matrix by working more efficiently with an approximation built from ``randomly sampled features''. As a result of this approximation, RFF involves an inherent tradeoff between computational cost and accuracy. 
However, managing this tradeoff in practice is complicated by the fact that the user \emph{does not know the actual error} of the approximation, or how this error may jeopardize downstream results. In addition, this uncertainty about error can lead the user to sample far more features than are really necessary, which erodes the computational gains of RFF.

At a conceptual level, the RFF literature is able to offer insights on these issues with various theoretical error bounds, which are surveyed in~\cite{Liu2020}. However, there has been a longstanding gap between theory and practice, because these results generally do not provide actionable guidance at a numerical level. One reason for this difficulty is that theoretical error bounds tend to be formulated to hold uniformly over a class of possible inputs, which often causes the bounds to be highly pessimistic for typical problem instances. (Empirical illustrations of this conservativeness can be found, for example, in Figures 4 and 5 of~\cite{Sutherland2015}.) Such bounds frequently also involve unspecified constants or unknown parameters, preventing the user from extracting any numerical information at all.

\textbf{Contributions.} To overcome the challenges above, we develop a systematic way to \emph{numerically estimate} the errors of RFF approximations. Our contributions are briefly summarized below.
\begin{enumerate}[leftmargin=0.5cm]
    \item The error estimates are fully-data driven, and hence tailored to the inputs in a given problem. This bypasses  practical limitations of worst-case error bounds.
    %
    \item The error estimates enhance the computational efficiency of RFF, by guiding the user to choose a number of features that is just enough for a preferred error tolerance.
    %
    \item We give a precise theoretical guarantee on the validity of the error estimates in the context of kernel matrix approximation (Theorem~\ref{thm:main}), holding under mild assumptions.
    %
    \item We demonstrate the versatility of the error estimates in several RFF use cases, including kernel matrix approximation, kernel ridge regression, and kernel-based hypothesis testing.
\end{enumerate}

\subsection{Peliminaries on kernels and RFF}\label{sec:prelim}
\textbf{Kernels.} Throughout the paper, we consider learning tasks involving a shift-invariant kernel \smash{$k:\R^d\times \R^d\to\R$}. This means that $k$ is positive definite and satisfies the relation $k(x,x')=k(x-x',0)$ for all $x,x'\in\R^d$.
In addition, we will always assume that $k$ is continuous and is normalized so that $k(0,0)=1$. Kernels with these properties are the ones most often studied in the RFF literature, and well-known examples include the Gaussian, Laplacian, Cauchy, and B-spline kernels, among others surveyed in~\textsection 4.4-4.5 of \cite{Scholkopf2002}.

\textbf{Random features.} From a mathematical perspective, the linchpin of RFF is
a classical result from Fourier analysis known as Bochner's Theorem, which ensures that if $k$ is of the stated type, then there exists a probability distribution $\rho$ on $\R^d$ such that $k$ can be represented as
\begin{equation}\label{eqn:Bochner}
    k(x,x') \ = \ \int_{\R^d} e^{\sqrt{-1} \langle x-x',w\rangle} \mathrm{d}\rho(w),
\end{equation}
for all $x,x'\in\R^d$, where $\langle \cdot,\cdot\rangle$ is the Euclidean inner product~\citep{Rudin}. Crucially, this integral representation allows the kernel to be viewed as an expectation, because if $W$ is a random vector drawn from $\rho$, and if we define the ``random feature'' $\zeta(x)=e^{\sqrt{-1}\langle x,W\rangle}$ for any fixed $x\in\R^d$, then
    $ k(x,x') = \E[\zeta(x)\zeta(-x')]$.
However, due to the fact that $k$ is real-valued, whereas $\zeta(x)$ is complex-valued, it has become common in the RFF literature to use a real-valued modification of $\zeta(x)$. Such a modification can be defined as 
$$Z(x)=\sqrt{2}\cos(\langle x,W\rangle+U),$$
where $U$ is drawn from the uniform distribution on $[0,2\pi]$ independently of $W$, leading to 
\begin{equation}\label{eqn:Ereln}
    k(x,x') \ = \ \E[Z(x)Z(x')].
\end{equation}
\textbf{Randomized kernel approximations.} The importance of viewing $k$ as an expectation is that it enables us to approximate $k$ with a sample average involving $s$ random features, where $s\ll n$. Specifically, let $W_1,\dots,W_s\sim \rho$ and $U_1,\dots,U_s\sim\text{Uniform}[0,2\pi]$ be independent sets of i.i.d.~samples, and for any fixed $x\in\R^d$, denote the associated random features as $Z_i(x)=\sqrt{2}\cos(\langle x,W_i\rangle+U_i)$, with $i=1,\dots,s$. In this notation, the RFF approximation to $k(x,x')$ is defined by
\begin{equation}\label{eqn:sampavg}
    \tilde{k}(x,x') \ = \  \frac{1}{s}\sum_{i=1}^s Z_i(x)Z_i(x'),
\end{equation}
which is unbiased,~$\E[\tilde{k}(x,x')]=k(x,x')$, due to~\eqref{eqn:Ereln}.

Regarding kernel matrices, consider a fixed set of data points $x_1,\dots,x_n\in\R^d$, and let $\mathsf{K}\in\R^{n\times n}$ have entries given by $\K_{jj'}=k(x_j,x_{j'})$.  An approximation to $\mathsf{K}$ can be developed by first defining a random matrix $\Z\in\R^{n\times s}$ whose $i$th column is $\frac{1}{\sqrt s}(Z_i(x_1),\dots,Z_i(x_n))$. Then, in light of~\eqref{eqn:sampavg}, the RFF approximate kernel matrix is defined as
\begin{equation}\label{eqn:ZZ}
    \tK \ = \ \Z\Z^{ \scriptscriptstyle{\top}}.
\end{equation}
To briefly describe the computational advantages of RFF, it is worth re-emphasizing that the number of random features $s$ is generally chosen so that $s\ll n$. Combining this with the fact that $\tK$ is automatically factorized in terms of the $n\times s$ matrix $\Z$, it follows that for any $v\in\R^n$, a matrix-vector product can be computed as $\tK v=\Z[\Z^{\scriptscriptstyle{\top}}v]$ with a cost of only $\mathcal{O}(sn)$. Hence, this is much less than the corresponding $\mathcal{O}(n^2)$ cost to compute $\mathsf{K}v$.
More generally, such savings in linear-algebraic operations have enabled RFF to speed up a variety of learning tasks---such as reducing cost from $\mathcal{O}(n^3)$ to $\mathcal{O}(s^2n)$ in both kernel PCA and kernel ridge regression~\citep{LopezPaz,Avron2017}.

\subsection{Formalizing the error estimation problem}\label{sec:formal}
\textbf{Errors with respect to norms.}  When assessing the error of $\tK$ in relation to the exact matrix $\mathsf{K}$, a variety of norms may be of interest. Since our approach is flexible with respect to this choice, we let $\|\cdot\|_{\diamond}$ denote a generic norm on $\R^{n\times n}$.
For any such choice, it should be stressed that the actual error $\|\tK-\K\|_{\diamond}$ is both \emph{random} and \emph{unknown} to the user. Also, we regard the exact matrix $\K$ as fixed, and so the randomness in $\|\tK-\K\|_{\diamond}$ arises entirely from the random features used to construct $\tK$.

Our goal is to numerically estimate the tightest possible upper bound on $\|\tK-\K\|_{\diamond}$ that holds with a given probability, say $1-\alpha$, where $\alpha\in(0,1)$.
More formally, this ideal (unknown) bound is called the $(1-\alpha)$-quantile of $\|\tK-\K\|_{\diamond}$, and is defined as
\begin{equation*}
   \vealph \, = \, \inf\Big\{c\in[0,\infty) \ \Big| \ \P\Big(\|\tK-\K\|_{\diamond}\leq c\Big)\,\geq\, 1-\alpha\Big\}.
\end{equation*}
Below, Figure~\ref{fig:intro} illustrates how the quantile $\ve_{1-\alpha}$ can be interpreted in relation to the fluctuations of the random variable $\|\tK-\K\|_{\diamond}$, in the particular case when $\|\cdot\|_{\diamond}$ is the operator (spectral) norm and $1-\alpha=90\%$. 

To explain Figure~\ref{fig:intro}, consider a hypothetical scenario where it is possible to track the random variable $\|\tK-\K\|_{\diamond}$ as the number of features $s$ is increased over a grid ranging from 1 to 1600. The result is displayed with the red curve. Similarly, by repeating this experiment many times, a large collection of such random curves can be generated, and these are displayed in blue. (This scenario would not occur in practice, and is only for conceptual illustration.) In addition, the 90\% quantile of the curves at each value of $s$ is plotted in black, which represents $\ve_{1-\alpha}$.
\begin{figure}[H]
	\centering
		\DeclareGraphicsExtensions{.pdf}
		%
		\begin{overpic}[width=0.43\textwidth]{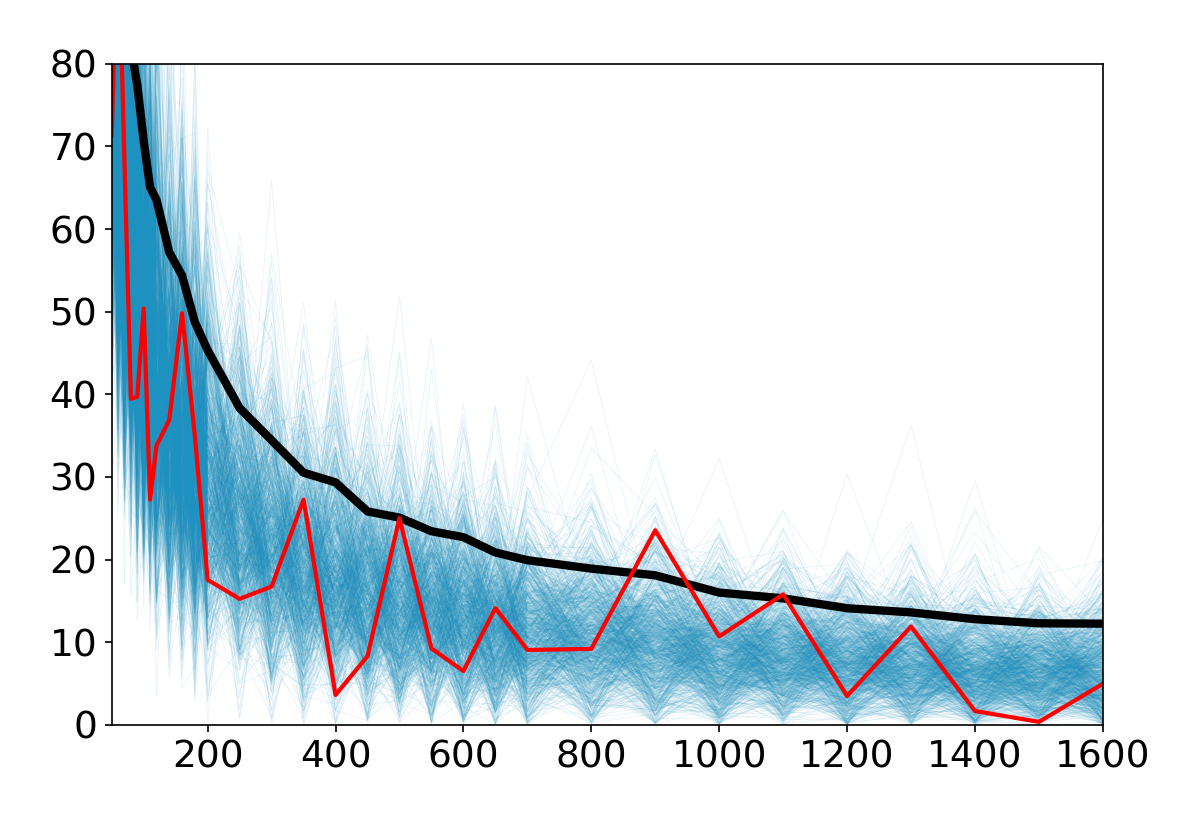}
			\put(35,-3){\color{black}{number of features $s$}}   
			\put(-4,27){\rotatebox{90}{$\|\tK-\K\|_{\diamond}$}}
			\put(33,57){\textbf{-----} 90\% quantile $\ve_{0.90}$ }
			\put(31.7,48){ {\color{red}\textbf{\textbf{-----}}} fluctuations of $\|\tK-\K\|_{\diamond}$}
		\end{overpic}	
		\vspace{+0.2cm}	
		\caption{
		(Interpretation of $\ve_{1-\alpha}$ when $1-\alpha=90\%$.) The plot was generated in the setting of the ``Swiss roll'' dataset described in Appendix~\ref{sec:add_matrix}. In particular, this involved a Gaussian kernel with bandwidth $\sigma=4\%$.} 
\label{fig:intro}
\end{figure}
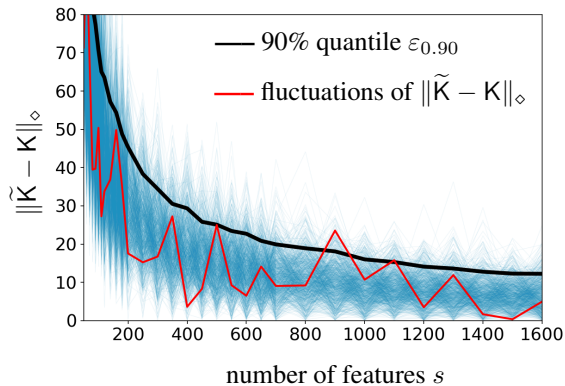
Hence, if the user had access to the black curve for $\vealph$, it would be possible to know if a given number of features $s_0$ is adequate, or to predict what larger value $s>s_0$ should be used to achieve a higher level of accuracy.
Despite the fact that none of the curves in Figure~\ref{fig:intro} are available to the user in practice, our work will show that, for a given value $s_0$, there is enough information in a \emph{single instance} of the $n\times s_0$ matrix $\mathsf{Z}$ (as in~\eqref{eqn:ZZ}) to closely estimate $\ve_{1-\alpha}$ for that value of $s_0$. Furthermore, it will also be shown in Section~\ref{sec:extrap} that a simple extrapolation rule can be used to rapidly estimate $\ve_{1-\alpha}$ for all larger values $s>s_0$.

\noindent\textbf{Estimation criteria.} When computing a numerical estimate, say $\tve$, for the true quantile $\vealph$, there are several important criteria to be met. 
First, the estimate should serve as a good proxy for $\vealph$, in the sense that the inequality
\begin{equation*}
    \|\tK-\K\|_{\diamond} \ \leq  \ \tve
\end{equation*}
holds with a probability that is close to $1-\alpha$ (cf. Theorem~\ref{thm:main}). Second, the estimate should not require any access to the full kernel matrix $\K$. Third, the algorithm used to compute $\tve$ should be efficient, so that the cost of error estimation does not outweigh the benefit of using RFF. In the remainder of this work, our proposed approach will be shown to meet all of these criteria.

\textbf{Errors with respect to functionals.} In addition to measuring error through norms, it is also of interest to measure error by comparing the kernel functions $\tilde k$ and $k$
with respect to various functionals $\psi$. For example, the values $\psi(\tilde k)$ and $\psi(k)$ could be measures of prediction error for learning algorithms based on $\tilde k$ and $k$ respectively, so that the difference $\psi(\tilde k)-\psi(k)$ represents how much predictive performance is sacrificed by the RFF approximation.
More generally, there are many other possibilities for comparing $\tilde k$ and $k$ in different contexts, such as letting $\psi$ represent eigenvalues in kernel PCA, or letting $\psi$ represent statistics for testing hypotheses. In these broader scenarios, the previous formulation of the error estimation problem can be extended by defining a counterpart for $\vealph$ according to
\begin{equation*}
\dealph  =  \inf\Big\{c\in[0,\infty) \, \Big| \, \P\Big(|\psi(\tilde k)-\psi(k)|\leq c\Big)\geq 1-\alpha\Big\}.
\end{equation*}
Likewise, our proposed approach can be applied to compute an estimate $\tdealph$ for $\dealph$ that meets the criteria mentioned previously. Also, the approach can be applied just as easily if the user prefers to define $\delta_{1-\alpha}$ by replacing $|\psi(\tilde k)-\psi(k)|$ with $\psi(\tilde k)-\psi(k)$.

\subsection{Related work}\label{sec:contrib} 
The existing literature on theoretical error bounds for RFF has grown substantially over the years, and so we only provide an illustrative sample. Results on kernel approximation can be found in~\cite{Rahimi2007, Sutherland2015, Sriperumbudur2015, Liu2020}. With regard to error analysis in other applications, such as  such as kernel-based regression, classification, and hypothesis testing, we refer to~\cite{yang2012nystrom, Sutherland2015, Avron2017, rudi2017generalization, sun2018but, li2019towards, Liu2020}.

To situate the current paper in the general context of numerical computation,
our work can be viewed as part of a topic known as \emph{a posteriori error estimation}---which refers to the process of estimating the error of a numerical solution after it has been computed. Although this topic has a mature literature in areas such as numerical PDE and finite element methods, an important distinction to make is that a posteriori error estimation has focused historically on deterministic algorithms \citep[e.g.][]{Babuska2:1978,Bank1985,Verfurth:1994,Ainsworth:2011}. Meanwhile, from a different perspective, our work can also be viewed as part of the extensive literature on \emph{bootstrap methods} for statistical inference~\citep[e.g.][]{Davison,HallBootstrap,Shao2012}.  Yet, from the standpoint of the statistics literature, relatively little attention has been given to bootstrap methods in the service of randomized algorithms for large-scale computation. Hence, our work sits at the border of two fields that have traditionally been quite distinct.

Nevertheless, the possibility of bridging this gap has not been overlooked completely, and there has been nascent interest in applying statistical ideas to estimate the errors of randomized algorithms, as noted in the recent survey~\citep{Tropp2020}. For instance, such interest has led to error estimation methods for randomized solutions to low-rank approximation~\citep{liberty2007,woolfe2008,Halko:2011}, matrix multiplication~\citep{lopes2019JMLR,lopes2023Bernoulli}, least-squares~\citep{Lopes2018,Ahfock2021}, singular value decomposition~\citep{Lopes2020svd}, and principal component analysis~\citep{Lunde2021}. However, to the best of our knowledge, statistical error estimation techniques for RFF have not previously been explored in a systematic way.
Therefore, given that RFF has been highly impactful, our work may offer new opportuntities to enhance many applications.

\textbf{Notation.} For any $\alpha\in(0,1)$, the empirical $(1-\alpha)$-quantile of a finite set of real numbers $A=\{a_1,\dots,a_N\}$ is defined as the smallest $a\in A$ such that $G_N(a)\geq 1-\alpha$, where $G_N(a)=\frac{1}{N}\sum_{j=1}^N 1\{a_j\leq a\}$, and $1\{\cdot\}$ denotes an indicator function. To denote this quantile, we write $\textup{quantile}(A; 1-\alpha)$. If $\mathbf{i}=(i_1,\dots,i_s)$ is a vector with entries taken from $\{1,\dots,s\}$, then $\Z(:,\mathbf{i})$ refers to the $n\times s$ matrix whose $l$th column is the $i_l$th column of $\Z\in\R^{n\times s}$. Similarly, for a vector $\mathsf{b}\in\R^s$, we define $\mathsf{b}(\mathbf{i})\in\R^s$ as the vector whose $l$th entry is the $i_l$th entry of $\mathsf{b}$. 
%
%
%
%
%
\section{METHOD}\label{sec:method}
Conceptually, the proposed bootstrap method for estimating $\vealph$ is based on generating a collection of ``pseudo error variables'' $\ve_1^{\star},\dots,\ve_N^{\star}$ that behave approximately like i.i.d.~samples of the unknown error variable $\|\tK-\K\|_{\diamond}$. Once the pseudo error variables have been generated, their empirical $(1-\alpha)$-quantile can then be used to define the estimate $\tve$. For example, if the user chooses $\alpha=0.1$ and $N=100$, then the estimate $\tve$ is defined as the 90th percentile among $\ve_1^{\star},\dots,\ve_{100}^{\star}$.

The subtlety of this approach consists in finding an effective way to generate $\ve_1^{\star},\dots,\ve_N^{\star}$. 
As a heuristic, we can imagine generating a random matrix $\tK^{\star}$  such that the difference $(\tK^{\star}-\tK)$ is  ``statistically similar'' to the difference $(\tK-\K)$, and then defining each $\ve_j^{\star}$ to be of the form $\|\tK^{\star}-\tK\|_{\diamond}$.

To explain this in more detail, it is important to notice that the matrix $\mathsf{Z}$ used to define $\tK$ has two special properties: (1) The columns of $\Z$ are i.i.d. (2) The columns of $\Z$ are generated so that $\E[\Z\Z^{ \scriptscriptstyle{\top}}]=\K$. 
Accordingly, we can try to generate an analogous matrix $\Z^{\star}$ having columns that are conditionally i.i.d.~given $\Z$, and satisfying the conditional expectation relation $\E[\Z^{\star}(\Z^{\star})^{ \!\scriptscriptstyle{\top}}|\Z]=\tK$. Then, we can let $(\Z^{\star}(\Z^{\star})^{ \!\scriptscriptstyle{\top}}-\Z\Z^{ \scriptscriptstyle{\top}})$ play the role of the matrix \smash{$(\tK^{\star}-\tK)$} mentioned earlier, and define pseudo error variables $\ve_j^{\star}$ having the form $\|\Z^{\star}(\Z^{\star})^{ \!\scriptscriptstyle{\top}}-\Z\Z^{ \scriptscriptstyle{\top}}\|_{\diamond}$. Furthermore, it turns out that these desired characteristics of $\Z^{\star}$ can be achieved by sampling its columns with replacement from the columns of $\Z$, which leads to the formulation of Algorithm 1 below.
 
In settings where RFF approximation error is measured in terms of $|\psi(\tilde k)-\psi(k)|$, the principles just discussed carry over analogously, and Algorithm 1 provides corresponding pseudo error variables $\delta_1^{\star},\dots,\delta_N^{\star}$.

\textbf{Algorithm 1. (Error estimation for RFF)}
~\\[-0.2cm]
\hrule
\textbf{Input:} A positive integer $N$, a number $\alpha\in(0,1)$, the matrix of random features $\Z\in\R^{n\times s}$, and the 
random functions $Z_1(\cdot),\dots,Z_s(\cdot)$.\\[0.2cm]
\textbf{For:} $j=1,\dots,N$ \textbf{ do in parallel }\\[-0.7cm]
\begin{itemize}[leftmargin=.5cm]
\item Draw a random vector $\textbf{i}=(i_1,\dots,i_s)$ by sampling $s$ numbers with replacement from $\{1,\dots,s\}$.
\item Define the $n\times s$ matrix $\Z^{\star}=\Z(:,\textbf{i})$.
\item Define the function 
$$\tilde k^{\star}(\cdot,\cdot')=\frac{1}{s}\Big(Z_{i_1}(\cdot)Z_{i_1}(\cdot')+\cdots+Z_{i_s}(\cdot)Z_{i_s}(\cdot')\Big).$$
\item Compute the pseudo error variables 
   $$ \ve_j^{\star}:=\|\Z^{\star}(\Z^{\star})^{ \!\scriptscriptstyle{\top}} - \Z\Z^{ \scriptscriptstyle{\top}}\|_{\diamond} \\
    \text{ \ \ \ and \ \ \  } \delta_j^{\star}:=|\psi(\tilde k^{\star}) - \psi(\tilde k)|.$$
\end{itemize}
\vspace{-0.2cm}
\noindent \textbf{Return:} The estimates  $\tve:=\text{quantile}(\ve_1^{\star},\dots,\ve_N^{\star}; 1-\alpha)$ \ and \  $\tdealph:=\text{quantile}(\delta_{1}^{\star}\dots,\delta_{N}^{\star}; 1-\alpha)$.
\vspace{0.2cm}
\hrule

\textbf{Remarks.} 
There are a few basic aspects of Algorithm 1 that are helpful to note for practical purposes. First, it is not always necessary to explicitly form the matrix $\Z^{\star}(\Z^{\star})^{ \!\scriptscriptstyle{\top}} - \Z\Z^{ \scriptscriptstyle{\top}}$, and this will be explained in greater detail in Section~\ref{sec:comp}. Second, the matrix $\Z^{\star}$ and function $\tilde{k}^{\star}$  can be overwritten after each iteration, which is why they are not marked with a subscript $j$.  Third, the number of bootstrap iterations $N$ generally does not need to be very large, and our experiments in Section~\ref{sec:expt} illustrate that $N\sim 50$ is often sufficient for a variety of tasks.

\section{COMPUTATIONAL EFFICIENCY}\label{sec:comp}
This section highlights the computational merits of Algorithm 1, and describes techniques for accelerating both error estimation and RFF. Since most of the ideas apply equally well to estimating both types of error, $\vealph$ and $\dealph$, we mainly address the former. 

\subsection{Selecting the number of features by extrapolation}\label{sec:extrap}
In the literature on bootstrap methods, a classical approach to speeding up computations is through the use of extrapolation techniques~\citep{BickelYahav}. In our current setting, this approach can be adapted as a two-step process: In the first step, we estimate the error of a ``preliminary'' RFF approximation that is computed from a small number of random features, say $s_0$. In the second step, we use this error estimate to predict how much $\vealph$ will decrease with a larger number of features, say $s_1 \gg s_0$. More concretely, if we make  the dependence of  $\vealph$ and $\tve$ on a generic value of $s$ explicit by writing $\vealph(s)$ and $\tve(s)$, then we seek to estimate  $\vealph(s_1)$ by extrapolating from $\tve(s_0)$.

There are two key benefits of this type of extrapolation. First, it can substantially speed up the error estimation process, because extrapolation only relies on  $\tve(s_0)$, which is computed by running Algorithm 1 on a small instance of $\Z$ with size $n\times s_0$. (If extrapolation is not used, then computing $\tve(s_1)$ requires a much larger instance of $\Z$ with size $n\times s_1$.) Second, extrapolation enhances RFF by enabling the user to choose a value of $s_1$ that is ``just large enough'' so that $\vealph(s_1)$ nearly matches a preferred error tolerance. In other words, this avoids the wasted computation that occurs when a user selects a highly excessive number of features due to uncertainty about accuracy.

From an algorithmic standpoint, an extrapolation rule can be developed as follows. Since it is possible to write $(\tK-\K)$ as a sample average of $s$ independent and zero-mean random matrices, the central limit theorem suggests heuristically that $\|\tK-\K\|_{\diamond}$ should decrease stochastically like $1/\sqrt{s}$ as a function of $s$. This also suggests that $\vealph(s_1)$ should be smaller than $\vealph(s_0)$ by a factor of $\sqrt{s_0/s_1}$, and so we define the extrapolated estimate of $\vealph(s_1)$ as
\begin{equation}\label{eqn:extrap}
    \tilde{\ve}_{1-\alpha}^{\textsc{ \,ext}}(s_1) \ = \ \sqrt{\ts\frac{s_0}{s_1}}\,\tve(s_0).
\end{equation}
Hence, if a user wants to select $s_1$ so that $\vealph(s_1)= \ve_{\text{tol}}$ for some tolerance $\ve_{\text{tol}}$, then $s_1$ can be chosen by setting $\tilde{\ve}_{1-\alpha}^{\textsc{ ext}}(s_1) = \ve_{\text{tol}}$, which leads to the choice $s_1= s_0 (\tve(s_0)/\ve_{\text{tol}})^2$. In Section~\ref{sec:expt}, our experiments illustrate the effectiveness of this rule when $s_1$ is larger than $s_0$ by \emph{two orders of magnitude}, demonstrating that extrapolation can yield major computational savings.

\subsection{Low communication and parallel processing\label{SEC:boot_parallel}}
In modern computing environments, communication costs are often of greater concern than processing costs~\citep[][\textsection 16.2]{Tropp2020}. For this reason, it is important to emphasize that when $\vealph$ is being estimated, Algorithm 1 \emph{does not require any access} to the $n\times n$ matrices $\K$ or $\tK$, but only to the much smaller matrix $\Z$. In fact, when extrapolation is used, Algorithm 1 only needs access to a ``preliminary'' instance of $\Z$ with $s_0$ columns, rather than a ``full'' instance of $\Z$ with $s_1\gg s_0$ columns that will be used for a high-quality RFF approximation. 

Another valuable feature of Algorithm 1 is its ``embarrassingly parallel'' structure. This means that the $N$ iterations of the for-loop can be trivially distributed across a collection of, say $m$, processors. Furthermore, our experiments will demonstrate that $N\sim 50$  is sufficient in many situations, and so if the user has access to just one or two dozen processors, it is often realistic to treat the number of bootstrap iterations per processor $N/m$ as being $\mathcal{O}(1)$.

\subsection{Computational cost in illustrative cases}
In this subsection, we quantify the computational cost of Algorithm 1 in some specific cases, with the benefits of extrapolation and parallel processing taken into account. The overall point of these examples is to show that the added cost of error estimation is manageable in comparison to the typical cost of RFF itself. As a benchmark for comparisons, it is worth noting that common learning tasks performed with RFF, such as kernel PCA and kernel ridge regression, have costs that are $\mathcal{O}(s_1^2n)$~\citep{LopezPaz,Avron2017}. (Here and below, we continue to use $s_0$ and $s_1$ respectively to denote number of features used for preliminary and high-quality RFF approximations.)

\textbf{Kernel matrix approximation.} First, we consider the cost of computing $\tilde{\ve}_{1-\alpha}^{\textsc{\,ext}}$ when error is measured through the operator norm $\|\tK-\K\|_{\textup{op}}$. Importantly, the matrix $\Z^{\star}(\Z^{\star})^{ \!\scriptscriptstyle{\top}} - \Z\Z^{ \scriptscriptstyle{\top}}$ in Algorithm 1 does not need to be explicitly formed when computing each $\ve_j^{\star}$. The reason is that the norm $\|\Z^{\star}(\Z^{\star})^{ \!\scriptscriptstyle{\top}} - \Z\Z^{ \scriptscriptstyle{\top}}\|_{\textup{op}}$ can be computed with variants of the power method, whose iterations are based on matrix-vector products $\Z^{\star}[(\Z^{\star})^{ \!\scriptscriptstyle{\top}}v] - \Z[\Z^{ \scriptscriptstyle{\top}}v]$ with $v\in\R^n$~\citep{Golub}. Also, as a basic guideline, the number of power iterations may be taken as $\mathcal{O}(\log(n))$~\citep[][\textsection 6.2.3]{Tropp2020}. In this case, each iteration of Algorithm 1 incurs a cost of $\mathcal{O}(s_0n\log(n))$.
Hence, if the iterations are computed in parallel, and the number of iterations per processor is $N/m=\mathcal{O}(1)$ (as described above), then the overall runtime is $\mathcal{O}(s_0n\log(n))$. Altogether, this compares well with the 
benchmark cost of $\mathcal{O}(s_1^2n)$ when $s_1\gg s_0$.

Alternatively, there is a second way to compute  the norm $\|\Z^{\star}(\Z^{\star})^{ \!\scriptscriptstyle{\top}} - \Z\Z^{ \scriptscriptstyle{\top}}\|_{\textup{op}}$ with lower communication costs. This approach originates from ideas in~\cite{epperly2022jackknife} and is based on computing a QR factorization $\Z = \Q \RR$, where $\Q\in\R^{n\times s_0}$ has orthonormal columns, and \smash{$\RR\in\R^{s_0\times s_0}$} is upper-triangular. By noting that the relation $\Z(:,\i)=\Q (\RR(:,\i))$ holds for every index vector $\mathbf{i}$ appearing in Algorithm 1, it follows from the unitary invariance of the operator norm that
%
%
\begin{equation} \label{EQN:unit_invariant_norm}
\small
\|(\mathsf{Z}^{\star})(\mathsf{Z}^{\star})\ttop-\mathsf{Z}\mathsf{Z}\ttop\|_{\textup{op}}=\|\RR(:,\i) \RR(:,\i)^\top) -\RR\RR\ttop \|_{\textup{op}}.
\end{equation}
\normalsize
So, if the bootstrap iterations are distributed across many processors, then the identity~\eqref{EQN:unit_invariant_norm} shows that it is only necessary to communicate copies of the $s_0\times s_0$ matrix $\RR$ to the processors, rather than copies of the $n\times s_0$ matrix $\Z$. Also, the cost of computing the right side of~\eqref{EQN:unit_invariant_norm} at each iteration is only $\mathcal{O}(s_0^2\log(s_0))$. However, these gains are offset by a one-time cost of $\mathcal{O}(s_0^2n)$ that must be paid to extract $\RR$. In the case when $N/m=\mathcal{O}(1)$, this approach leads to an overall runtime of $\mathcal{O}(s_0^2n)$. Although this nominally exceeds the $\mathcal{O}(s_0n\log(n))$ runtime of the previous approach when $\log(n)=\mathcal{O}(s_0)$, the reduced communication of this approach might still lead to better performance in practice. Also, this approach is favorable when parallel processing is limited, because its cost per iteration is lower.

\textbf{Kernel ridge regression.} As our second illustration of computational cost, we consider the use of Algorithm 1 in estimating the extra mean-squared test error that arises from RFF in kernel ridge regression. 

However, before diving into the details of cost, we first review the basic elements of kernel ridge regression and its associated RFF approximation. For a kernel $k$, let $f_k$ denote a kernel ridge regression function trained on $n$ data points in $\R^d$. This means that if the training points are denoted as $(x_1,y_1),\dots,(x_n,y_n)\in\R^{d}\times \R$ with $\mathsf{y}=(y_1,\dots,y_n)$, then 
\begin{equation}
 f_k(\cdot)=\sum_{i=1}^n \beta_i k(x_i,\cdot),
 \end{equation}
 where the vector $\beta\in\R^n$ solves $(\K+\lambda \mathsf{I}_n)\beta=\mathsf{y}$, and $\lambda>0$ is a tuning parameter. For the RFF approximation $\tilde k$, the associated regression function is defined as 
 %
\begin{equation}\label{eqn:fktildedef}
f_{\tilde k}(\cdot)=\sum_{i=1}^s \tilde\beta_i Z_i(\cdot),
\end{equation}
where the vector $\tilde \beta\in\R^s$ solves $(\Z^{\scriptstyle \top}\Z+\lambda \mathsf{I}_s)\tilde\beta=\Z^{\scriptstyle \top}\mathsf{y}$ and the functions $Z_1(\cdot),\dots,Z_s(\cdot)$ are as defined in Section~\ref{sec:prelim}.
Next, let $\psi(k)$ denote the mean-squared test error of $f_k$. More specifically, if there are $t$ test points denoted as $(x_1',y_1'),\dots,(x_t',y_t')\in\R^d\times \R$, then we have 
\begin{equation}\label{eqn:psikdef}
\psi(k)=\frac{1}{t}\sum_{i=1}^t (y_i'-f_k(x_i'))^2.
\end{equation}
Likewise, let $\psi(\tilde k)$ denote the corresponding quantity involving $f_{\tilde k}$.

Returning our attention to error estimation, let $\delta_{1-\alpha}$ denote the $(1-\alpha)$-quantile of $\psi(\tilde k)-\psi(k)$. Our goal here is to quantify the cost of computing an extrapolated estimate $\tilde{\delta}_{1-\alpha}^{\,\textsc{ext}}$ for $\delta_{1-\alpha}$. In this particular setting, there are a few ways to reduce the cost of Algorithm 1 by doing some one-time computations before starting the for-loop.  Namely, it is helpful to compute the scalar value $\psi(\tilde k)$, as well as the vector $\mathsf{b}=\Z\ttop \mathsf y$, and the QR factorization $\Z=\Q\RR$. (The motivation for the QR factorization is similar to that discussed earlier in connection with the work of~\cite{epperly2022jackknife}.) 

Inside the for-loop, each iteration computes a separate instance of the pseudo error variable $\psi(\tilde k^{\star})-\psi(\tilde k)$, with $\tilde k^{\star}$ being as defined in Algorithm 1. Since $\psi(\tilde k)$ has been pre-computed, it is only necessary to compute $\psi(\tilde k^{\star})$. This requires computing the solution $\tilde\beta^{\star}\in\R^{s_0}$ of the equation $((\Z^{\star})^{\scriptstyle \top}(\Z^{\star})+\lambda \mathsf{I}_{s_0})\tilde\beta^{\star}=\mathsf{b}^{\star}$, where $\Z^{\star}=\Z(:,\mathbf{i})$ and $\mathsf{b}^{\star}=\mathsf{b}(\mathbf{i})$. But instead of solving this equation directly, the initial QR factorization allows it to be solved more efficiently as \smash{$\big(\RR(:,\mathbf{i})^{\scriptstyle \top}\RR(:,\mathbf{i}) +\lambda \mathsf{I}_{s_0}\big)\tilde\beta^{\star}=\mathsf{b}^{\star}$.} 
Once the solution $\tilde\beta^{\star}$ is in hand, the scalar $\psi(\tilde k^{\star})$ can be computed similarly to~\eqref{eqn:psikdef}, by replacing $f_k$ with $f_{\tilde k^{\star}}(\cdot)= \tilde\beta_1^{\star} Z_{i_1}(\cdot)+\cdots+\tilde\beta_{s}^{\star}Z_{i_{s_0}}(\cdot)$, where it should be noted that the subscripts $i_1,\dots,i_{s_0}$ are the entries of $\mathbf{i}$.

To arrive at a simple overall runtime for computing $\tilde{\delta}_{1-\alpha}^{\,\textsc{ext}}$, suppose the for-loop is distributed so that the number of iterations per processor satisfies $N/m=\mathcal{O}(1)$. In addition, suppose that the number of test points satisfies $t=\mathcal{O}(n)$, and the data dimension satisfies $d=\mathcal{O}(s_0)$. Under these assumptions, the overall runtime to compute $\tilde{\delta}_{1-\alpha}^{\,\textsc{ext}}$, is $\mathcal{O}(s_0^2 n)$, which is quite manageable in comparison to the $\mathcal{O}(s_1^2n)$ cost of kernel ridge regression using RFF.

\section{THEORY}\label{sec:theory}
Here, we analyze the performance of Algorithm 1 when the RFF kernel approximation error is measured in a uniform entrywise sense, which is common in the literature
~\citep[e.g.][]{Rahimi2007,Sutherland2015,Liu2020}.
In particular, we use the norm \smash{$\|\tK-\K\|_{\infty}=\max_{1\leq j,j'\leq n}|\tK_{jj'}-\K_{jj'}|$.} Our main theoretical result shows that in the limit of large problem sizes $(n\to\infty)$, the estimate $\tve$ constructed in Algorithm 1 matches the performance of the ideal value $\vealph$ with respect to coverage probability.

\textbf{Assumptions.} We consider a sequence of kernel approximation problems indexed by $n=1,2,\dots$, where the dimension $d=d_n$ of the point set $\{x_1,\dots,x_n\}\subset\R^{d}$ is allowed to vary in an unrestricted manner as $n\to\infty$.  In addition, the kernel function $k=k_n$ may vary as $n\to\infty$, provided that it is of the type described in Section~\ref{sec:prelim}. That is, the kernel function $k$ is assumed to be shift-invariant and continuous with $k(0,0)=1$ for every $n$.

With regard to RFF and error estimation, the number of random features $s=s_n$ and bootstrap iterations $N=N_n$ in Algorithm 1 are both allowed to vary as $n\to\infty$, subject to two basic conditions: $N\to\infty$ and $\frac{\log(n)^5}{s}\to 0$.
\begin{theorem}\label{thm:main}
Suppose that the aforementioned assumptions hold.
Also let $\tve$ be computed with Algorithm 1 from an input matrix $\Z\in\R^{n\times s}$ that is generated as described in Section~\ref{sec:prelim}.
Then, for any fixed $\alpha\in(0,1)$, as $n\to\infty$, 
\begin{equation}\label{eqn:thmlim}
 \P\Big(\|\tK-\K\|_{\infty} \, \leq \, \tve\Big)   \ \to \ 1-\alpha.
\end{equation}
\end{theorem}
\textbf{Remarks.} Theorem~\ref{thm:main} has been presented in an asymptotic form for the sake of simplicity. An explicit rate of convergence can be found in the proof in Appendix A, which shows that the probability in~\eqref{eqn:thmlim} differs from $(1-\alpha)$ by a quantity that is at most $\mathcal{O}\big((\log(N)/N)^{1/2} + (\log(sn)^5/s)^{1/4} \big)$.
To interpret some other aspects of the result, it should be emphasized that the assumptions are mild, insofar far as the point set $\{x_1,\dots,x_n\}\subset\R^d$ is unrestricted with respect to its geometric structure and dimension $d$. Also, there are no extra assumptions on the kernel function beyond those that are ordinarily used in the study of RFF. Furthermore, the conditions on $N$ and $s$ are mild, since they allow both $N$ and $s$ to grow very slowly compared to $n$. On the other hand, to note a limitation of Theorem 1, it only deals with the typical version of RFF where the columns of the random matrix $\Z$ are independent (as in Section~\ref{sec:prelim}), and it does not cover some particular versions of RFF in which these columns may not be independent~\citep{Le2013,Recycling2016}. However, even in the typical setting, our proof utilizes cutting-edge results on the central limit theorem in high dimensions~\citep{CCKK2022}, and the challenge extending such results to account for dependence is at the frontier of research in high-dimensional probability.

\section{EXPERIMENTS}\label{sec:expt}
We demonstrate the empirical performance of our error estimates in three settings: kernel matrix approximation (Section~\ref{sec:matrix}), kernel ridge regression (Section~\ref{sec:ridge}), and kernel-based hypothesis testing (Appendix B). There are two main takeaways: First, the extrapolated estimates $\tilde{\ve}_{1-\alpha}^{\,\textsc{ext}}$ and $\tilde{\delta}_{1-\alpha}^{\,\textsc{ext}}$ closely track their targets $\vealph$ and $\dealph$ across different settings. Second, these estimates can be quickly computed with modest values of $s_0$ and $N$. A Python implementation of the experiments is available at the GitHub repository~\cite{junwenrepo}.

\subsection{Error estimation for RFF in kernel matrix approximation}\label{sec:matrix}
\begin{figure*}[!t]
	\centering
	\begin{subfigure}{1\textwidth}	
		\centering
		\DeclareGraphicsExtensions{.pdf}
		\begin{overpic}[width=0.31\textwidth]{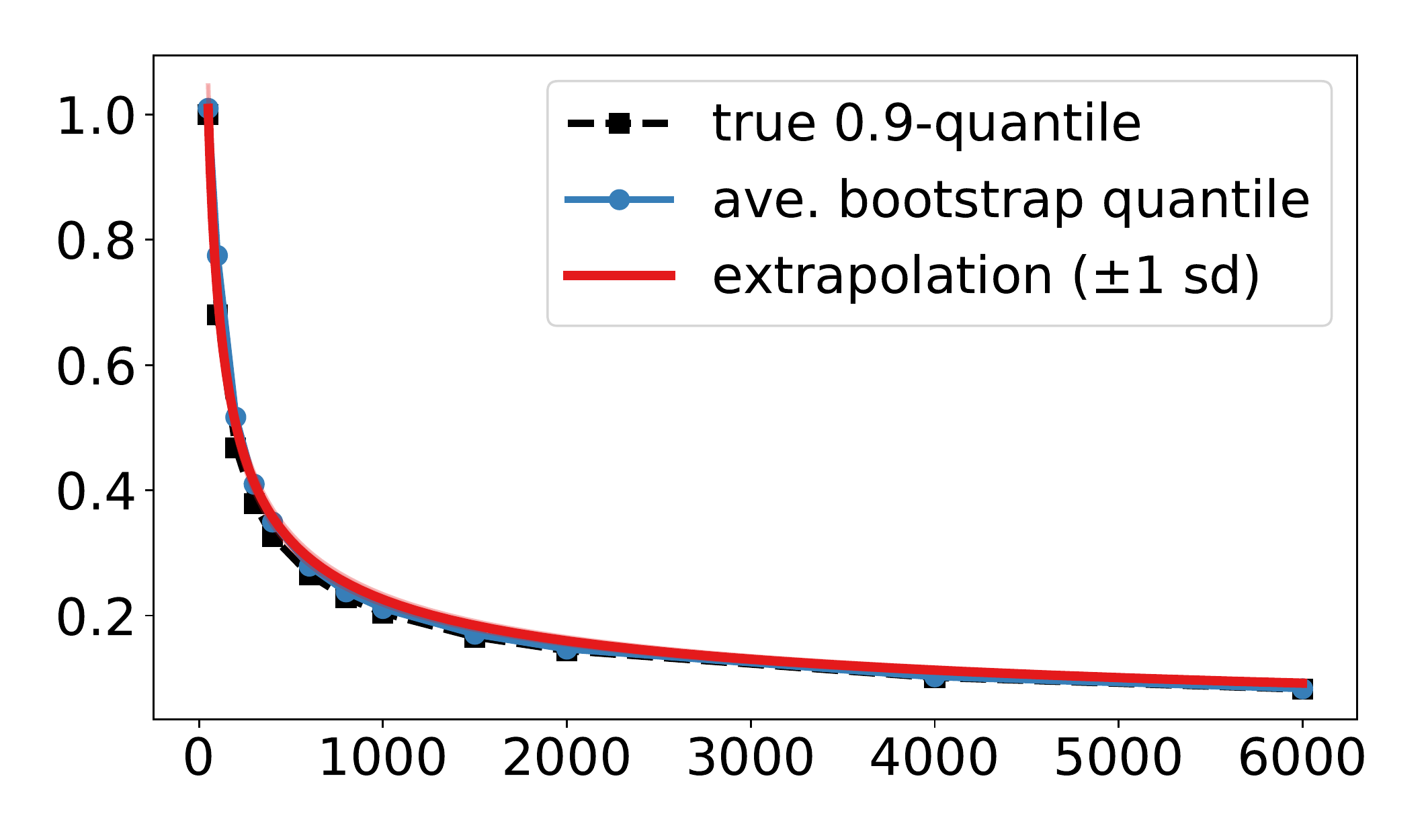} 
			\put(-6,24){\rotatebox{90}{\small $\ve_{0.9}$}} 
			\put(42,58){\color{black}{\small $\sigma=0.5$}} 
		\end{overpic}\hspace*{-0.2cm}
		~
		\begin{overpic}[width=0.31\textwidth]{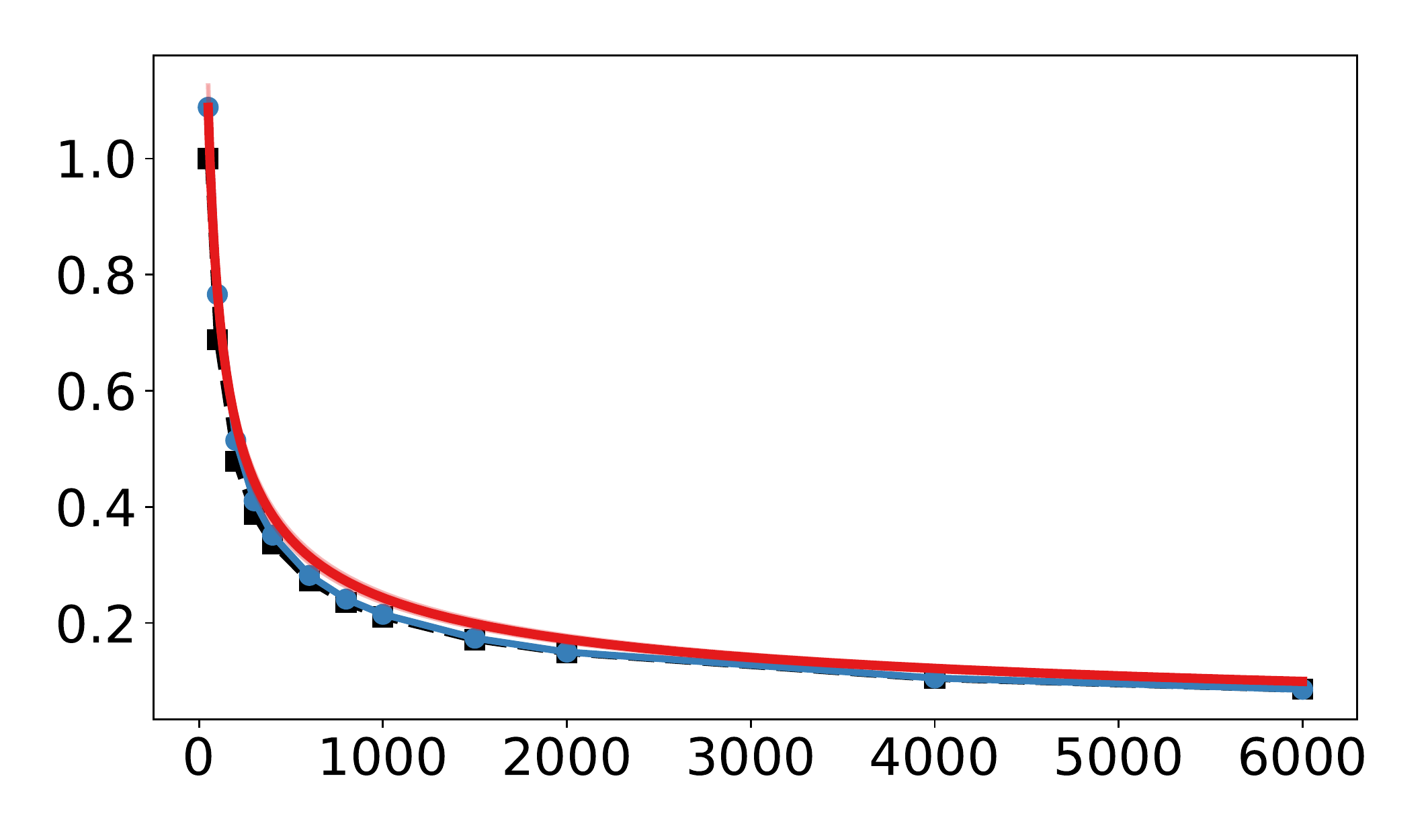} 
			\put(42,58){\color{black}{\small $\sigma=1.0$}} 
		\end{overpic}\hspace*{-0.2cm}
		~
		\begin{overpic}[width=0.31\textwidth]{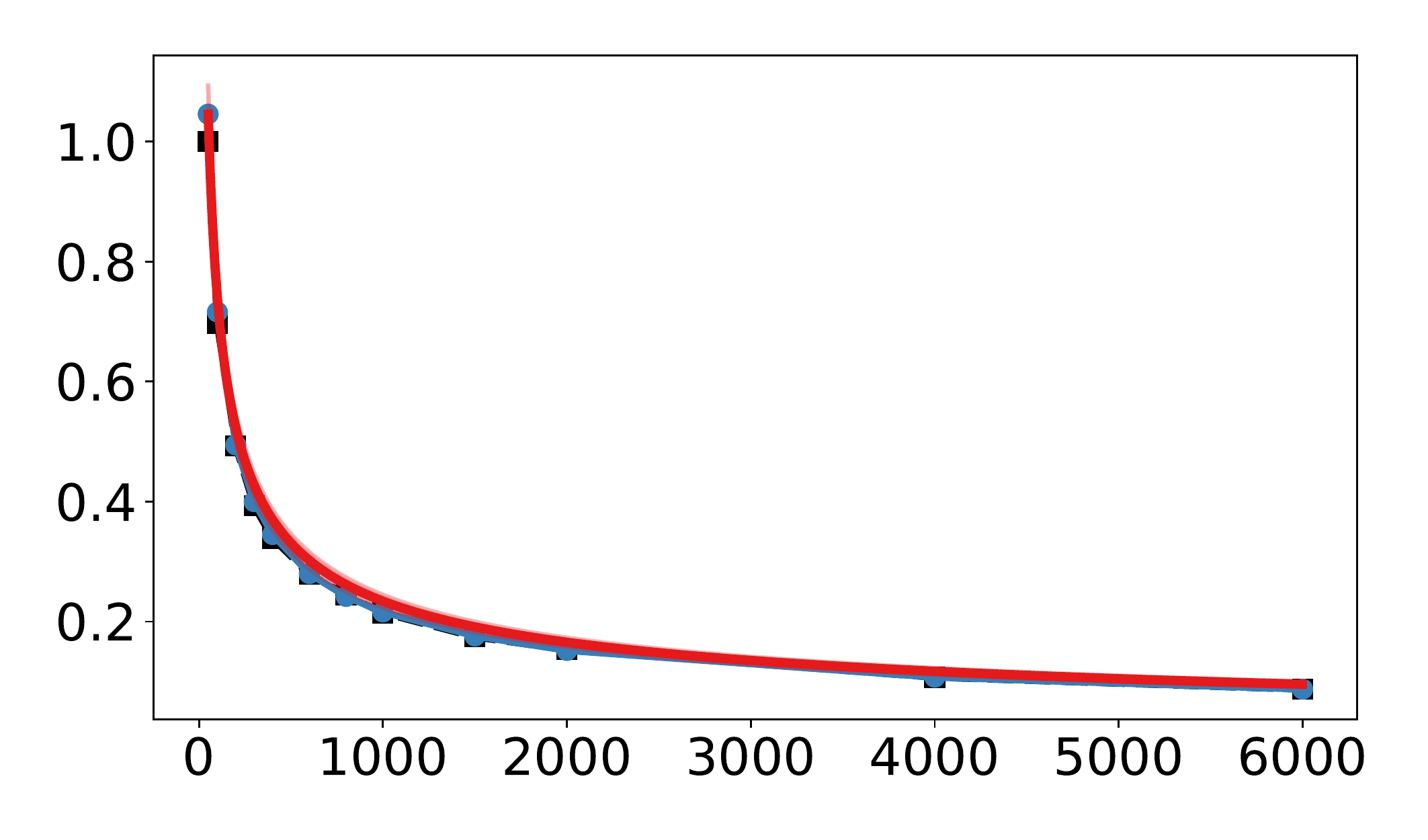} 
			\put(42,58){\color{black}{\small $\sigma=4.0$}} 
			\put(100,16){\rotatebox{90}{\scriptsize (Lorenz system)}}			
		\end{overpic}
	\end{subfigure}

	\begin{subfigure}{1\textwidth}	
		\centering
		\DeclareGraphicsExtensions{.pdf}
		\begin{overpic}[width=0.31\textwidth]{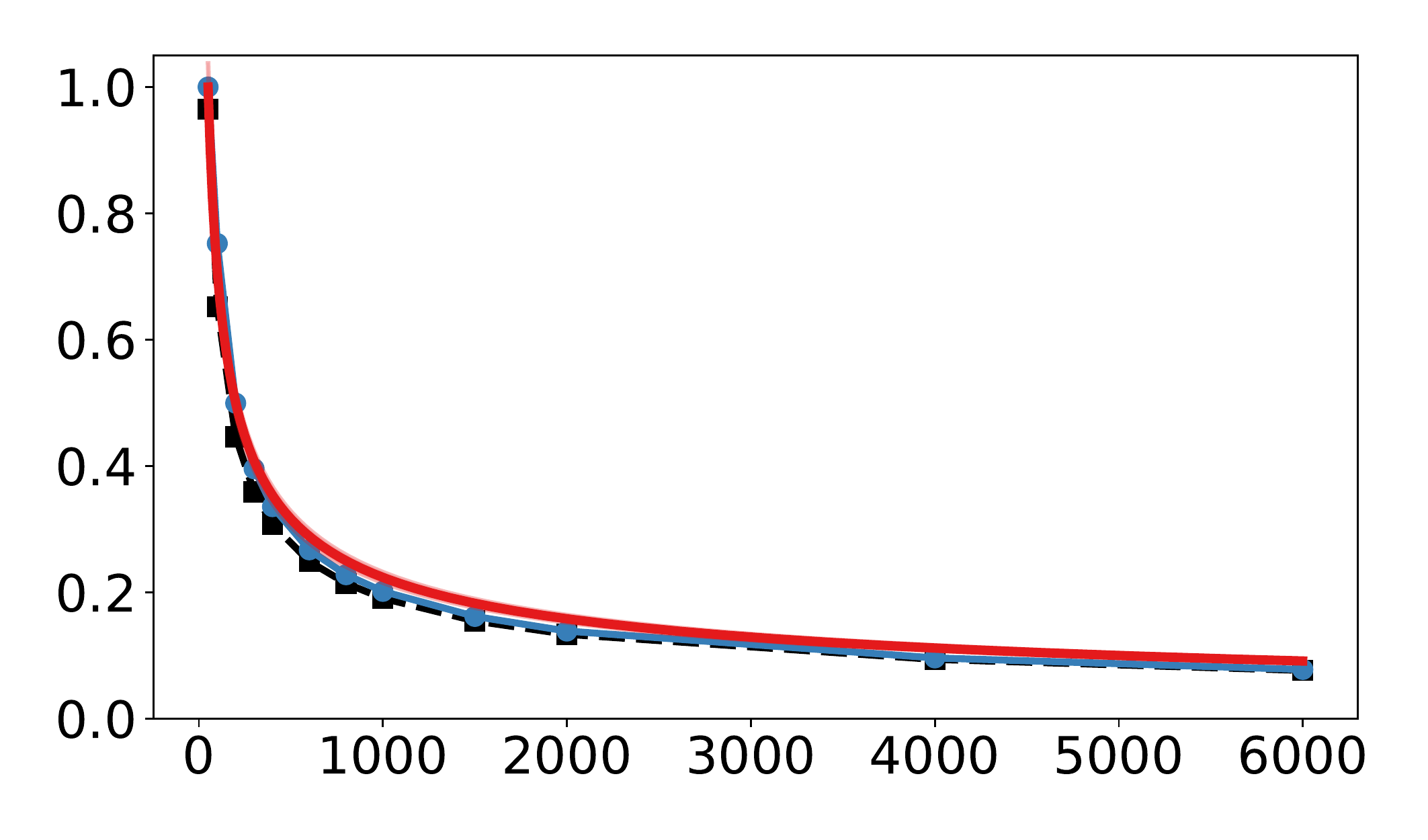} 
			\put(50,-2){\color{black}{\footnotesize $s$}}   
			\put(-6,24){\rotatebox{90}{\small $\ve_{0.9}$}}
		\end{overpic}\hspace*{-0.2cm}
		~
		\begin{overpic}[width=0.31\textwidth]{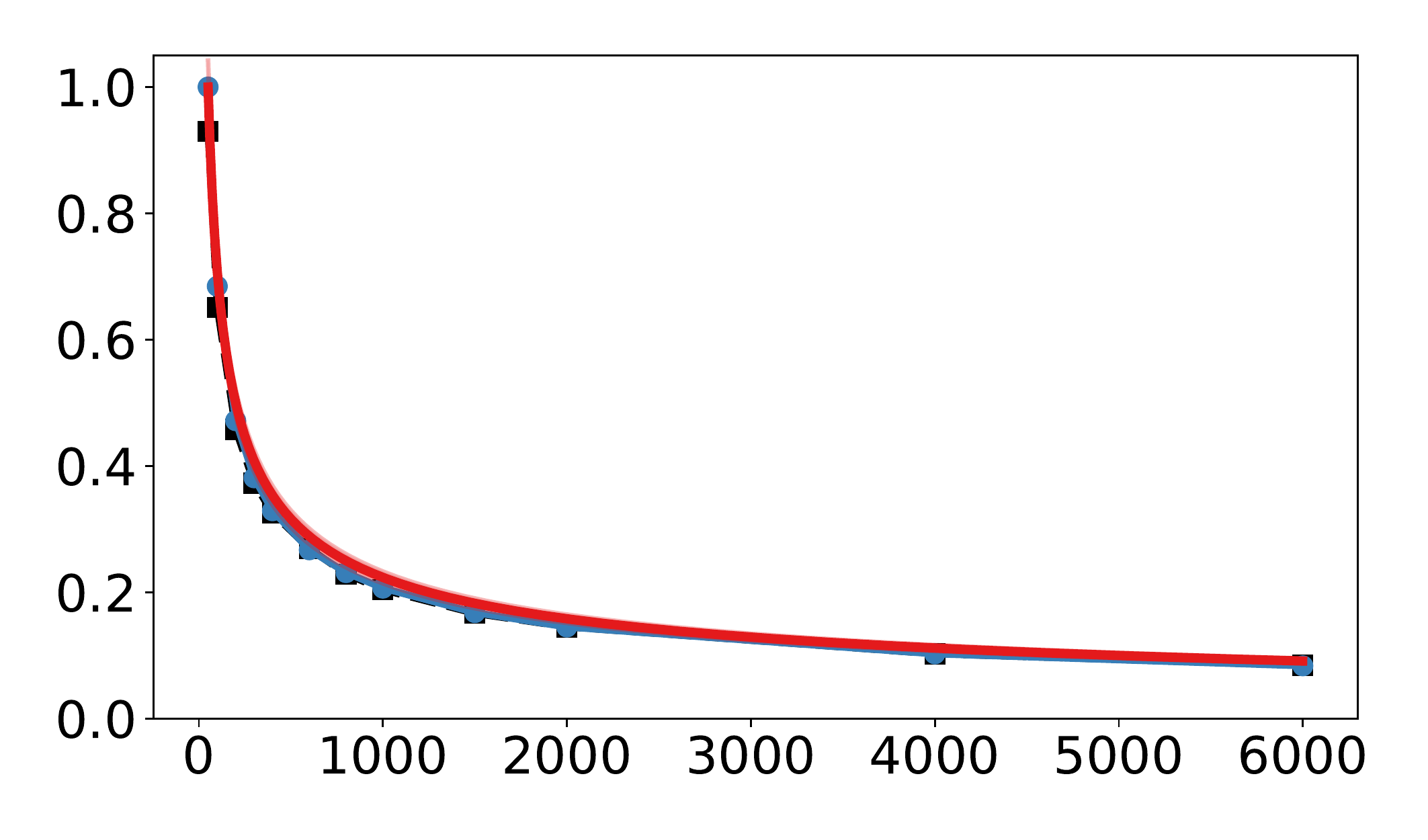} 
			\put(50,-2){\color{black}{\footnotesize $s$}}   
		\end{overpic}\hspace*{-0.2cm}
		~
		\begin{overpic}[width=0.31\textwidth]{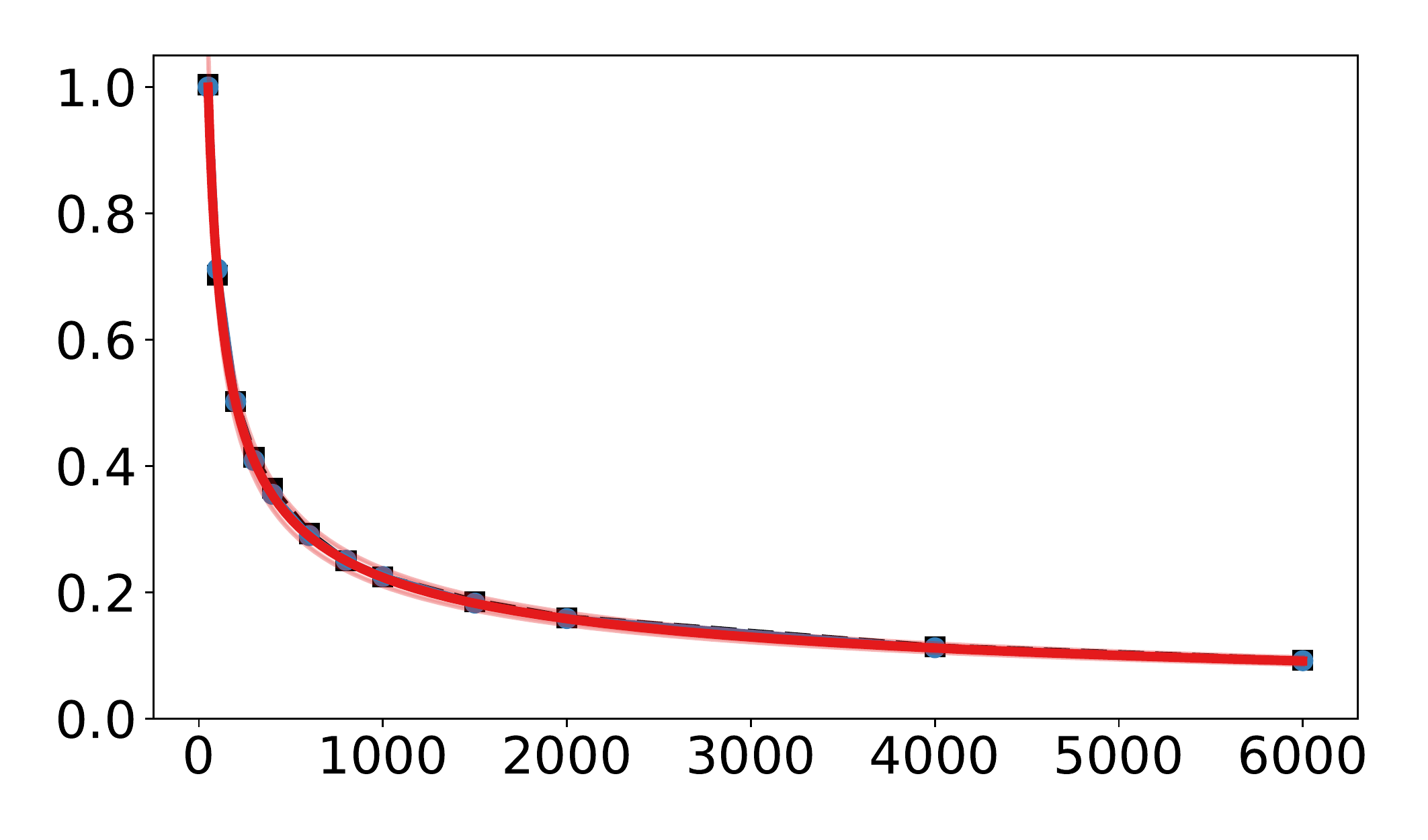} 
			\put(50,-2){\color{black}{\footnotesize $s$}}   
			\put(100,19){\rotatebox{90}{\scriptsize (MNIST)}}			
		\end{overpic}
	\end{subfigure}
%
	%
	\caption{(Estimation of $\ve_{0.9}$ for $\|\tK-\K\|_{\infty}$.) The top and bottom rows correspond respectively to the Lorenz system and MNIST datasets.	The columns correspond to choices of the kernel bandwidth $\sigma$. }
    \vspace{.1cm}
\label{fig:results_KMF_inf}
\end{figure*}

\begin{figure*}[!t]
	\centering
	\begin{subfigure}{1\textwidth}	
		\centering
		\DeclareGraphicsExtensions{.pdf}
		\begin{overpic}[width=0.31\textwidth]{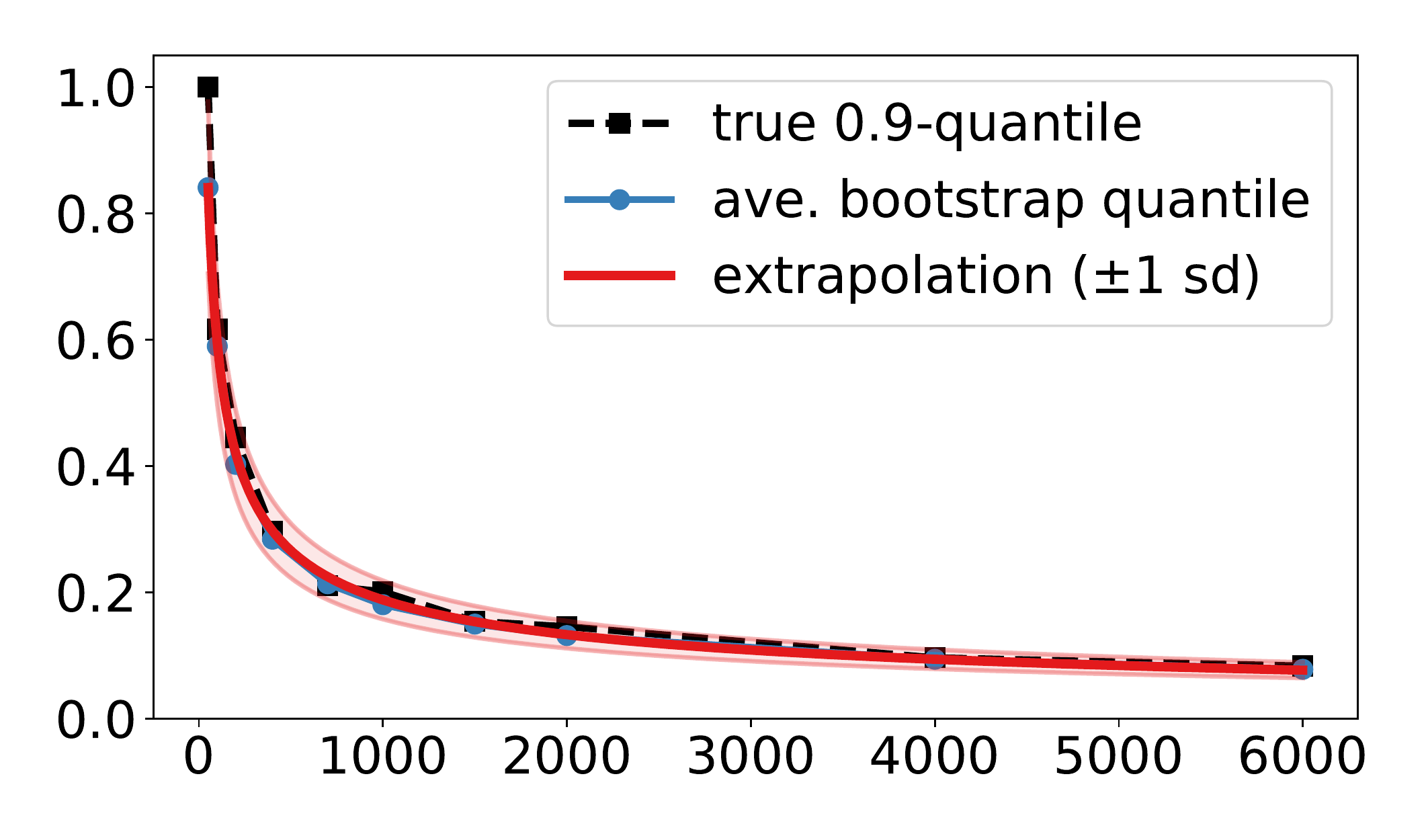} 
			\put(-6,26){\rotatebox{90}{\small $\ve_{0.9}$}}
			\put(42,58){\color{black}{\small $\sigma=0.5$}} 			
		\end{overpic}\hspace*{-0.2cm}
		~
		\begin{overpic}[width=0.31\textwidth]{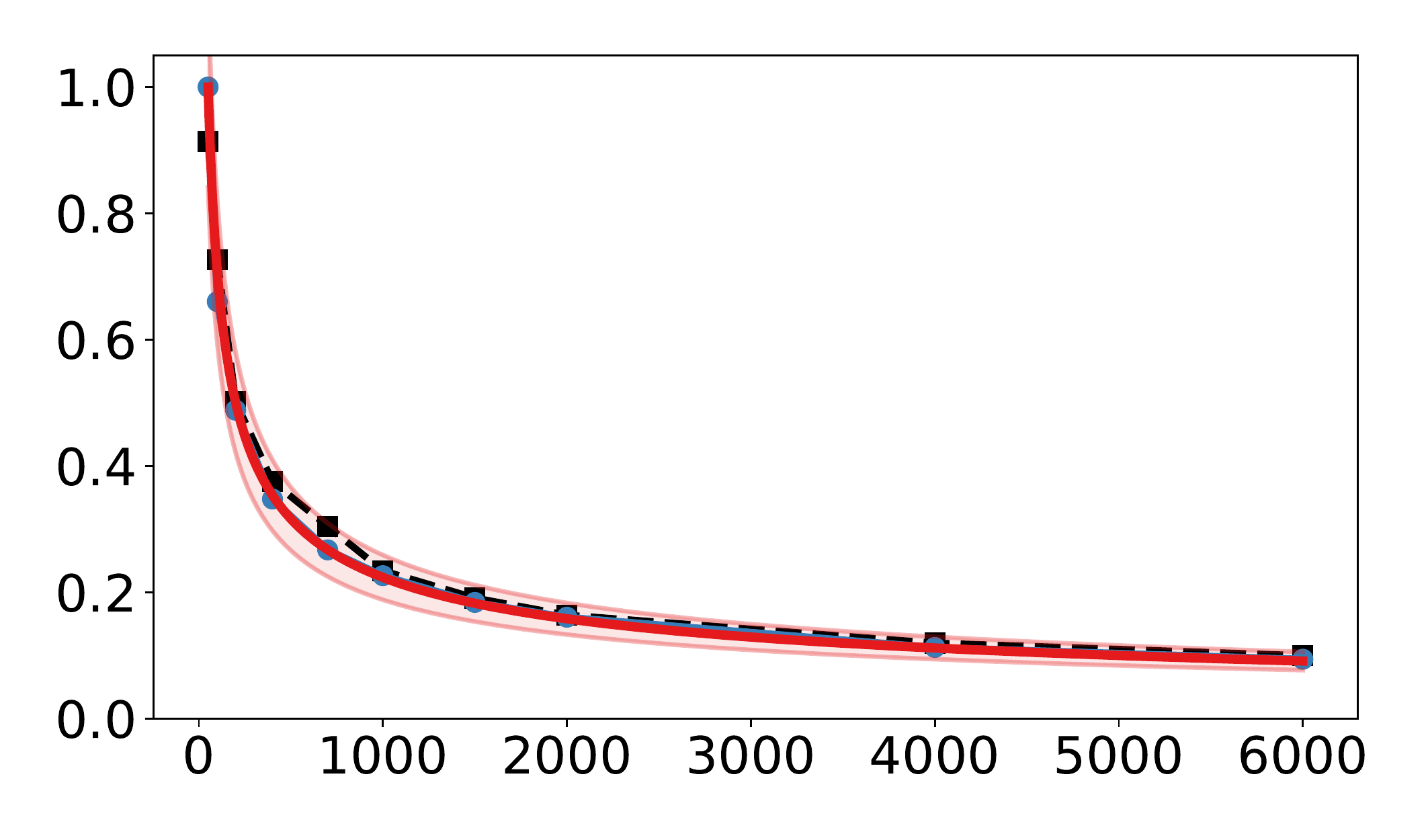} 
			\put(42,58){\color{black}{\small $\sigma=1.0$}} 			 			
		\end{overpic}\hspace*{-0.2cm}
		~
		\begin{overpic}[width=0.31\textwidth]{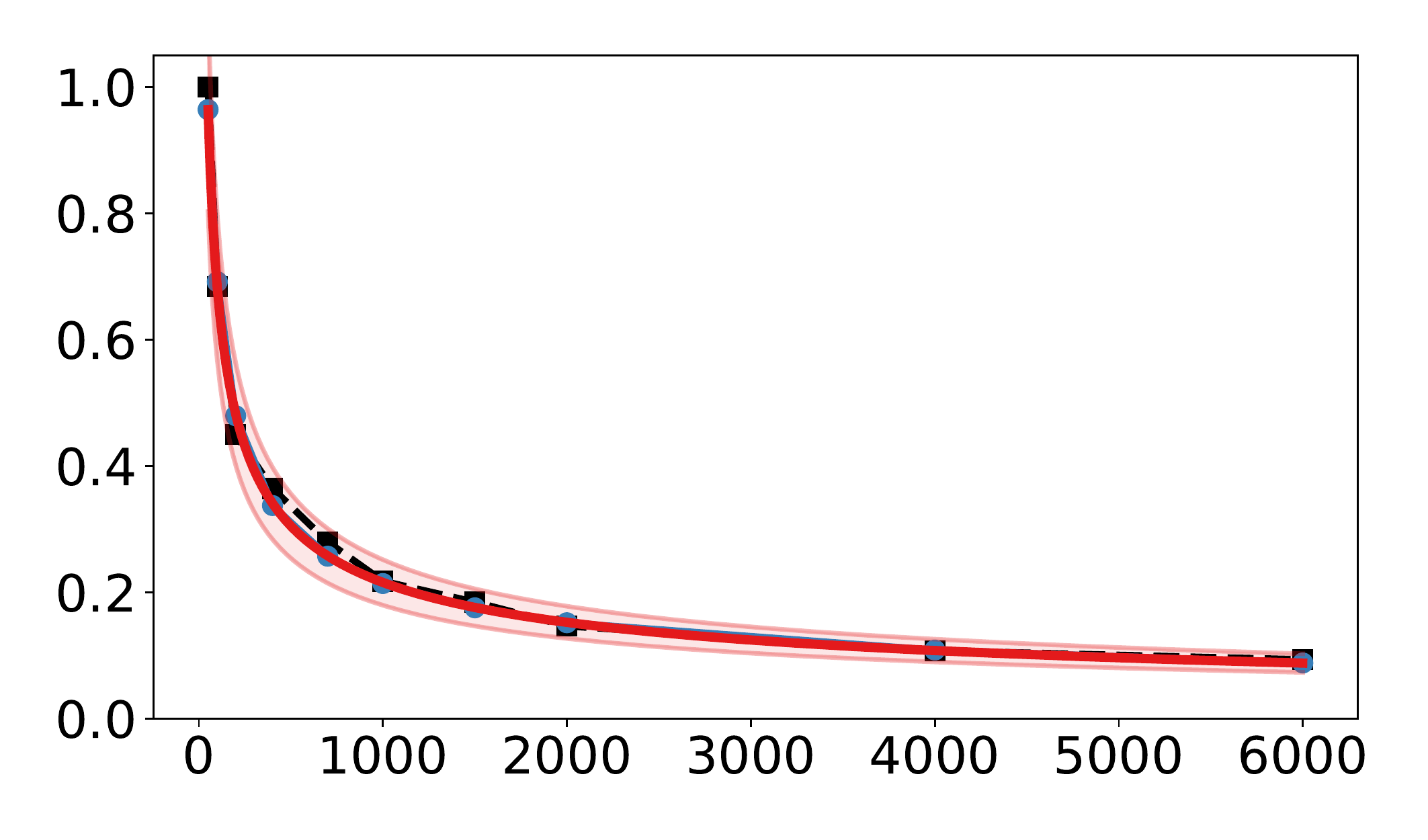} 
			\put(42,58){\color{black}{\small $\sigma=4.0$}} 
			\put(100,16){\rotatebox{90}{\scriptsize (Lorenz system)}}
		\end{overpic}
	\end{subfigure}
	
	\begin{subfigure}{1\textwidth}	
		\centering
		\DeclareGraphicsExtensions{.pdf}
		\begin{overpic}[width=0.31\textwidth]{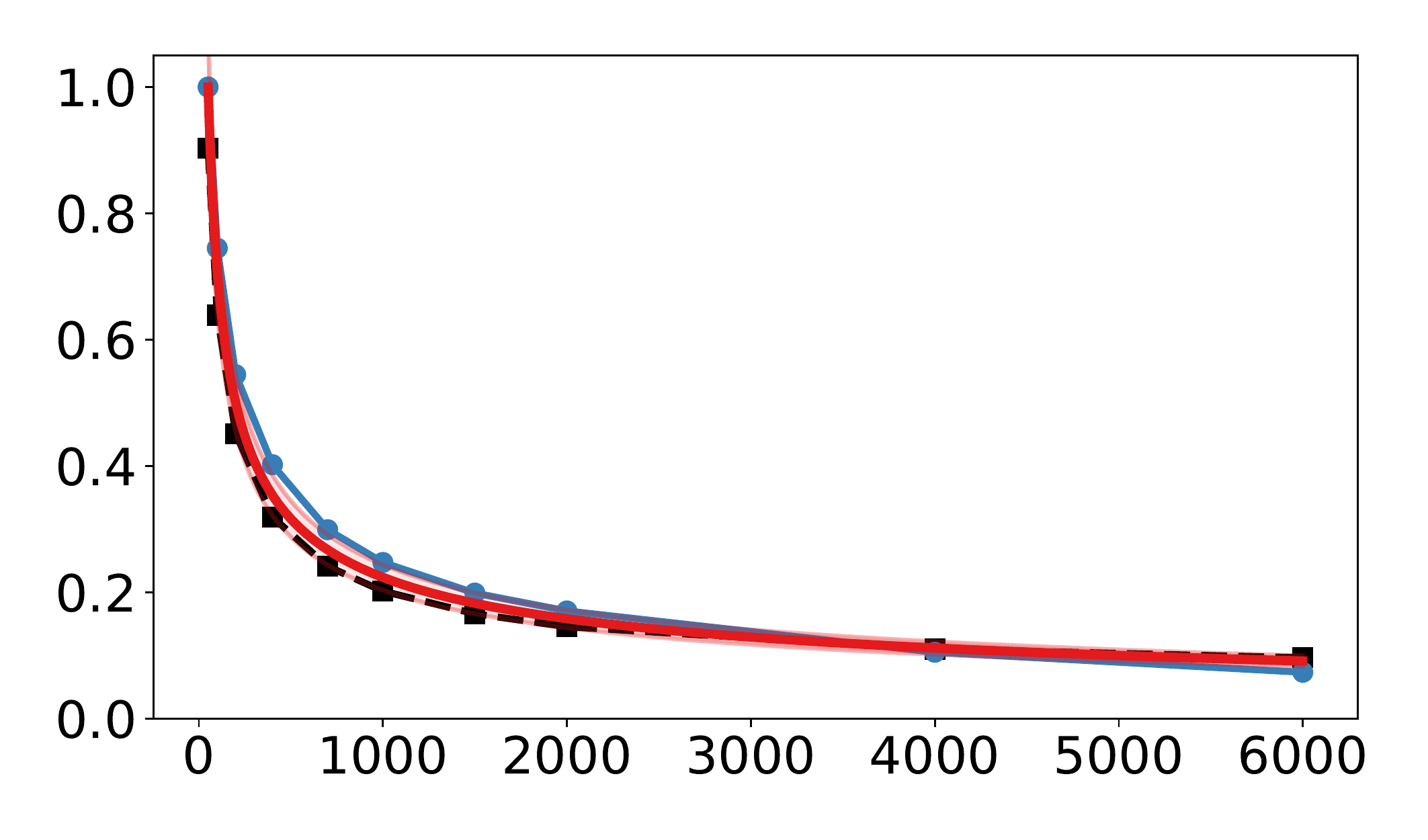} 
			\put(50,-2){\color{black}{\footnotesize $s$}}   
			\put(-6,24){\rotatebox{90}{\footnotesize $\ve_{0.9}$}}
		\end{overpic}\hspace*{-0.2cm}
		~
		\begin{overpic}[width=0.31\textwidth]{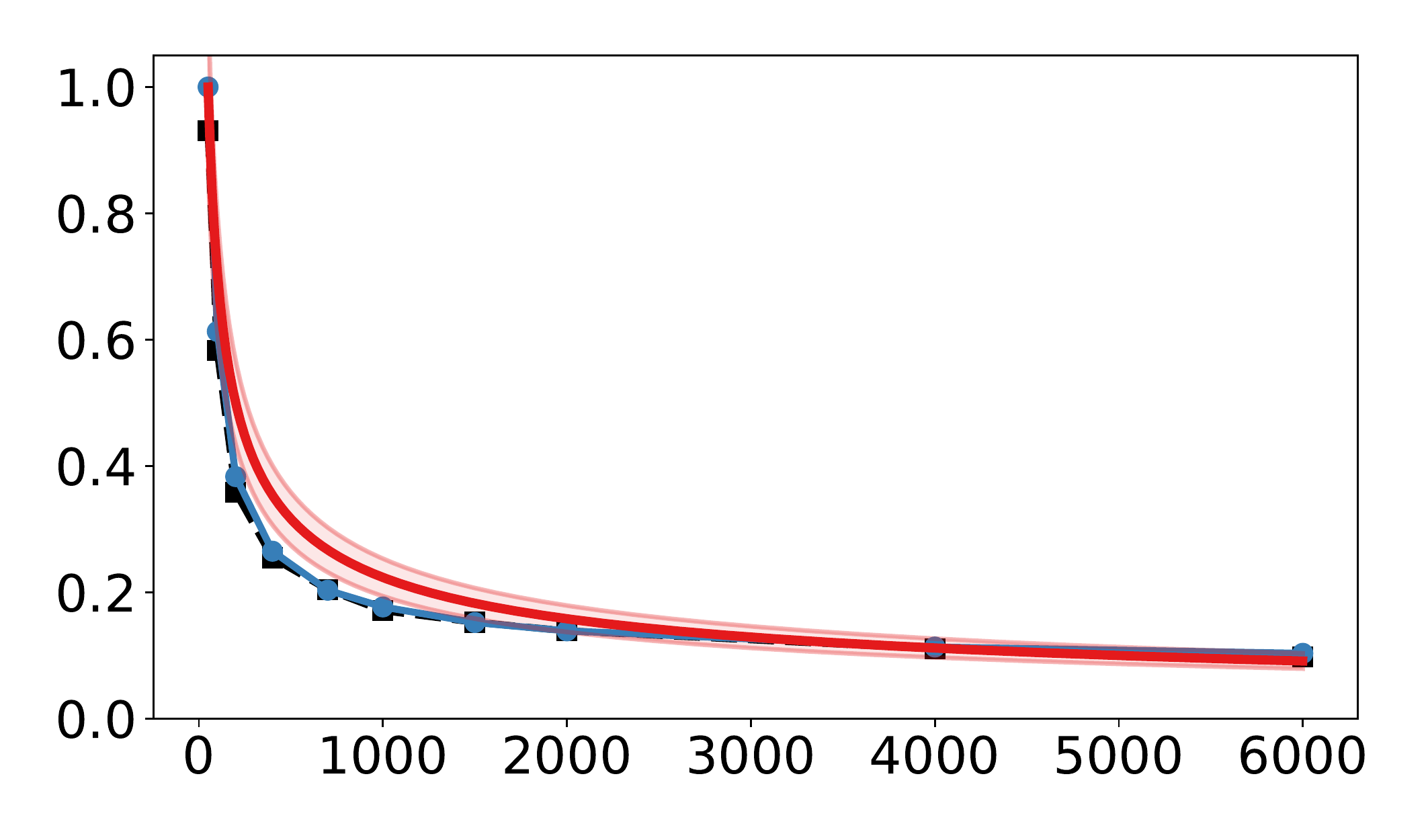} 
			\put(50,-2){\color{black}{\footnotesize $s$}}  
		\end{overpic}\hspace*{-0.2cm}
		~
		\begin{overpic}[width=0.31\textwidth]{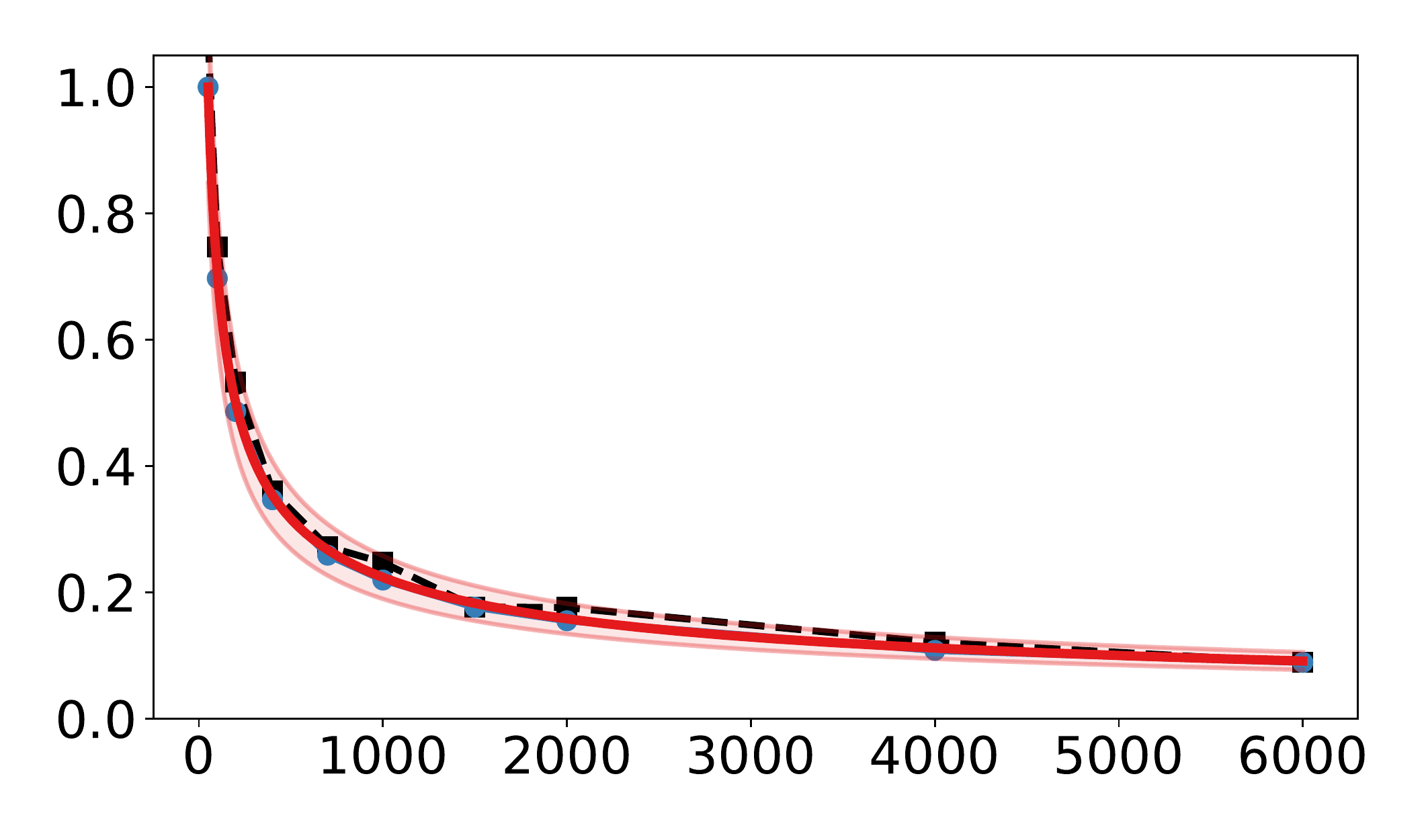} 
			\put(100,19){\rotatebox{90}{\scriptsize (MNIST)}}
			\put(50,-2){\color{black}{\footnotesize $s$}}  
		\end{overpic}
	\end{subfigure}

	\caption{(Estimation of $\ve_{0.9}$ for $\|\tK-\K\|_{\textup{op}}$.) The plots are organized analagously to Figure~\ref{fig:results_KMF_inf}.}	
    \vspace{.3cm}
\label{fig:results_KMF_op}
\end{figure*}

Here, we examine how accurate $\tve$ and $\tilde{\ve}_{1-\alpha}^{\textsc{ ext}}$ 
are as estimates of $\ve_{1-\alpha}$. This is done when matrix approximation error is measured through the $\ell_{\infty}$-norm $\|\tK-\K\|_{\infty}$ (Figure~\ref{fig:results_KMF_inf}), as well as the operator norm $\|\tK-\K\|_{\textup{op}}$ (Figure~\ref{fig:results_KMF_op}).

\textbf{Data examples.} The results are based on two datasets derived from: (1) the \emph{Lorenz system}~\citep{lorenz1963deterministic} and (2) the training set portion of MNIST~\citep{lecun1998gradient}. The Lorenz system is a well-known chaotic dynamical system, and we followed~\citep{erichson2018diffusion} by generating \smash{$n=25000$} points that reside on a trajectory in $\R^3$. The training set portion of MNIST consists of $n=50000$ points that represent 784-pixel images.

\textbf{Design of experiments.} The following procedures were used for both datasets, with the kernel matrix $\K\in\R^{n\times n}$ being computed directly from the data. For each value of $s$ in a grid ranging from 50 to 6000, we generated 300 realizations of the random matrix $\Z\in\R^{n\times s}$ as described in Section~\ref{sec:prelim}, using the probability distribution $\rho$ corresponding to the Gaussian kernel $\exp(-\|x-x'\|_2^2/(2\sigma^2))$ with $\sigma\in\{0.5,1,4\}$. In addition, for each realization of $\Z$, we computed the associated error variables $\|\tK-\K\|_{\infty}$ and $\|\tK-\K\|_{\textup{op}}$, where $\tK=\Z\Z^{\scriptstyle \top}$.
This provided us with a set of 300 realizations of each type of error variable, and we computed the 90th percentile of each set, treating it as ground truth for $\ve_{0.9}$ at each $s$.
In Figures~\ref{fig:results_KMF_inf} and~\ref{fig:results_KMF_op}, the value of $\ve_{0.9}$ at each $s$ is plotted with a black curve. To ease comparisons, the black curve was rescaled so that its initial value is 1 in each plot, and the associated blue and red curves (described below) were rescaled by the same factor.

Next, we applied Algorithm 1 with $N=30$ iterations to each realization of $\Z$, yielding 300 corresponding estimates $\tilde{\ve}_{0.9}$ at each $s$, and we plotted the average of these estimates with a blue curve. Also, from each of the 300 realizations of $\tilde{\ve}_{0.9}$ computed at $s_0=50$, we obtained the extrapolated estimates $\tilde{\ve}_{0.9}^{\!\textsc{ ext}}(s)$ using formula~\eqref{eqn:extrap} for all $50\leq s\leq 6000$. The average of the extrapolated estimates is plotted with a red curve, and a pink envelope signifies $\pm 1$ standard deviation. (Note that in some plots within Figure~\ref{fig:results_KMF_inf}, the pink envelope is almost entirely covered by the red curve.)

\textbf{Discussion of results.} It is clear that both the blue and red curves for $\tilde{\ve}_{0.9}$ and $\tilde{\ve}_{0.9}^{\,\textsc{ext}}$ closely track the black curve representing ground truth. Beyond this main point, the red curve deserves special attention---because it is based on extrapolation from only $s_0=50$ features. So, if the user constructs a ``preliminary'' kernel approximation with $50$ features, they can use Algorithm 1 to ``look ahead'' and accurately predict how error will decrease for larger choices of $s$, e.g.~up to $s=6000$. Computationally, this means Algorithm 1 can be run with a matrix $\Z$ that is $n\times 50$, rather than $n\times 6000$ for a non-extrapolated estimate, i.e. \emph{two orders of magnitude reduction}. Another important point is that the number of bootstrap iterations $N=30$ is so small that, with a dozen processors, only a few iterations are needed per processor. 
Lastly, the two figures show that the error estimates behave 
reliably across different norms, datasets, and bandwidths. 
%
%
%
%
%
%
%
\vspace{0.3cm}
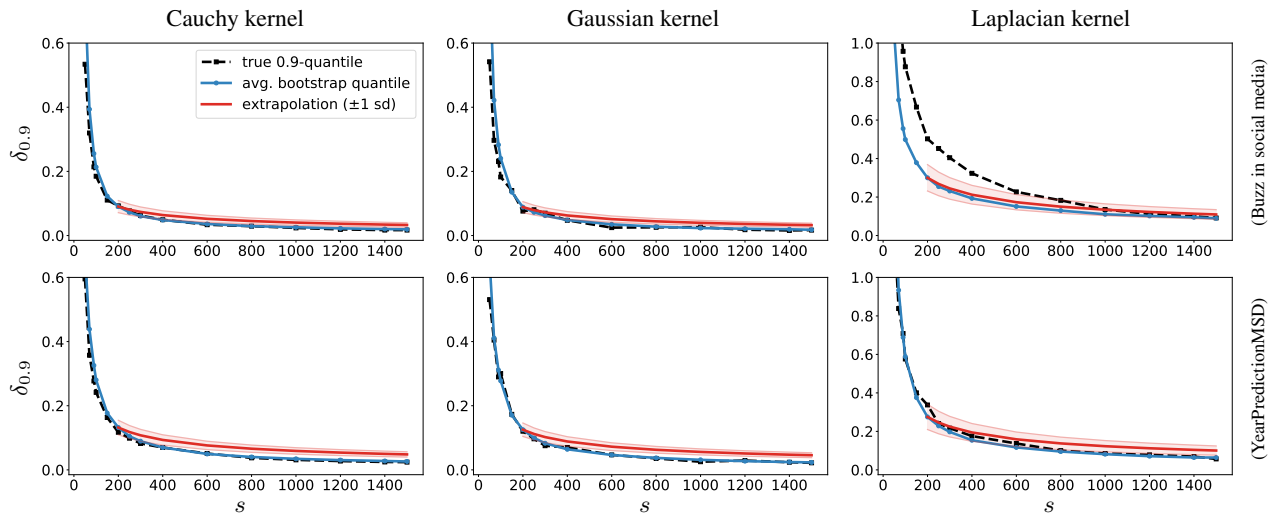
\begin{figure*}[!t]
\centering
\begin{subfigure}{1\textwidth}	
\centering
\DeclareGraphicsExtensions{.pdf}
\begin{overpic}[width=0.31\textwidth]{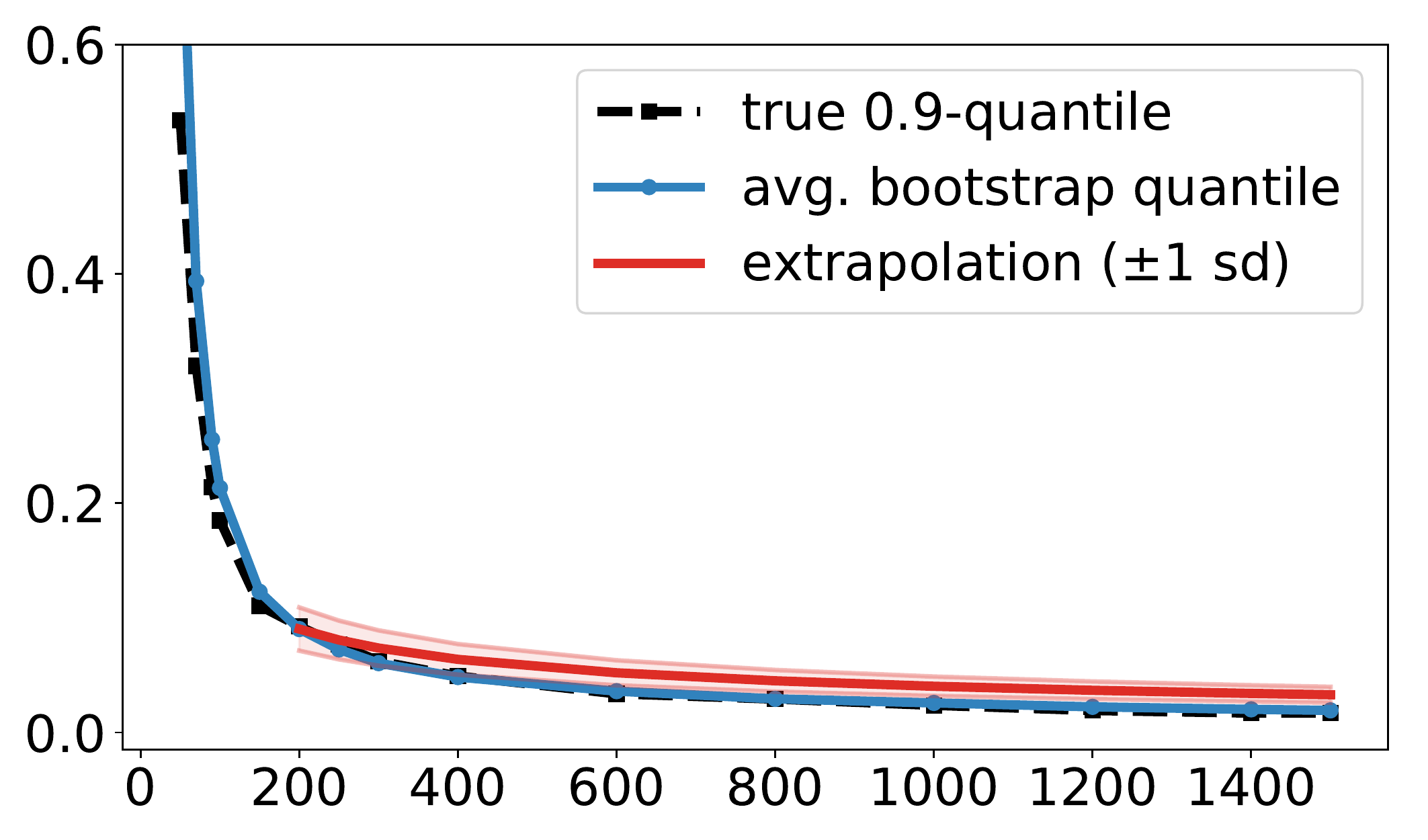} 
    \put(-6,26){\rotatebox{90}{\small $\delta_{0.9}$}}
	\put(33,60){\color{black}{\small Cauchy kernel }} 			
\end{overpic}\hspace*{-0.2cm}
~
\begin{overpic}[width=0.31\textwidth]{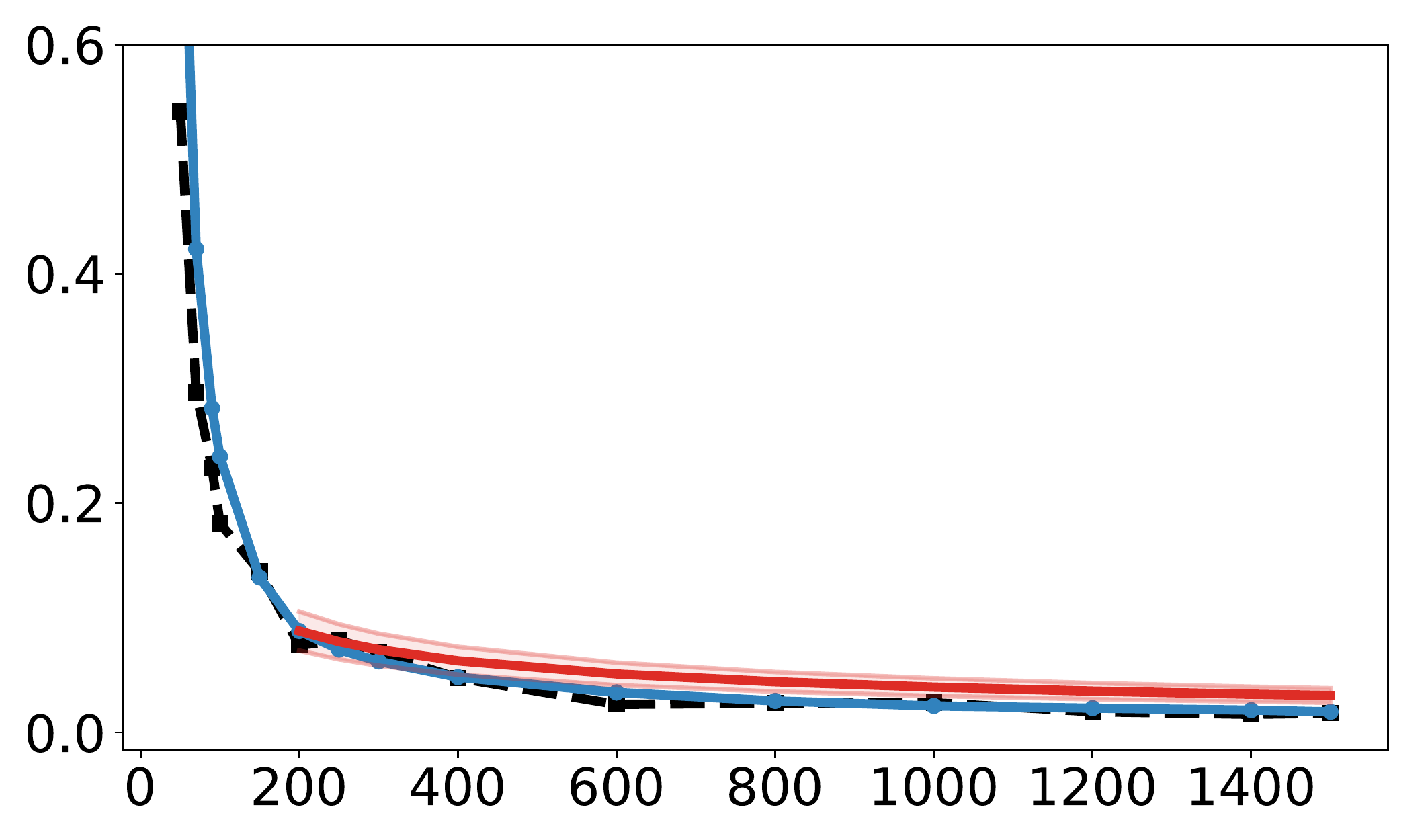} 
	\put(32,60){\color{black}{\small Gaussian kernel}} 			
\end{overpic}\hspace*{-0.2cm}
~
\begin{overpic}[width=0.31\textwidth]{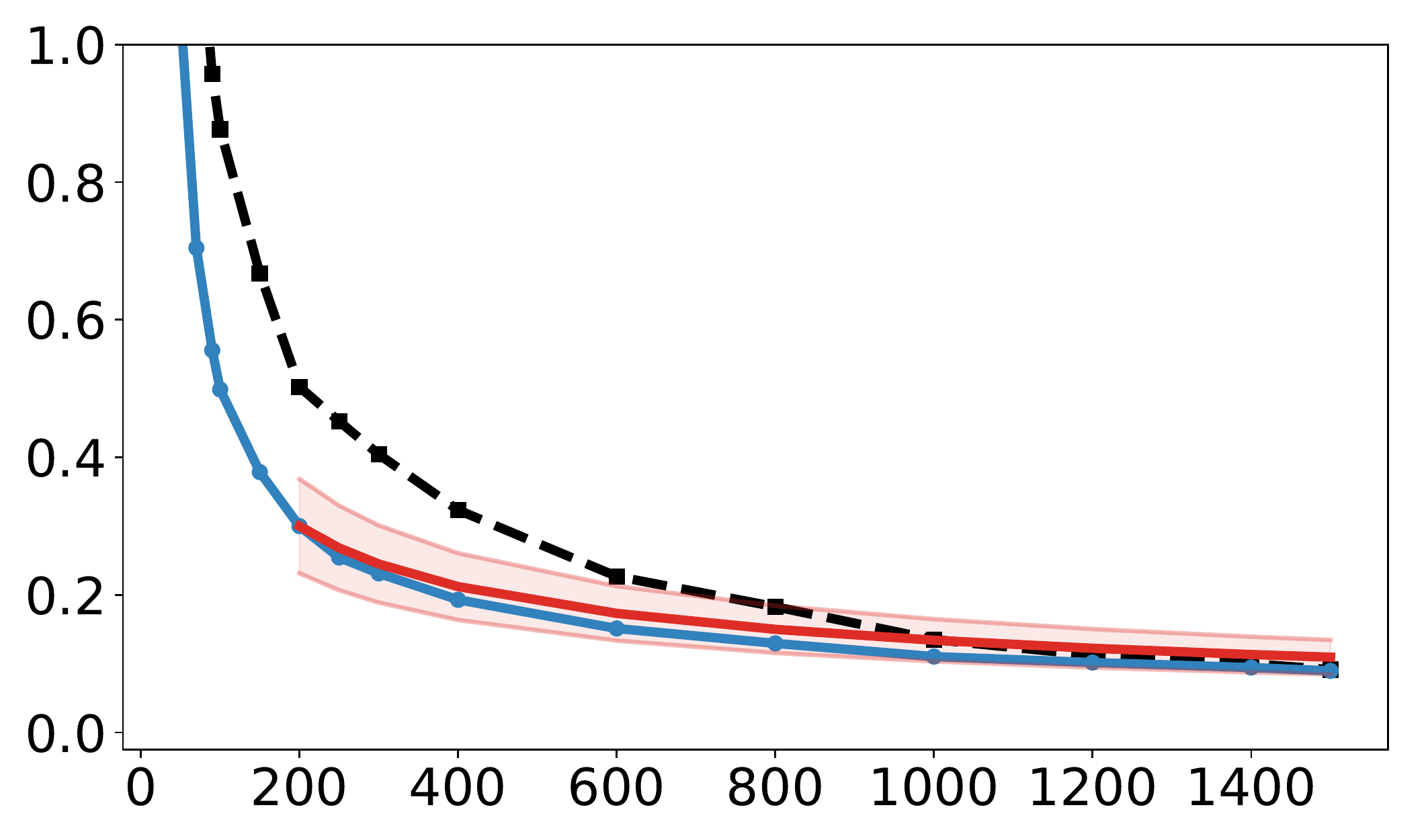} 
	\put(32,60){\color{black}{\small Laplacian kernel}} 
	\put(102,10){\rotatebox{90}{\scriptsize (Buzz in social media)}}
\end{overpic}
\end{subfigure}

\begin{subfigure}{1\textwidth}	
\centering
\DeclareGraphicsExtensions{.pdf}

\begin{overpic}[width=0.31\textwidth]{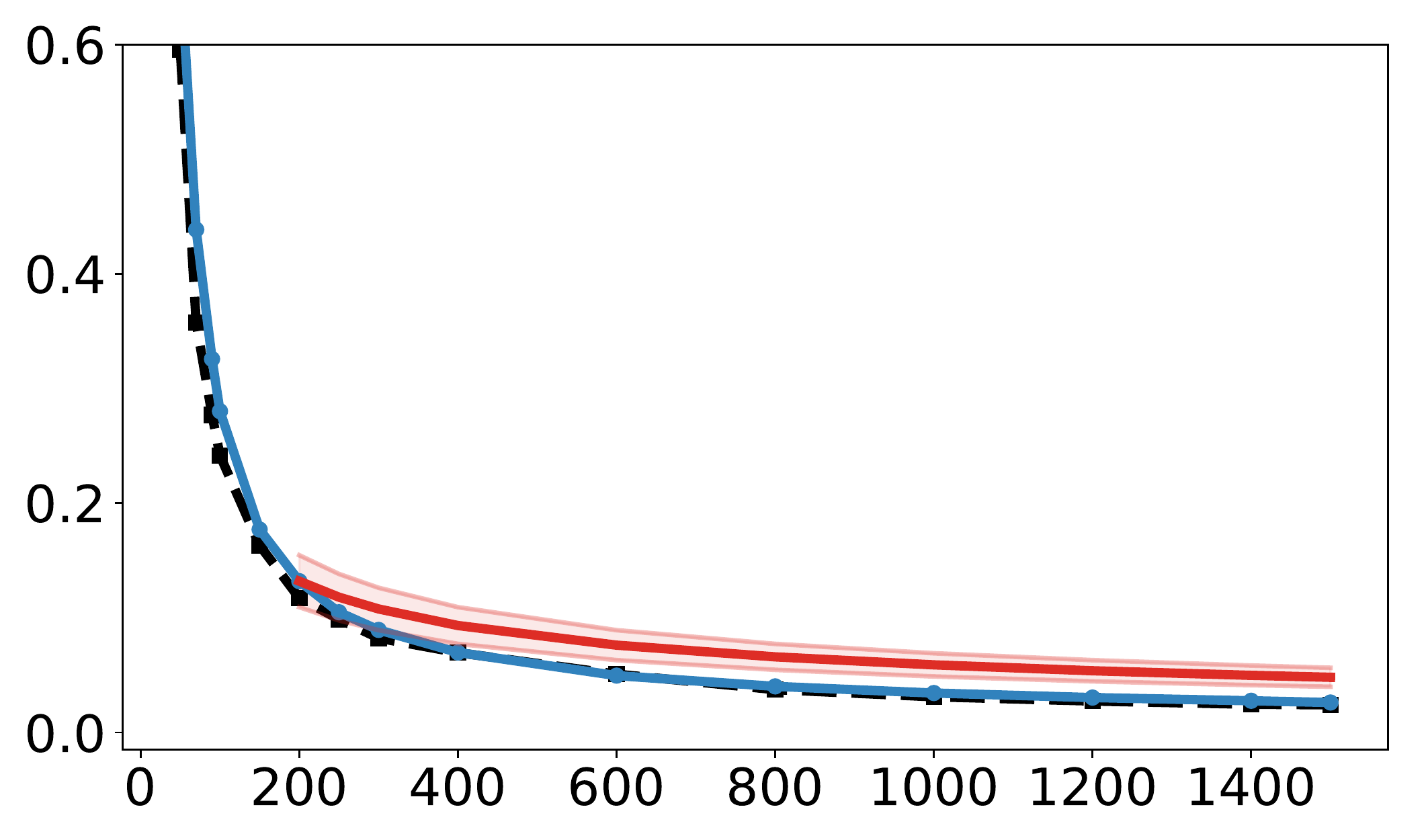} 
	\put(-6,26){\rotatebox{90}{\small $\delta_{0.9}$}}
	\put(50,-3){\color{black}{\footnotesize $s$}} 
\end{overpic}\hspace*{-0.2cm}
~
\begin{overpic}[width=0.31\textwidth]{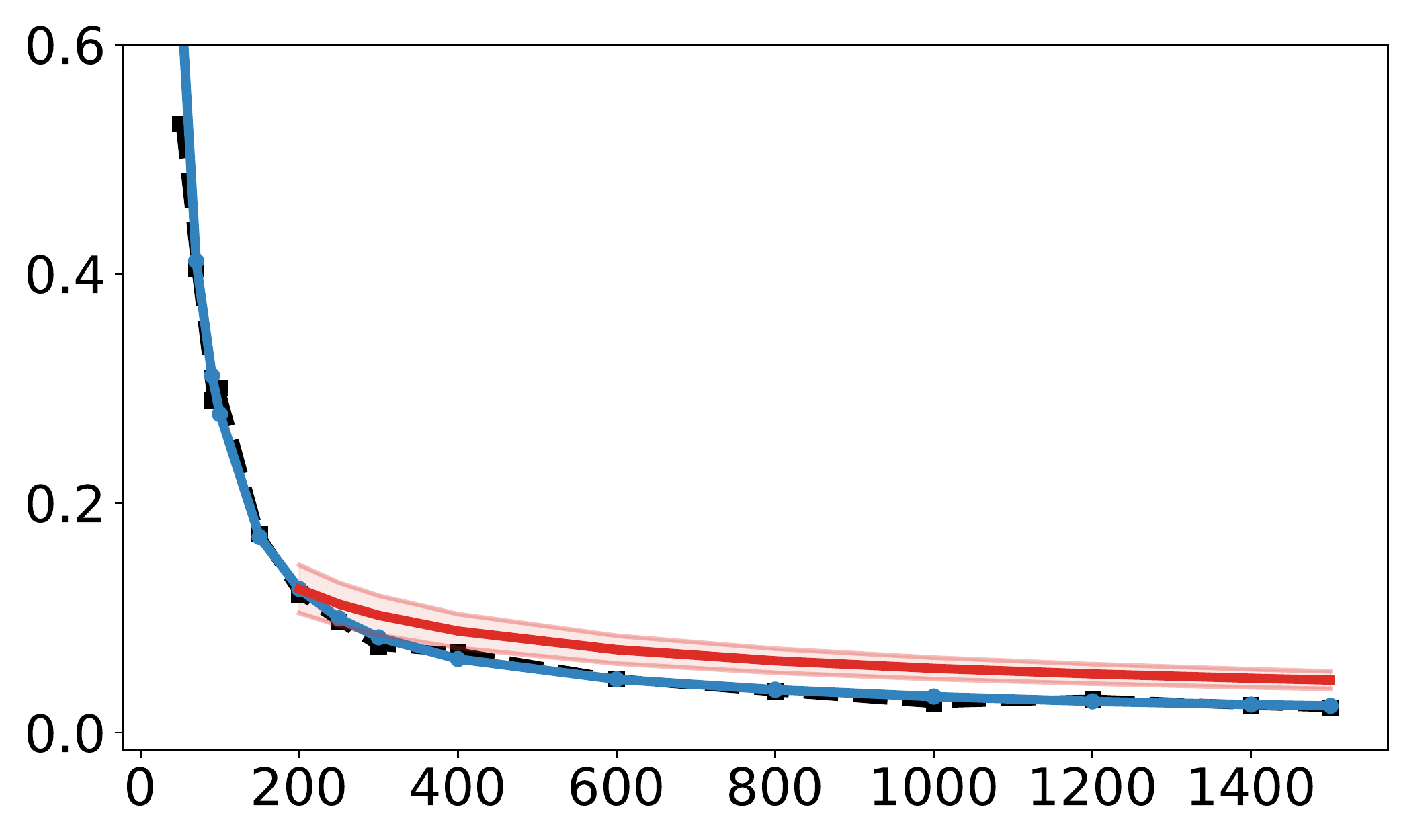} 
	\put(50,-3){\color{black}{\footnotesize $s$}} 
\end{overpic}\hspace*{-0.2cm}
~
\begin{overpic}[width=0.31\textwidth]{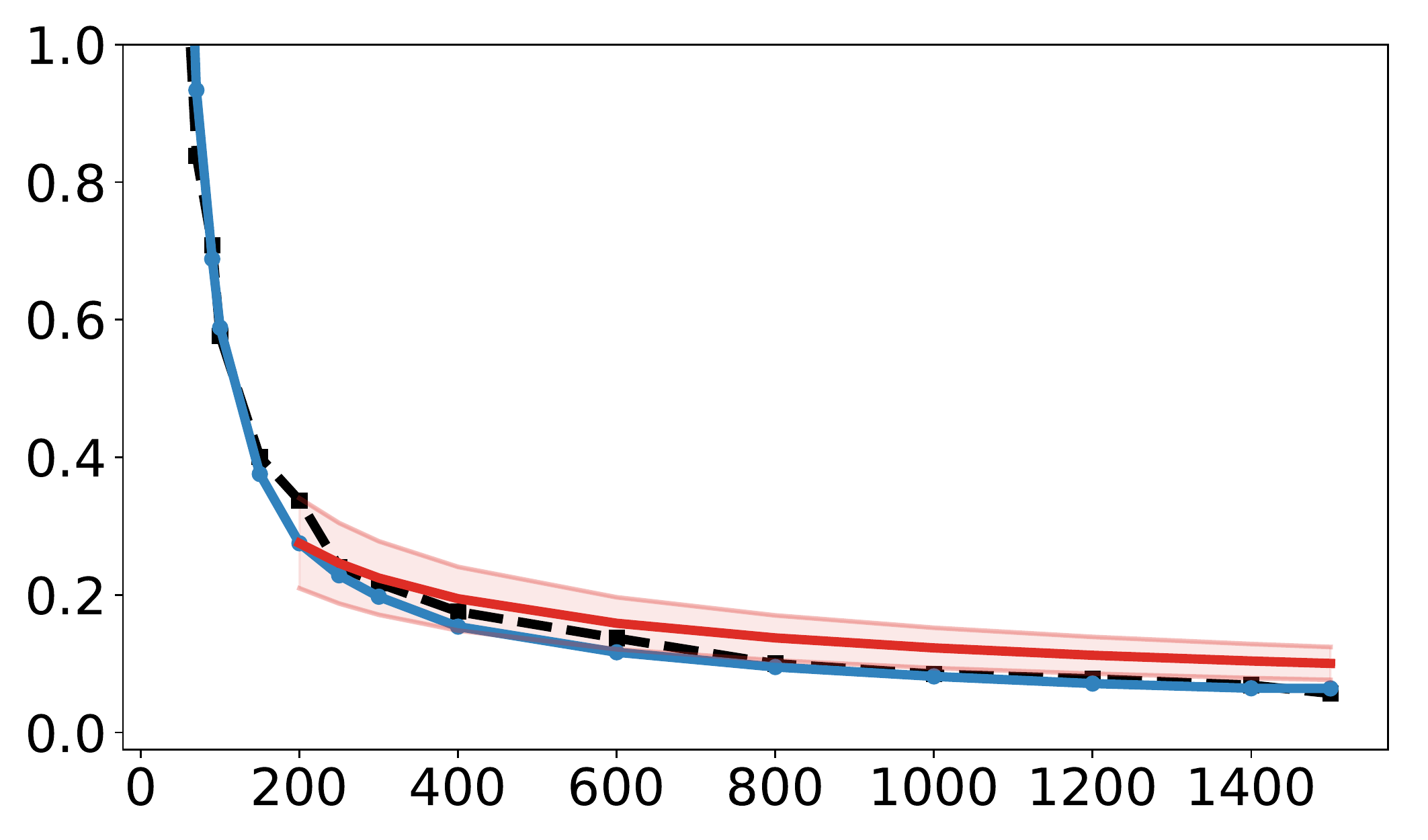} 
	\put(50,-3){\color{black}{\footnotesize $s$}} 
	\put(102,10){\rotatebox{90}{\scriptsize (YearPredictionMSD)}}
\end{overpic}
	
\end{subfigure}
%
\caption{(Estimation of $\delta_{0.9}$ for $\psi(\tilde k)-\psi(k)$.) The top and bottom rows correspond respectively to the two regression datasets. The columns correspond to the three different kernels.}
%
%
\label{fig:results_kernel_ridge_regrssion}
\end{figure*}
\subsection{Error estimation for RFF in kernel ridge regression}\label{sec:ridge}

Now we turn our attention to estimating how much error is created by RFF in kernel ridge regression.

\textbf{Data examples.} We used two regression datasets, each consisting of $(x,y)$ pairs in $\R^d\times \R$ with $d=50$. Each dataset $\D$ was partitioned as $\D=\D_{\text{train}}\cup\D_{\text{test}}$, with $|\D_{\text{test}}|=3000$ and  $n=|\D_{\text{train}}|=27000$. To obtain two different versions of $\D$ with these specifications, we uniformly subsampled 30000 rows and 50 columns from the datasets \emph{YearPredictionMSD} and \emph{Buzz in social media} in the repository~\citep{UCI}. For both versions of $\D$, we applied the standard normalization function `MinMaxScaler' from scikit-learn
to all the $x$ vectors, and in the case of YearPredictionMSD we took the square-root of the $y$ values due to their wide range.

\textbf{Design of experiments.} For a kernel $k$, let $\psi(k)$ denote the mean-squared test error of the associated ridge regression function, as defined in~\eqref{eqn:psikdef}. Also, let $\delta_{0.9}$ denote the 90th percentile of the random variable \smash{$\psi(\tilde k)-\psi(k)$}, which measures the extra prediction error due to RFF. 
The experiments here were organized analogously to those in Section~\ref{sec:matrix}, with $(\delta_{0.9}$, $\tilde{\delta}_{0.9}$, $\tilde{\delta}_{\,0.9}^{\textsc{\,ext}})$ playing the roles of $(\ve_{0.9}$, $\tilde{\ve}_{0.9}$, $\tilde{\ve}_{\,0.9}^{\textsc{\,ext}})$. Hence, the colored curves and the envelope can be interpreted in the same way. Also, as before, we generated 300 realizations of $\Z$ and used $N=30$ at each value of $s$. There are only a few notable details that are specific to the current setting. First, we computed $\tilde{\delta}_{0.9}^{\,\textsc{ext}}$ by extrapolating from the initial value $s_0=200$, and we always fixed the regression tuning parameter at $\lambda=1$. Second, all the curves were multiplied by the number $1/\psi(k)$ so that they can be more naturally viewed on a scale relative to the mean-squared test error of $f_k$.  Third, we performed the experiments using three different kernels: the Gaussian kernel $ \exp(-\|x-x'\|_2^2/10)$, the Laplacian kernel $\exp(-\|x-x'\|_1/10)$, and the Cauchy kernel $\prod_{j=1}^d 1/(1 + \Delta_j^2 / 10)$ where $\Delta=x-x'$.

\textbf{Discussion of results.} Figure~\ref{fig:results_kernel_ridge_regrssion} shows that in kernel ridge regression, the error estimates $\tilde{\delta}_{0.9}$ and $\tilde{\delta}_{\,0.9}^{\textsc{\,ext}}$ perform well, and with qualitatively similar characteristics to the error estimates in Section~\ref{sec:matrix}. However, this setting is more challenging, since a larger value of $s_0=200$ is needed, and since $\tilde{\delta}_{\,0.9}^{\textsc{\,ext}}$ shows a slight upward bias for large $s$.
Nevertheless, an upward bias may be preferred as being safer than a downward bias in the context of error estimation. 
In addition, Figure~\ref{fig:results_kernel_ridge_regrssion} shows that the error estimates largely maintain their accuracy across different choices of kernels.

\section{CONCLUSION}
Despite the broad impact that RFF has had in scaling up kernel methods,
a longstanding difficulty for users is that they do not know the actual errors of RFF approximations. This paper offers the first systematic approach to numerically estimate these errors. Our approach also overcomes practical limitations of analytical worst-case error bounds, because the error estimates are tailored to the user's specific inputs, and are very flexible with respect to different problem settings and error metrics.
Computationally, our approach leverages both parallelism and extrapolation so that the additional step of error estimation is affordable in relation to RFF itself. Also, our approach can enhance the efficiency of RFF by guiding the user to select $s$ in a data-adaptive way. From the standpoint of theory, we have provided a guarantee in the context of kernel matrix approximation, showing that our error estimates perform properly under mild assumptions. Furthermore, we have demonstrated empirically that our error estimates are quite accurate in a variety of tasks.

Looking ahead to future work, it is important to recognize that there are many variants and uses of RFF that go beyond the setup considered here. 
For example, our approach might be adapted to settings involving rotation-invariant kernels~\citep{lyu2017spherical,choromanski2017angular}, low-precision and quantized kernel estimators \citep{zhang2019low, li2021quantization}, or random features that are not independent~\citep{Le2013,Recycling2016}.

\section*{Acknowledgements}
MEL gratefully acknowledges partial support from NSF grant DMS-1915786. NBE would like to acknowledge partial support from the U.S. Department of Energy, Office of Science, Office of Advanced Scientific Computing Research, Scientific Discovery through Advanced Computing (SciDAC) program, under Contract Number DE-AC02-05CH11231, and the National Energy Research Scientific Computing Center (NERSC), operated under Contract No. DE-AC02-05CH11231 at Lawrence Berkeley National Laboratory. The authors thank all the reviewers for their helpful and constructive feedback.

\nocite{}
\bibliographystyle{apalike}
\bibliography{reference}

\begin{thebibliography}{}

\bibitem[Ahfock et~al., 2021]{Ahfock2021}
Ahfock, D.~C., Astle, W.~J., and Richardson, S. (2021).
\newblock Statistical properties of sketching algorithms.
\newblock {\em Biometrika}, 108(2):283--297.

\bibitem[Ainsworth and Oden, 2011]{Ainsworth:2011}
Ainsworth, M. and Oden, J.~T. (2011).
\newblock {\em A Posteriori Error Estimation in Finite Element Analysis},
  volume~37.
\newblock John Wiley \& Sons.

\bibitem[Avron et~al., 2017]{Avron2017}
Avron, H., Kapralov, M., Musco, C., Musco, C., Velingker, A., and Zandieh, A.
  (2017).
\newblock Random {F}ourier features for kernel ridge regression:
  {A}pproximation bounds and statistical guarantees.
\newblock In {\em International Conference on Machine Learning}.

\bibitem[Babu{\v{s}}ka and Rheinboldt, 1978]{Babuska2:1978}
Babu{\v{s}}ka, I. and Rheinboldt, W.~C. (1978).
\newblock Error estimates for adaptive finite element computations.
\newblock {\em SIAM Journal on Numerical Analysis}, 15(4):736--754.

\bibitem[Bank and Weiser, 1985]{Bank1985}
Bank, R.~E. and Weiser, A. (1985).
\newblock Some a posteriori error estimators for elliptic partial differential
  equations.
\newblock {\em Mathematics of Computation}, 44(170):283--301.

\bibitem[Bickel and Yahav, 1988]{BickelYahav}
Bickel, P.~J. and Yahav, J.~A. (1988).
\newblock Richardson extrapolation and the bootstrap.
\newblock {\em Journal of the American Statistical Association},
  83(402):387--393.

\bibitem[Chernozhuokov et~al., 2022]{CCKK2022}
Chernozhuokov, V., Chetverikov, D., Kato, K., and Koike, Y. (2022).
\newblock {Improved central limit theorem and bootstrap approximations in high
  dimensions}.
\newblock {\em The Annals of Statistics}, 50(5):2562 -- 2586.

\bibitem[Choromanski and Sindhwani, 2016]{Recycling2016}
Choromanski, K. and Sindhwani, V. (2016).
\newblock Recycling randomness with structure for sublinear time kernel
  expansions.
\newblock In {\em International Conference on Machine Learning}.

\bibitem[Choromanski et~al., 2017]{choromanski2017angular}
Choromanski, K.~M., Rowland, M., and Weller, A. (2017).
\newblock The unreasonable effectiveness of structured random orthogonal
  embeddings.
\newblock In {\em Advances in Neural Information Processing Systems}.

\bibitem[Dai et~al., 2014]{dai2014double}
Dai, B., Xie, B., He, N., Liang, Y., Raj, A., Balcan, M.-F.~F., and Song, L.
  (2014).
\newblock Scalable kernel methods via doubly stochastic gradients.
\newblock In {\em Advances in Neural Information Processing Systems}.

\bibitem[Davison and Hinkley, 1997]{Davison}
Davison, A.~C. and Hinkley, D.~V. (1997).
\newblock {\em Bootstrap Methods and Their Application}.
\newblock Cambridge.

\bibitem[Dua and Graff, 2017]{UCI}
Dua, D. and Graff, C. (2017).
\newblock {UCI} machine learning repository.

\bibitem[Epperly and Tropp, 2022]{epperly2022jackknife}
Epperly, E.~N. and Tropp, J.~A. (2022).
\newblock Jackknife variability estimation for randomized matrix computations.
\newblock {\em arXiv:2207.06342}.

\bibitem[Erichson et~al., 2018]{erichson2018diffusion}
Erichson, N.~B., Mathelin, L., Brunton, S.~L., and Kutz, J.~N. (2018).
\newblock Diffusion maps meet {N}ystr\"{o}m.
\newblock {\em arXiv:1802.08762}.

\bibitem[Giannakis et~al., 2022]{giannakis2021learning}
Giannakis, D., Henriksen, A., Tropp, J.~A., and Ward, R. (2022).
\newblock Learning to forecast dynamical systems from streaming data.
\newblock {\em SIAM Journal on Applied Dynamical Systems}.

\bibitem[Golub and Van~Loan, 2013]{Golub}
Golub, G.~H. and Van~Loan, C.~F. (2013).
\newblock {\em Matrix Computations}.
\newblock JHU Press.

\bibitem[Gretton et~al., 2012]{Gretton2012}
Gretton, A., Borgwardt, K.~M., Rasch, M.~J., Sch{\"o}lkopf, B., and Smola, A.
  (2012).
\newblock A kernel two-sample test.
\newblock {\em Journal of Machine Learning Research}, 13(1):723--773.

\bibitem[Halko et~al., 2011]{Halko:2011}
Halko, N., Martinsson, P.-G., and Tropp, J.~A. (2011).
\newblock Finding structure with randomness: Probabilistic algorithms for
  constructing approximate matrix decompositions.
\newblock {\em SIAM review}, 53(2):217--288.

\bibitem[Hall, 2013]{HallBootstrap}
Hall, P. (2013).
\newblock {\em The Bootstrap and Edgeworth Expansion}.
\newblock Springer.

\bibitem[Kiessling et~al., 2021]{kiessling2021wind}
Kiessling, J., Str{\"o}m, E., and Tempone, R. (2021).
\newblock Wind field reconstruction with adaptive random {F}ourier features.
\newblock {\em Proceedings of the Royal Society A}, 477(2255):20210236.

\bibitem[Le et~al., 2013]{Le2013}
Le, Q., Sarl{\'o}s, T., and Smola, A. (2013).
\newblock Fastfood - {A}pproximating kernel expansions in loglinear time.
\newblock In {\em International Conference on Machine Learning}.

\bibitem[LeCun et~al., 1998]{lecun1998gradient}
LeCun, Y., Bottou, L., Bengio, Y., and Haffner, P. (1998).
\newblock Gradient-based learning applied to document recognition.
\newblock {\em Proceedings of the IEEE}, 86(11):2278--2324.

\bibitem[Li and Li, 2021]{li2021quantization}
Li, X. and Li, P. (2021).
\newblock Quantization algorithms for random {F}ourier features.
\newblock In {\em International Conference on Machine Learning}.

\bibitem[Li et~al., 2019]{li2019towards}
Li, Z., Ton, J.-F., Oglic, D., and Sejdinovic, D. (2019).
\newblock Towards a unified analysis of random {F}ourier features.
\newblock In {\em International Conference on Machine Learning}, pages
  3905--3914. PMLR.

\bibitem[Liberty et~al., 2007]{liberty2007}
Liberty, E., Woolfe, F., Martinsson, P.-G., Rokhlin, V., and Tygert, M. (2007).
\newblock Randomized algorithms for the low-rank approximation of matrices.
\newblock {\em Proceedings of the National Academy of Sciences},
  104(51):20167--20172.

\bibitem[Liu et~al., 2021]{Liu2020}
Liu, F., Huang, X., Chen, Y., and Suykens, J.~A. (2021).
\newblock Random features for kernel approximation: A survey on algorithms,
  theory, and beyond.
\newblock {\em IEEE Transactions on Pattern Analysis and Machine Intelligence},
  44(10):7128--7148.

\bibitem[Lopes, 2022]{lopes2022aos}
Lopes, M.~E. (2022).
\newblock Central limit theorem and bootstrap approximation in high dimensions:
  Near $1/\sqrt{n}$ rates via implicit smoothing.
\newblock {\em The Annals of Statistics}, 50(5):2492--2513.

\bibitem[Lopes et~al., 2020]{Lopes2020svd}
Lopes, M.~E., Erichson, N.~B., and Mahoney, M.~W. (2020).
\newblock Error estimation for sketched {SVD} via the bootstrap.
\newblock In {\em International Conference on Machine Learning}.

\bibitem[Lopes et~al., 2023]{lopes2023Bernoulli}
Lopes, M.~E., Erichson, N.~B., and Mahoney, M.~W. (2023).
\newblock Bootstrapping the operator norm in high dimensions: {E}rror
  estimation for covariance matrices and sketching.
\newblock {\em Bernoulli}, 29(1):428--450.

\bibitem[Lopes et~al., 2018]{Lopes2018}
Lopes, M.~E., Wang, S., and Mahoney, M.~W. (2018).
\newblock Error estimation for randomized least-squares algorithms via the
  bootstrap.
\newblock In {\em International Conference on Machine Learning}.

\bibitem[Lopes et~al., 2019]{lopes2019JMLR}
Lopes, M.~E., Wang, S., and Mahoney, M.~W. (2019).
\newblock A bootstrap method for error estimation in randomized matrix
  multiplication.
\newblock {\em The Journal of Machine Learning Research}, 20(1):1434--1473.

\bibitem[Lopez-Paz et~al., 2014]{LopezPaz}
Lopez-Paz, D., Sra, S., Smola, A., Ghahramani, Z., and Sch{\"o}lkopf, B.
  (2014).
\newblock Randomized nonlinear component analysis.
\newblock In {\em International Conference on Machine Learning}.

\bibitem[Lorenz, 1963]{lorenz1963deterministic}
Lorenz, E.~N. (1963).
\newblock Deterministic nonperiodic flow.
\newblock {\em Journal of Atmospheric Sciences}, 20(2):130--141.

\bibitem[Lunde et~al., 2021]{Lunde2021}
Lunde, R., Sarkar, P., and Ward, R. (2021).
\newblock Bootstrapping the error of {O}ja's algorithm.
\newblock In {\em Advances in Neural Information Processing Systems}.

\bibitem[Lyu, 2017]{lyu2017spherical}
Lyu, Y. (2017).
\newblock Spherical structured feature maps for kernel approximation.
\newblock In {\em International Conference on Machine Learning}, pages
  2256--2264. PMLR.

\bibitem[Marsland, 2011]{marsland2011machine}
Marsland, S. (2011).
\newblock {\em Machine Learning: An Algorithmic Perspective}.
\newblock Chapman and Hall/CRC.

\bibitem[Martinsson and Tropp, 2020]{Tropp2020}
Martinsson, P.-G. and Tropp, J.~A. (2020).
\newblock Randomized numerical linear algebra: {F}oundations and algorithms.
\newblock {\em Acta Numerica}, 29:403--572.

\bibitem[Rahimi and Recht, 2007]{Rahimi2007}
Rahimi, A. and Recht, B. (2007).
\newblock Random features for large-scale kernel machines.
\newblock In {\em Advances in Neural Information Processing Systems}.

\bibitem[Rudi and Rosasco, 2017]{rudi2017generalization}
Rudi, A. and Rosasco, L. (2017).
\newblock Generalization properties of learning with random features.
\newblock In {\em Advances in Neural Information Processing Systems}.

\bibitem[Rudin, 1990]{Rudin}
Rudin, W. (1990).
\newblock {\em Fourier Analysis on Groups}.
\newblock Wiley.

\bibitem[Sch{\"o}lkopf and Smola, 2002]{Scholkopf2002}
Sch{\"o}lkopf, B. and Smola, A.~J. (2002).
\newblock {\em Learning with {K}ernels: {S}upport {V}ector {M}achines,
  {R}egularization, {O}ptimization, and {B}eyond}.
\newblock MIT.

\bibitem[Shao and Tu, 2012]{Shao2012}
Shao, J. and Tu, D. (2012).
\newblock {\em The Jackknife and Bootstrap}.
\newblock Springer.

\bibitem[Shawe-Taylor and Cristianini, 2004]{Shawe2004}
Shawe-Taylor, J. and Cristianini, N. (2004).
\newblock {\em Kernel Methods for Pattern Analysis}.
\newblock Cambridge.

\bibitem[Sriperumbudur and Szab{\'o}, 2015]{Sriperumbudur2015}
Sriperumbudur, B. and Szab{\'o}, Z. (2015).
\newblock Optimal rates for random {F}ourier features.
\newblock In {\em Advances in Neural Information Processing Systems}.

\bibitem[Sun et~al., 2018]{sun2018but}
Sun, Y., Gilbert, A., and Tewari, A. (2018).
\newblock But how does it work in theory? {L}inear {SVM} with random features.
\newblock In {\em Advances in Neural Information Processing Systems}.

\bibitem[Sutherland and Schneider, 2015]{Sutherland2015}
Sutherland, D.~J. and Schneider, J. (2015).
\newblock On the error of random {F}ourier features.
\newblock In {\em Conference on Uncertainty in Artificial Intelligence}.

\bibitem[van~der Vaart, 2000]{Vaart}
van~der Vaart, A.~W. (2000).
\newblock {\em Asymptotic {S}tatistics}.
\newblock Cambridge.

\bibitem[Verf{\"u}rth, 1994]{Verfurth:1994}
Verf{\"u}rth, R. (1994).
\newblock A posteriori error estimation and adaptive mesh-refinement
  techniques.
\newblock {\em Journal of Computational and Applied Mathematics},
  50(1-3):67--83.

\bibitem[Woolfe et~al., 2008]{woolfe2008}
Woolfe, F., Liberty, E., Rokhlin, V., and Tygert, M. (2008).
\newblock A fast randomized algorithm for the approximation of matrices.
\newblock {\em Applied and Computational Harmonic Analysis}, 25(3):335--366.

\bibitem[Yang et~al., 2012]{yang2012nystrom}
Yang, T., Li, Y.-F., Mahdavi, M., Jin, R., and Zhou, Z.-H. (2012).
\newblock Nystr{\"o}m method vs random {F}ourier features: {A} theoretical and
  empirical comparison.
\newblock In {\em Advances in Neural Information Processing Systems}.

\bibitem[Yao et~al., 2023]{junwenrepo}
Yao, J., Erichson, N.~B., and Lopes, M.~E. (2023).
\newblock \url{https://github.com/jwyyy/bootstrappedRFF}.

\bibitem[Zhang et~al., 2019]{zhang2019low}
Zhang, J., May, A., Dao, T., and R{\'e}, C. (2019).
\newblock Low-precision random {F}ourier features for memory-constrained kernel
  approximation.
\newblock In {\em International Conference on Artificial Intelligence and
  Statistics}.

\bibitem[Zhao and Meng, 2015]{zhao2015fastmmd}
Zhao, J. and Meng, D. (2015).
\newblock Fast{MMD}: {E}nsemble of circular discrepancy for efficient
  two-sample test.
\newblock {\em Neural Computation}, 27(6):1345--1372.

\end{thebibliography}

\clearpage

\pagestyle{empty}
\onecolumn
\aistatstitle{Error Estimation for Random Fourier Features\\[0.2cm]Supplementary Material }
\counterwithin{figure}{section}
The supplementary material consists of three appendices. Appendix~\ref{sec:proof} contains the proof of Theorem 1 from the main text. Appendix~\ref{sec:appendix_mmd} presents empirical results on estimating the error of RFF in the context of kernel-based hypothesis testing. Appendix~\ref{sec:add_matrix} is a continuation of Section 5.1 from the main text, and presents empirical results for an additional dataset in the context of kernel matrix approximation.

\appendix
%
%
%
\section{Proof of Theorem~\ref{thm:main}}\label{sec:proof}
We begin by defining several distributions functions that will be needed throughout the proof. For any $t\in\R$, define
\begin{align}
    F_s(t) & \ = \ \P\Big(\|\tK-\K\|_{\infty} \leq t\Big),\\[0.2cm]
    \tilde F_s(t) & \ = \ \P\big(\ve_1^{\star} \leq t\,\big|\,\mathsf{Z}\big),\\[0.2cm]
    \tilde F_{s,N}(t) & \ = \ \frac{1}{N}\sum_{j=1}^N 1\{\ve_j^{\star}\leq t\},
\end{align}
where $\ve_1^{\star},\dots,\ve_N^{\star}$ are generated as in Algorithm 1, and $1\{\cdot\}$ is an indicator function. Note also that $\tilde F_s$ and $\tilde F_{s,N}$ are random functions.

Below, we develop two lemmas showing that these distribution functions are uniformly close with high probability. The uniform approximations are important, because they imply that the quantiles of $\tilde F_{s,N}$ and $F_s$ behave similarly---which is exactly what is needed to prove Theorem~\ref{thm:main}, since the $(1-\alpha)$-quantiles of $\tilde{F}_{s,N}$ and $F_s$ are respectively  $\tve$ and $\vealph$.

\begin{lemma}\label{lem:DKW}
Suppose the conditions of Theorem~\ref{thm:main} hold. Then, there is an absolute constant $c>0$ such that the event
\begin{equation}
    \sup_{t\in\R}\big|\tilde F_{s,N}(t) - \tilde{F}_s(t)\big| \ \leq \ \ts\frac{\sqrt{\log(N)}}{\sqrt N}
\end{equation}
holds with probability at least $1-c/N.$
\end{lemma}

\proof Conditioning on $\mathsf{Z}$, we can view $\tilde{F}_{s,N}$ as the empirical distribution function associated with $N$ i.i.d.~samples drawn from $\tilde F_s$. Consequently, the Dvoretzky-Kiefer-Wolfowitz inequality~\citep[][p.268]{Vaart} gives the following bound for any real number $r$,
\begin{equation}
   \P\Big( \sup_{t\in\R}\big|\tilde F_{s,N}(t) - \tilde{F}_s(t)\big| > \ts\frac{r}{\sqrt N}\,\Big|\, \mathsf{Z}\Big) \ \leq \ 2e^{-2r^2}.
\end{equation}
 Hence, the statement of the lemma follows by taking an expectation over $\mathsf{Z}$ and using the choice $r=\sqrt{\log(N)}$.\qed

~\\
\begin{lemma}\label{lem:boot}
Suppose the conditions of Theorem~\ref{thm:main} hold. Then, there is an absolute constant $c>0$ such that the event
\begin{equation}\label{eqn:lemboot}
    \sup_{t\in\R}\big|\tilde F_s(t) - F_s(t)\big| \ \leq \ \Big(\ts\frac{c \log(2sn)^5}{s}\Big)^{1/4}
\end{equation}
holds with probability at least $1-c(1/s+\sqrt{\log(sn)^3/s})$.
\end{lemma}

\proof For each $i=1,\dots,s$, define a random matrix $\mathsf{Y}(i)\in\R^{n\times n}$ whose $(j,j')$ entry is 
\begin{equation}
    \mathsf{Y}_{jj'}(i) \ = \ Z_i(x_j)Z_i(x_{j'}) \, - \, \E[Z_i(x_j)Z_i(x_{j'})],
\end{equation}
and let $\bar{\mathsf{Y}}=\frac{1}{s}\sum_{i=1}^s \mathsf{Y}(i)$. Since the expectation of $Z_i(x_j)Z_i(x_{j'})$ is equal to $k(x_j,x_j')$, we have
\begin{equation}
    \|\tK-\K\|_{\infty} \ = \max_{1\leq j,j'\leq n}\big|\bar{\mathsf{Y}}_{jj'}\big|.
\end{equation}

Next, let $\mathsf{Y}^{\star}(1),\dots,\mathsf{Y}^{\star}(s)$ be i.i.d.~samples with replacement from $(\mathsf{Y}(1),\dots,\mathsf{Y}(s))$. Based on the definition of the bootstrap sample $\ve_1^{\star}$ in Algorithm 1, it is straightforward to check that it can be expressed as
\begin{equation}
    \ve_1^{\star} \ = \ \max_{1\leq j,j'\leq n}\Big| \ts\frac{1}{s}\sum_{i=1}^s \mathsf{Y}_{jj'}^{\star}(i)-\bar{\mathsf{Y}}_{jj'}\Big|.
\end{equation}

Likewise, the left side of~\eqref{eqn:lemboot} satisfies
\begin{equation*}
\footnotesize
     \sup_{t\in\R}\big|\tilde F_s(t) - F_s(t)\big|  = \sup_{t\in\R}\Bigg|\P\bigg(\max_{1\leq j,j'\leq n}\Big| \ts\frac{1}{s}\sum_{i=1}^s \mathsf{Y}_{jj'}^{\star}(i)-\bar{\mathsf{Y}}_{jj'}\Big| \leq  t\,\bigg|\, \mathsf{Z}\bigg) - \P\bigg(\displaystyle\max_{1\leq j,j'\leq n}\big| \bar{\mathsf{Y}}_{jj'}\big|\leq  t\bigg)\Bigg|.
\end{equation*}
\normalsize
Due to this representation and the fact that the matrices $\mathsf{Y}(1),\dots,\mathsf{Y}(s)$ are i.i.d.,~the statement~\eqref{eqn:lemboot} follows as a consequence of Lemma 4.5 in~\citep{CCKK2022}, provided that we can verify three conditions: Specifically, it is enough to show that there exist absolute constants $c_1,c_2,C>0$ such that the following bounds (i), (ii), and (iii) hold for all $j,j'\in\{1,\dots,n\}$,
\begin{enumerate}
    \item[(i)]  \ \ \ \ \ $\textup{var}(\mathsf{Y}_{jj'}(1)) \ \geq \ c_1$,
    \item[(ii)] \ \ \ \ \  $\E[\mathsf{Y}_{jj'}^4(1)] \ \leq \ C^2 c_2$,
    \item[(iii)] \ \ \ \ \  $\E[\exp(|\mathsf{Y}_{jj'}(1)|/C)] \ \leq \ 2$.
\end{enumerate}

As a first step toward verifying these conditions, note that the bound $|Z_1(x_j)|\leq\sqrt{2}$ holds almost surely for all \smash{$j\in\{1,\dots,n\}$} by construction. This implies $|\mathsf{Y}_{jj'}(1)|\leq 4$ holds almost surely for all $j,j'$, and so the existence of the two absolute constants $c_2,C>0$ satisfying (ii) and (iii) is clear. 

The only remaining item to address is the lower bound in condition (i). For this purpose, we begin by noting that 
\begin{equation}\label{eqn:firststep}
        \begin{split}
        \var(\mathsf{Y}_{jj'}(1)) & \ = \ \var(Z_1(x_j)Z_1(x_{j'})) \\[0.2cm]
        & \ = \ \E\Big[\big(Z_1(x_j)Z_1(x_j')\big)^2\Big] \ - \ k(x_j,x_{j'})^2.
        \end{split}
\end{equation}
To handle the second moment in the last line, observe that the sum-of-angles identity $\cos(a)\cos(b)=\frac{1}{2}\cos(a-b)+\frac{1}{2}\cos(a+b)$ yields
\begin{equation}\label{eqn:secondstep}
    \begin{split}
        \E\Big[\big(Z_1(x_j)Z_1(x_j')\big)^2\Big]  &  \ = \ \E\bigg[\Big(2\cos\big(\langle W_1,x_j\rangle +U_1\big)\cos\big(\langle W_1,x_{j'}\rangle+ U_1\big)\Big)^2\bigg]\\[0.2cm]
        & \ = \E\bigg[\Big(\cos\big(\langle W_1,x_j-x_{j'}\rangle\big) + \cos\big(\langle W_1,x_j+ x_{j'}\rangle+ 2U_1\big)\Big)^2\bigg]\\[0.2cm]
        & \ = \ \textup{I} + \textup{II} + \textup{III},
    \end{split}
\end{equation}
where we let 
\begin{align*}
    \textup{I} & \ = \ \E\Big[\cos\big(\langle W_1,x_j-x_{j'}\rangle\big)^2\Big]\\
    \textup{II} & \ = \ \E\Big[2\cos\big(\langle W_1,x_j-x_{j'}\rangle\big)\cos\big(\langle W_1,x_j+ x_{j'}\rangle+ 2U_1\big)\Big]\\[0.2cm]
    \textup{III} & \ = \  \E\Big[\cos\big(\langle W_1,x_j+ x_{j'}\rangle+ 2U_1\big)^2\Big].
\end{align*}
For the term $\textup{I}$, we apply Jensen's inequality, followed by the formula~\eqref{eqn:Bochner} from Bochner's Theorem to obtain
\begin{equation}
\begin{split}
    \textup{I} & \ \ \geq \ \ \E\big[\!\cos(\langle W_1,x_j-x_{j'}\rangle)\big]^2\\[0.2cm]
    & \ =  \ \  k(x_j,x_{j'})^2.
\end{split}
\end{equation}
Next, the term $\textup{II}$ turns out to vanish. This is because we can apply the sum-of-angles identity again to obtain
\begin{equation}\label{eqn:vanish}
\begin{split}
    \textup{II} & \ = \ \E\big[\cos(\langle W_1,-2x_{j'}\rangle-2U_1)\big] \ + \ \E\big[\cos(\langle W_1,2x_j\rangle+2U_1)\big]\\[0.2cm]
    & \ = \ 0,
\end{split}
\end{equation}
where the last step uses the facts that $W_1$ and $U_1$ are independent and that for any fixed $r\in\R$, we have
$$\E[\cos(r\pm 2U_1)] \ = \ \textup{Re}\Big(\ts\frac{e^{\sqrt{-1}r}}{2\pi}\displaystyle\int_0^{2\pi} e^{\pm \sqrt{-1}(2u)}du\Big) \ = \ 0.$$
Lastly, for the term $\textup{III}$, we apply the sum-of-angles formula with $a=b$ to get
\begin{equation}
    \begin{split}
        \textup{III} & \ = \ \ts\frac{1}{2} \ + \ \ts\frac{1}{2}\E\Big[\cos\big(\langle W_1,2(x_j+ x_{j'})\rangle+ 4U_1\big)\Big]\\[0.2cm]
        & \ = \ \ts\frac{1}{2},
    \end{split}
\end{equation}
where the expectation on the right vanishes due to the same reasoning that was used in~\eqref{eqn:vanish}. Altogether, we see that $\textup{I}+\textup{II}+\textup{III}\geq 1/2+k(x_j,x_{j'})^2$, and combining this with equations~\eqref{eqn:firststep} and~\eqref{eqn:secondstep} gives the lower bound
\begin{equation}
    \var(\mathsf{Y}_{jj'}(1)) \ \geq \ \ts\frac{1}{2}.
\end{equation}
Hence, the condition (i) is satisfied with $c_1=1/2$, which completes the proof.\qed

\textbf{Concluding the proof of Theorem~\ref{thm:main}.} Combining Lemmas~\ref{lem:DKW} and~\ref{lem:boot} with the triangle inequality shows there is an absolute constant $c>0$ such that the bound
    \begin{equation}\label{eqn:almostdone}
    \sup_{t\in\R}\big|\tilde F_{s,N}(t) - F_s(t)\big| \ \leq \ \ts\frac{\sqrt{\log(N)}}{\sqrt N} + \Big(\ts\frac{c \log(2sn)^5}{s}\Big)^{1/4}
\end{equation}
holds with probability at least $1-c(1/s+\sqrt{\log(sn)^3/s}+1/N)$. Due to this uniform approximation, classical arguments can be used to show that the quantiles of $\tilde F_{s,N}$ and $F_s$ behave similarly, implying that the event $\|\tK-\K\|_{\infty}\leq \tve$ holds with probability close to $1-\alpha$.
For example, the arguments in the proof of Theorem 2.5 in \cite{CCKK2022} or the proof of Lemma 10.4 in~\cite{lopes2022aos} can be used to show that~\eqref{eqn:almostdone} implies 
\begin{equation}\label{eqn:finally}
    \bigg|\P\Big(\|\tK-\K\|_{\infty}\leq \tve\Big) \ - \ (1-\alpha)\bigg| \ \leq  \ \ts\frac{c\sqrt{\log(N)}}{\sqrt N} + \Big(\ts\frac{c \log(2sn)^5}{s}\Big)^{1/4}
\end{equation}
for some absolute constant $c>0$. Finally, as $n\to\infty$, the assumptions of Theorem~\ref{thm:main} ensure that the terms on the right side of~\eqref{eqn:finally} approach 0, which completes the proof. \qed

\section{Error estimation for RFF in hypothesis testing } \label{sec:appendix_mmd}

This section looks at using $\tdealph$ to estimate the error arising from RFF in the context of kernel-based hypothesis testing.

\textbf{MMD statistic.} Let $\D_x=\{x_1,\dots,x_n\}$ and $\D_y=\{y_1,\dots,y_n\}$ denote two datasets in $\R^d$, and consider the problem of testing the null hypothesis that both $\D_x$ and $\D_y$ were drawn in an i.i.d.~manner from the same distribution. A well-known approach for solving this problem is based on the notion of Maximum Mean Discrepancy (MMD), which is a statistical distance that can be formulated in terms of kernels~\citep{Gretton2012}.

For a given kernel $k$, an MMD test statistic can be defined as
\begin{equation}
    T \ = \  \frac{1}{n(n-1)}\sum_{i\neq i'}^n k(x_i,x_{i'})-\frac{2}{n^2}\sum_{i,j=1}^n k(x_i,y_j) +\frac{1}{n(n-1)}\sum_{j\neq j'}^n k(y_j,y_{j'}),
\end{equation}
which is referred to as $\text{MMD}_u^2$ in the paper~\citep{Gretton2012}. Alternatively, we may view $T$ as a functional of the kernel, say $T=\psi(k)$.

In order to compute an approximation to $T$ via RFF, one may use a corresponding statistic defined as $\tilde T=\psi(\tilde k)$ with the approximate kernel $\tilde k$. In particular, we have
\begin{equation}\label{eqn:naive}
    \tilde T \ = \  \frac{1}{n(n-1)}\sum_{i\neq i'}^n \tilde k(x_i,x_{i'})-\frac{2}{n^2}\sum_{i,j=1}^n \tilde k(x_i,y_j) +\frac{1}{n(n-1)}\sum_{j\neq j'}^n \tilde k(y_j,y_{j'}).
\end{equation}
It is also worth noting that $\tilde T$ can be obtained in an equivalent but computationally more efficient way. For this purpose, let $\mathsf{z}(\cdot)=\frac{1}{\sqrt s}(Z_1(\cdot),\dots,Z_s(\cdot))$, with the functions $Z_1(\cdot),\dots,Z_s(\cdot)$ defined as in Section~\ref{sec:prelim}, and let
\begin{equation*}
        \bar{\mathsf{z}}_x =\frac{1}{n}\sum_{j=1}^n \mathsf{z}(x_j) \ \ \ \ \text{  and  } \ \ \ \     \bar{\mathsf{z}}_y=\frac{1}{n}\sum_{j=1}^n\mathsf{z}(y_j),
\end{equation*}
which are both vectors in $\R^s$. Then, the statistic $\tilde T$ is expressible as
\begin{equation*}
    \tilde T \ =  \ \ts\frac{n^2}{n^2-n}\Big(\|\bar{\mathsf{z}}_x\|_2^2-\ts\frac{1}{n^2}\sum_{j=1}^n\|\mathsf{z}(x_j)\|_2^2\Big) \  -2\big\langle \bar{\mathsf{z}}_x,\bar{\mathsf{z}}_y\big\rangle \ + \ \frac{n^2}{n^2-n}\Big(\|\bar{\mathsf{z}}_y\|_2^2-\ts\frac{1}{n^2}\sum_{j=1}^n\|\mathsf{z}(y_j)\|_2^2\Big),
\end{equation*}
which has the advantage that it can be computed with a cost that is linear $n$, rather than quadratic in $n$ (as in~\eqref{eqn:naive}).

To assess the error of the RFF approximation using the framework developed in Sections~\ref{sec:intro} and~\ref{sec:method}, we estimate the 90\% and 99\% quantiles $\delta_{0.9}$ and $\delta_{0.99}$ of the error variable $|\psi(\tilde k)-\psi(k)|=|\tilde T-T|$ using Algorithm 1.

\textbf{Data examples.} We constructed three different versions of the pair $(\D_x,\D_y)$. Each version was constructed so that $|\D_x|=|\D_y|=25000$ and $d=10$. The first version of $(\D_x,\D_y)$ was obtained by uniformly subsampling 25000 rows and $10$ columns from the datasets \emph{YearPredictionMSD} (MSD) and \emph{Buzz in social media} (Buzz), and the second version of $(\D_x,\D_y)$ was obtained in the same way from the datasets \emph{SGEMM GPU kernel performance} (GPU) and \emph{Gas Turbine CO and NOx Emission} (Emission). (The four named datasets are available in the repository~\citep{UCI}.) In addition, the third version of $(\D_x,\D_y)$ was constructed with synthetic data by sampling 25000 points from the two multivariate Gaussian distributions $N(0,\frac{1}{10}\cdot \mathsf{I}_{10})$, and $N(0,(\frac{1}{10}+\eta) \cdot \mathsf{I}_{10})$, where $\eta>0$ was chosen small enough so that detecting a difference with $T$ was relatively challenging. More specifically, we selected $\eta=.0933$ so that the p-value derived from $T$ (as in Corollary 11 of~\citep{Gretton2012}) was nearly equal to 5\%.

\begin{figure*}[h]
\centering
\begin{subfigure}{1\textwidth}	
\centering
\DeclareGraphicsExtensions{.pdf}
\begin{overpic}[width=0.31\textwidth]{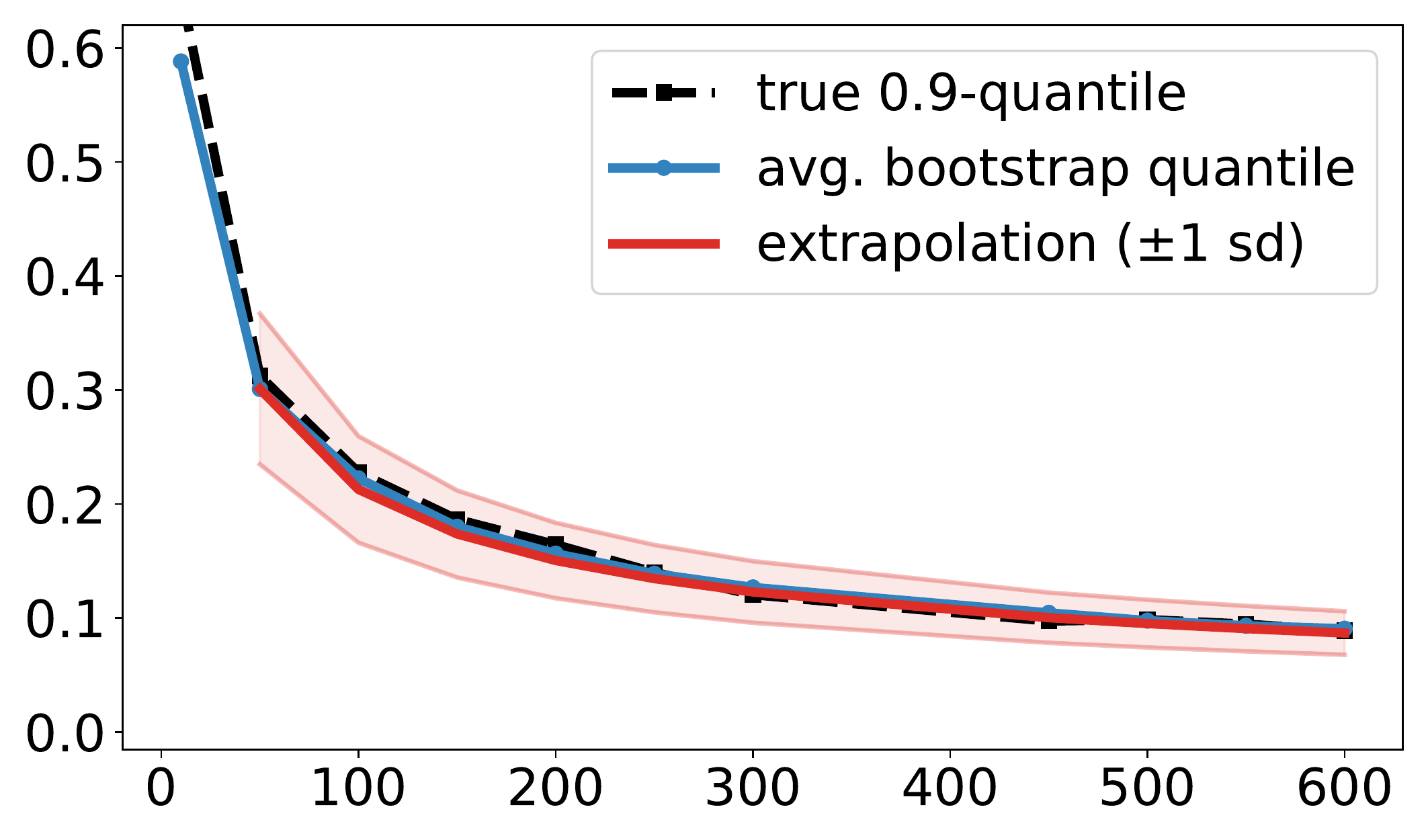} 
    \put(-7,27){\rotatebox{90}{\small $\delta_{0.9}$}}
	\put(33,60){\color{black}{\small Cauchy kernel}} 			
\end{overpic}\hspace*{-0.2cm}
~
\begin{overpic}[width=0.31\textwidth]{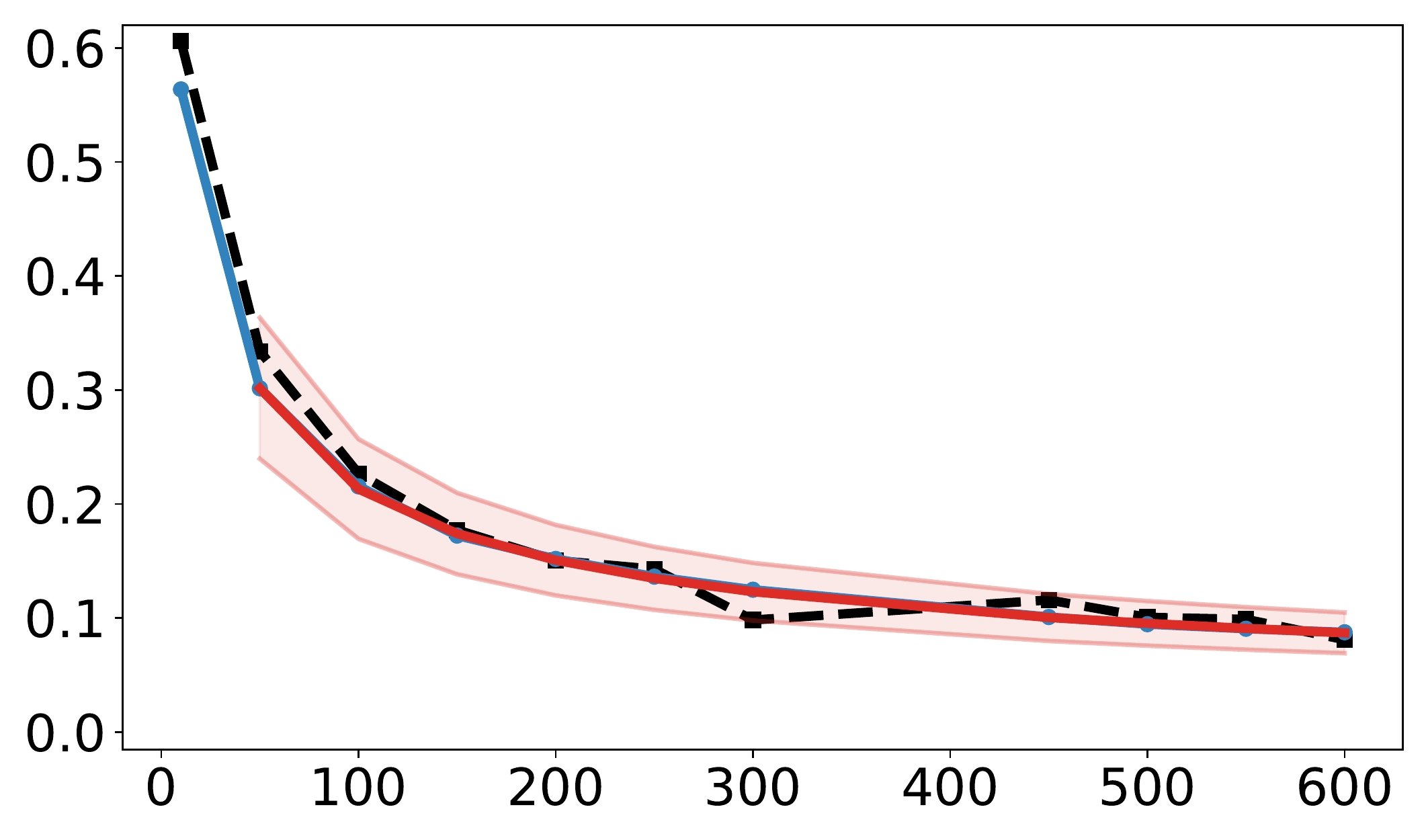} 
	\put(31,60){\color{black}{\small Gaussian kernel}} 			
\end{overpic}
~
\begin{overpic}[width=0.31\textwidth]{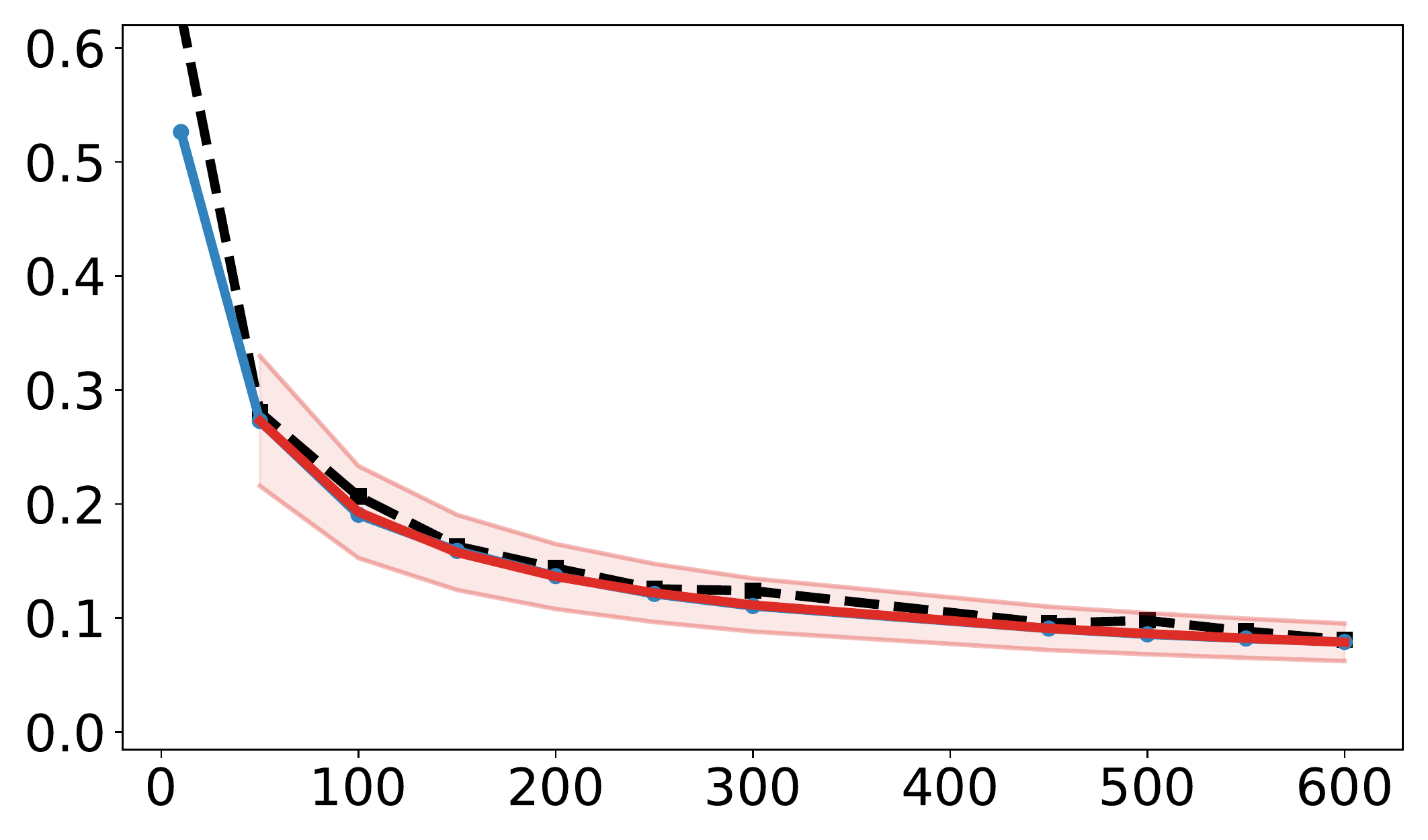} 
	\put(30,60){\color{black}{\small Laplacian kernel}} 
	\put(101,18){\rotatebox{90}{\scriptsize (Buzz-MSD)}}
\end{overpic}\hspace*{-0.2cm}
\end{subfigure}

\begin{subfigure}{1\textwidth}	
\centering
\DeclareGraphicsExtensions{.pdf}

\begin{overpic}[width=0.31\textwidth]{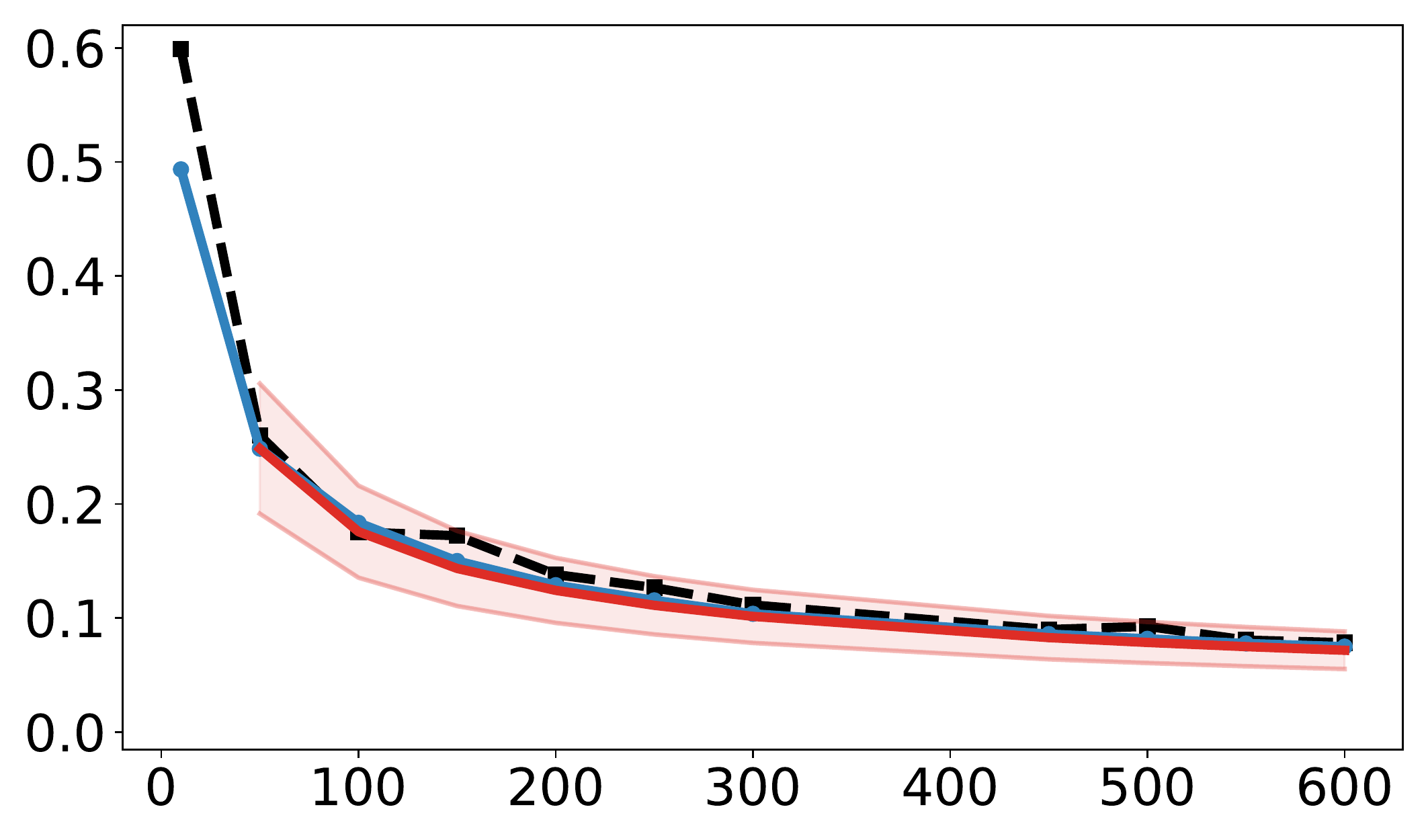} 
	\put(-7,27){\rotatebox{90}{\small  $\delta_{0.9}$}}
				
\end{overpic}\hspace*{-0.2cm}
~
\begin{overpic}[width=0.31\textwidth]{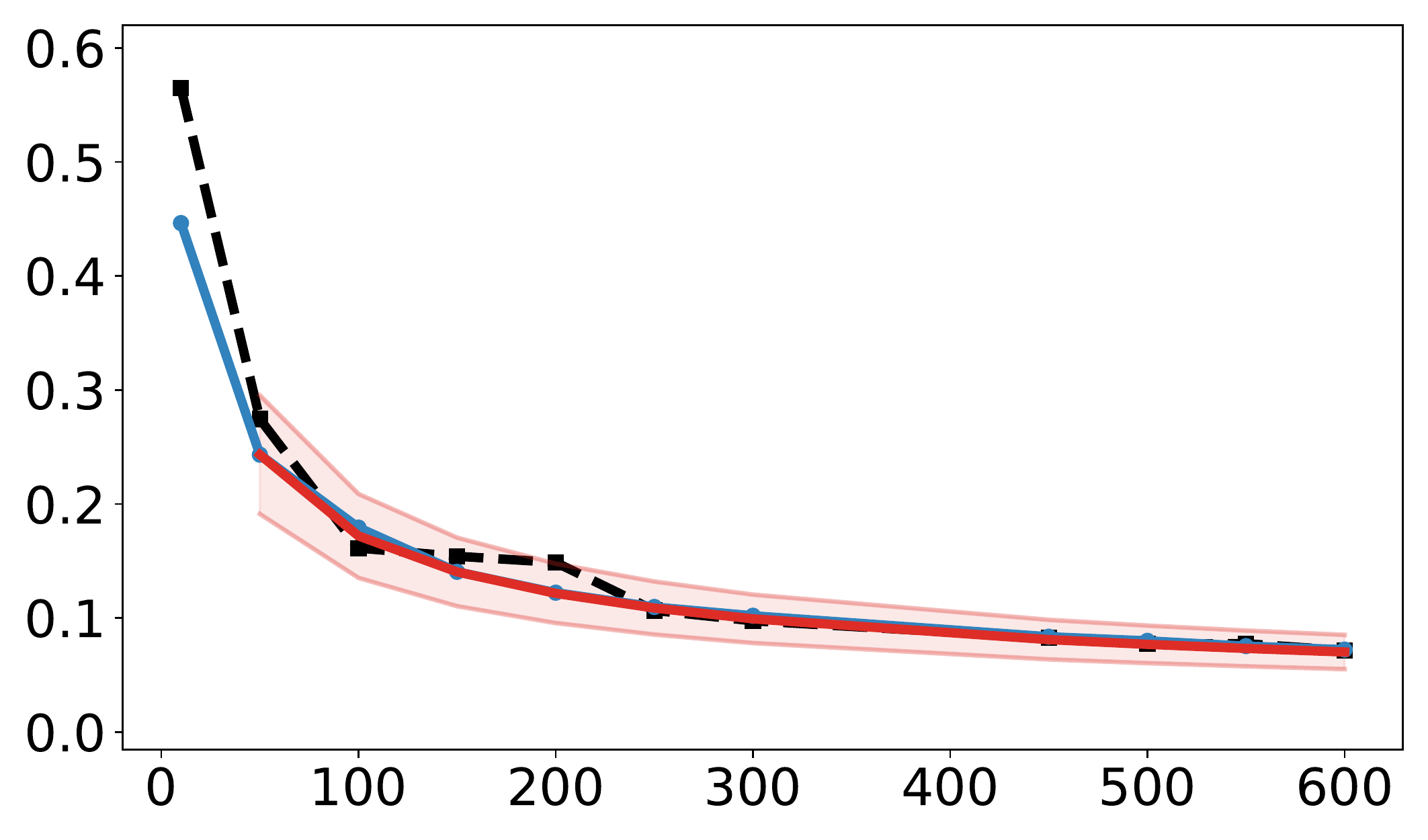} 
\end{overpic}
~
\begin{overpic}[width=0.31\textwidth]{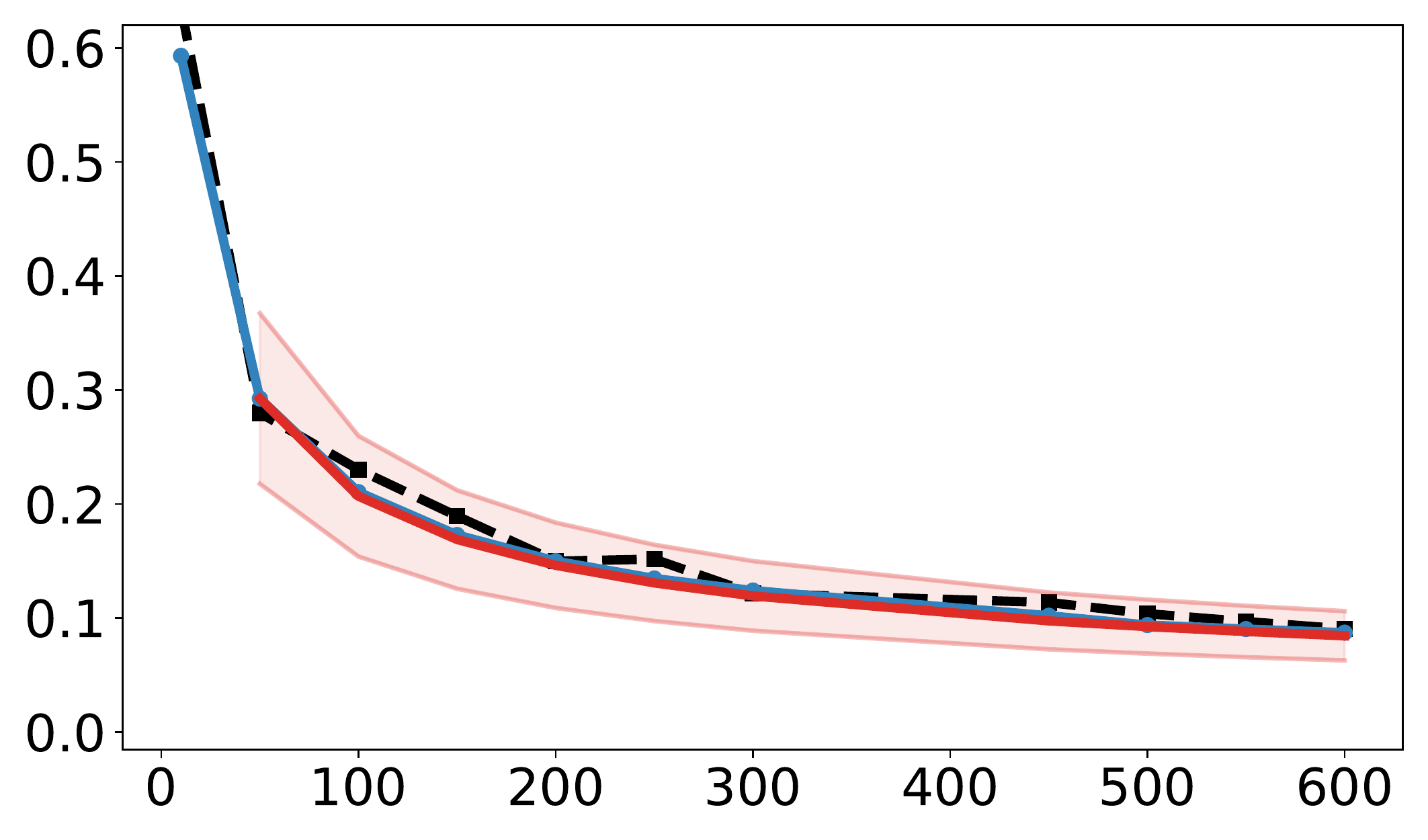} 
	\put(101,15){\rotatebox{90}{\scriptsize (GPU-Emission)}}
\end{overpic}\hspace*{-0.2cm}
	
\end{subfigure}

\begin{subfigure}{1\textwidth}	
\centering
\DeclareGraphicsExtensions{.pdf}

\begin{overpic}[width=0.31\textwidth]{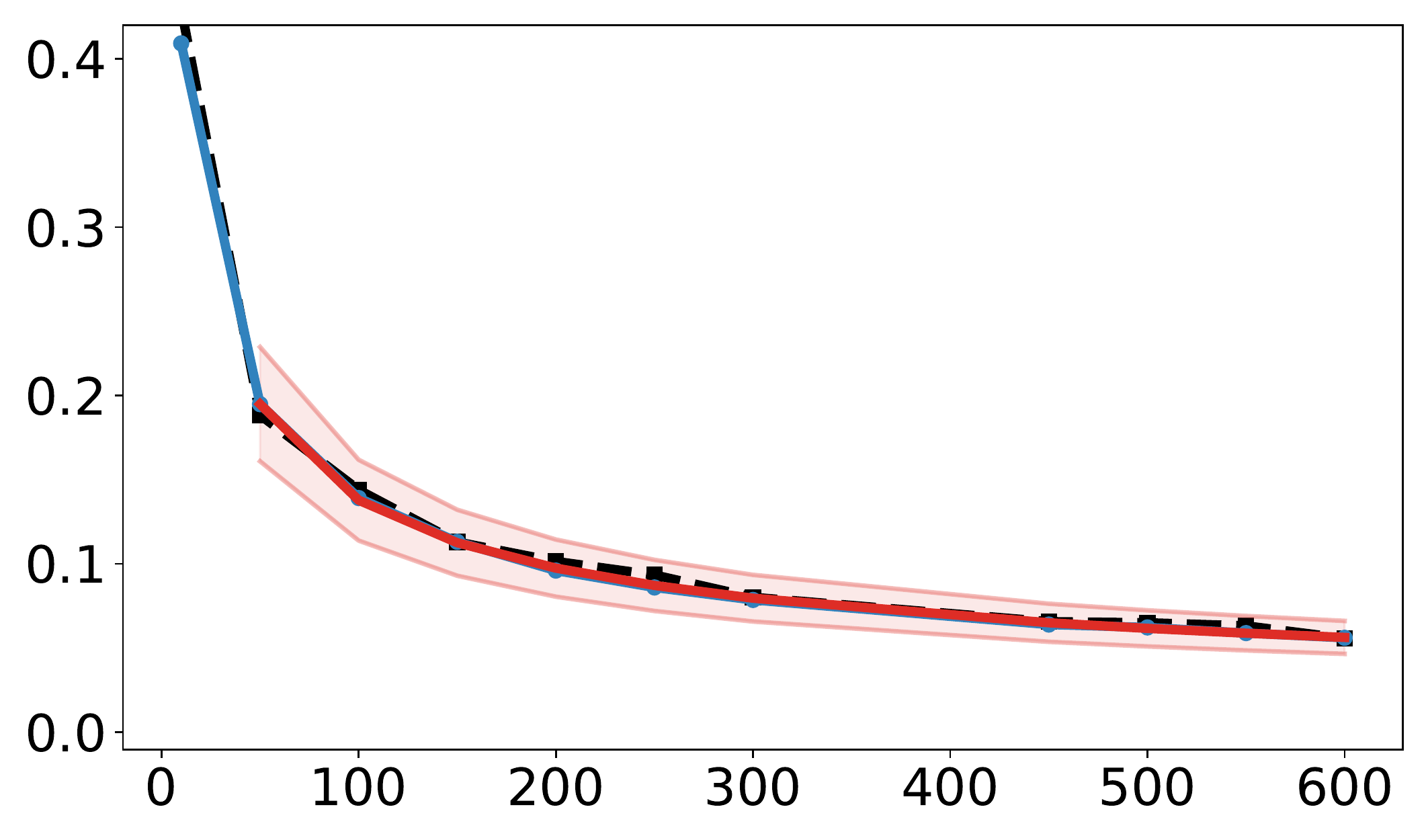} 
	\put(-7,27){\rotatebox{90}{\small  $\delta_{0.9}$}}
	\put(50,-3){\color{black}{\footnotesize $s$}} 
\end{overpic}\hspace*{-0.2cm}
~
\begin{overpic}[width=0.31\textwidth]{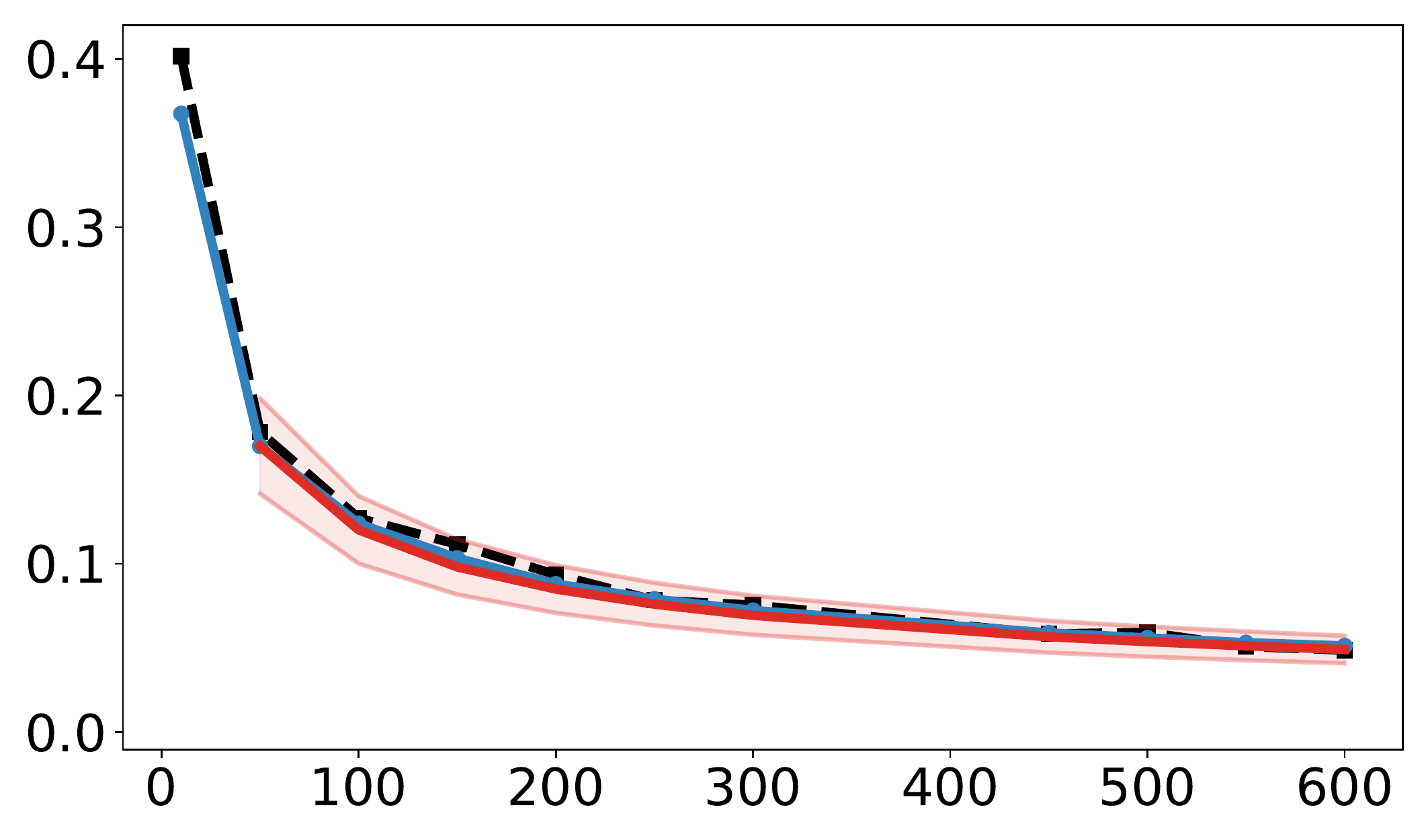} 
	\put(50,-3){\color{black}{\footnotesize $s$}} 
\end{overpic}
~
\begin{overpic}[width=0.31\textwidth]{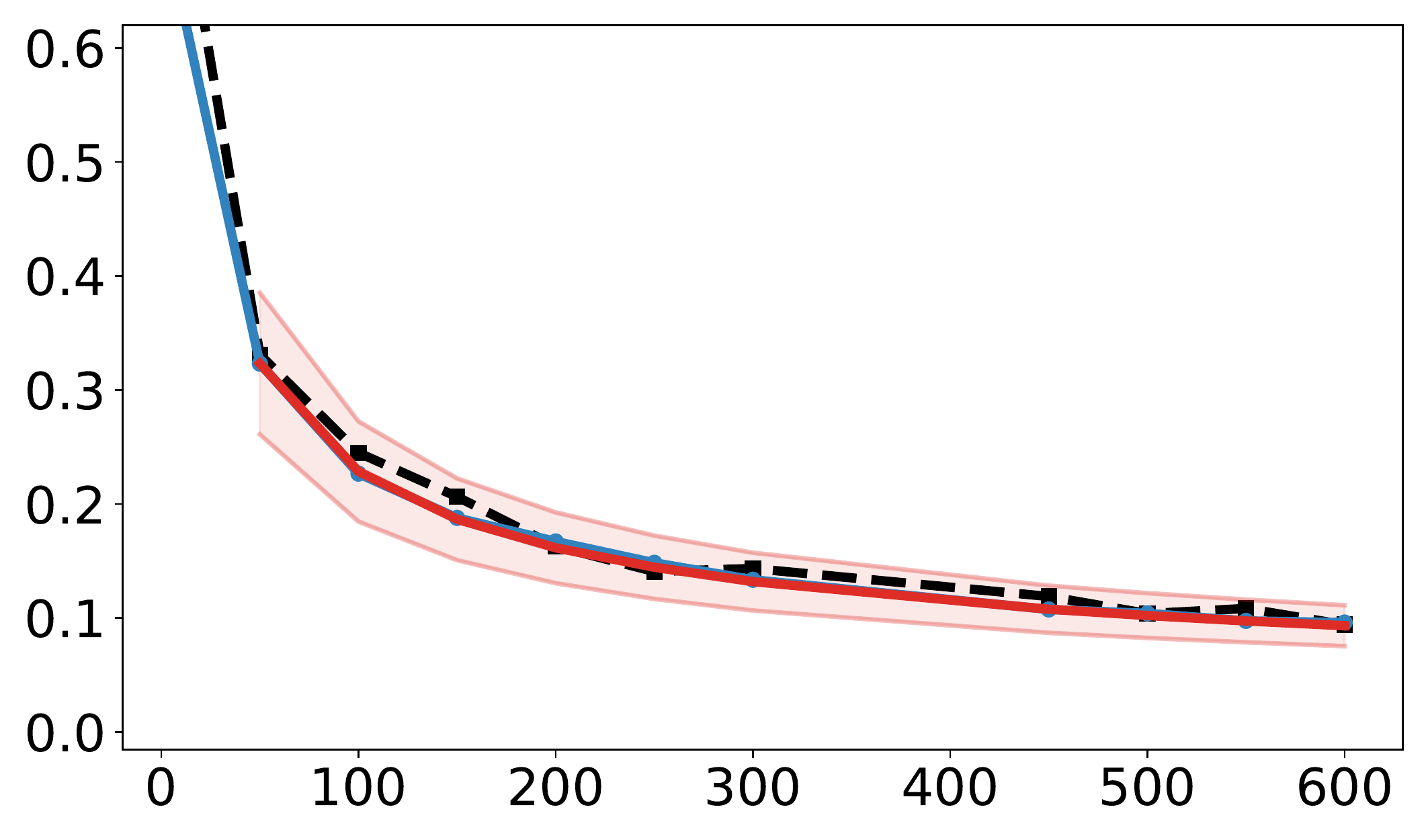} 
	\put(50,-3){\color{black}{\footnotesize $s$}} 
	\put(101,11){\rotatebox{90}{\scriptsize (Synthetic-Synthetic)}}
\end{overpic}\hspace*{-0.2cm}
	
\end{subfigure}

\caption{(Estimation of $\delta_{0.9}$ for $|\psi(\tilde k)-\psi(k)|=|\tilde T-T|$.) The rows correspond to different pairs of datasets, and the columns correspond to different kernels.}
	
\label{fig:results_mmd}
\end{figure*}

\begin{figure*}[h]
\centering
\begin{subfigure}{1\textwidth}	
\centering
\DeclareGraphicsExtensions{.pdf}
\begin{overpic}[width=0.31\textwidth]{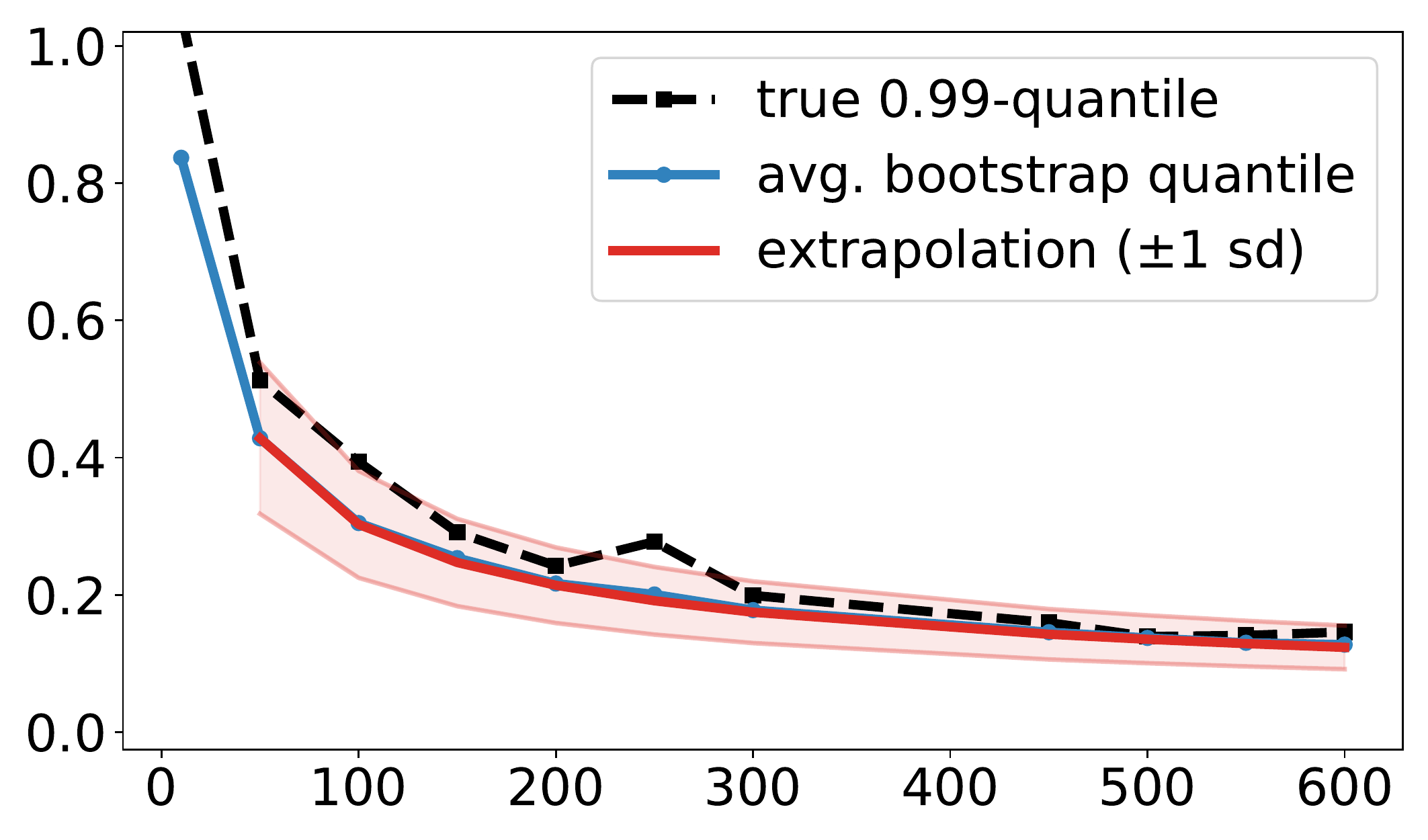} 
    \put(-7,27){\rotatebox{90}{\small $\delta_{0.99}$}}
	\put(33,60){\color{black}{\small Cauchy kernel}} 			
\end{overpic}\hspace*{-0.2cm}
~
\begin{overpic}[width=0.31\textwidth]{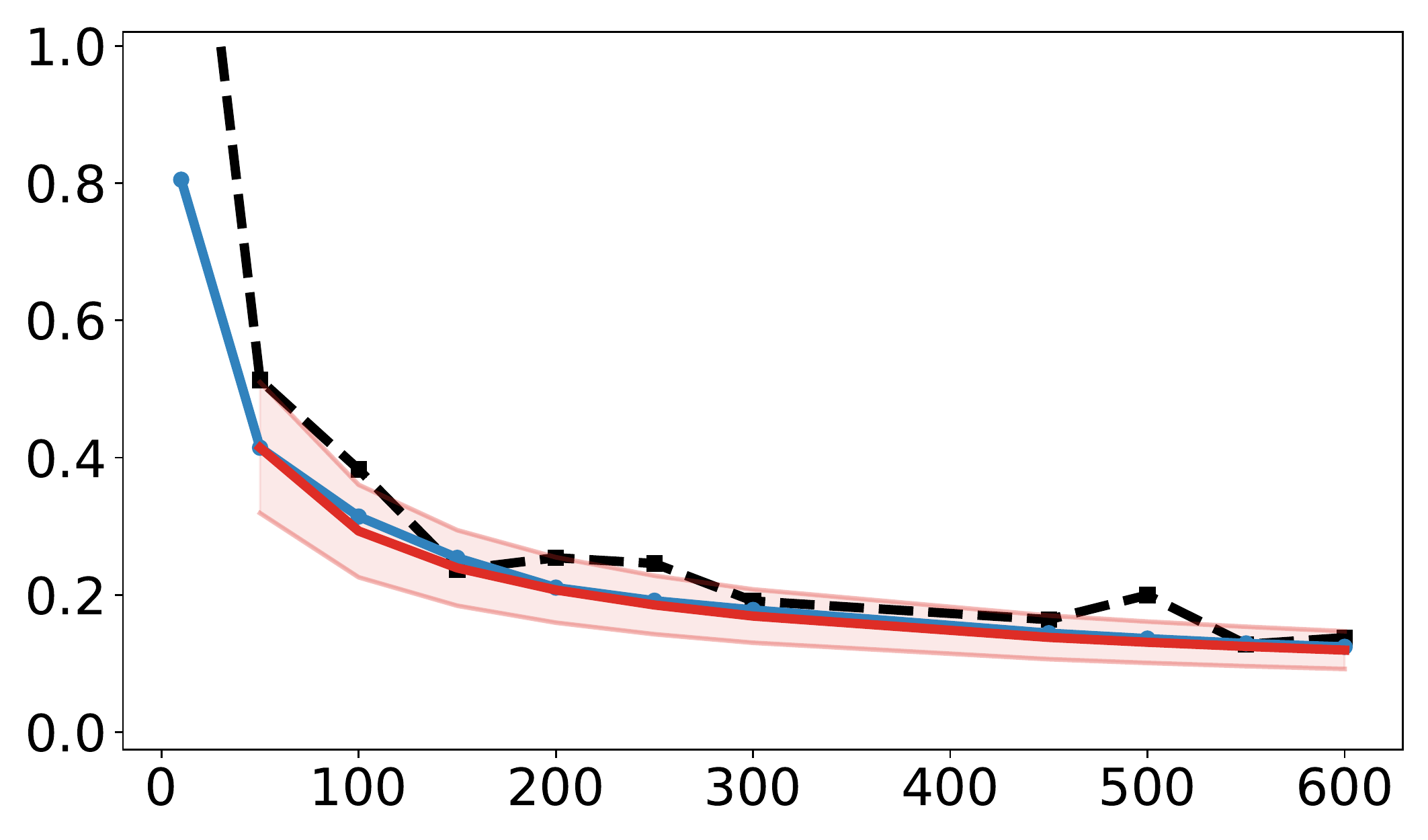} 
	\put(31,60){\color{black}{\small Gaussian kernel}} 			
\end{overpic}
~
\begin{overpic}[width=0.31\textwidth]{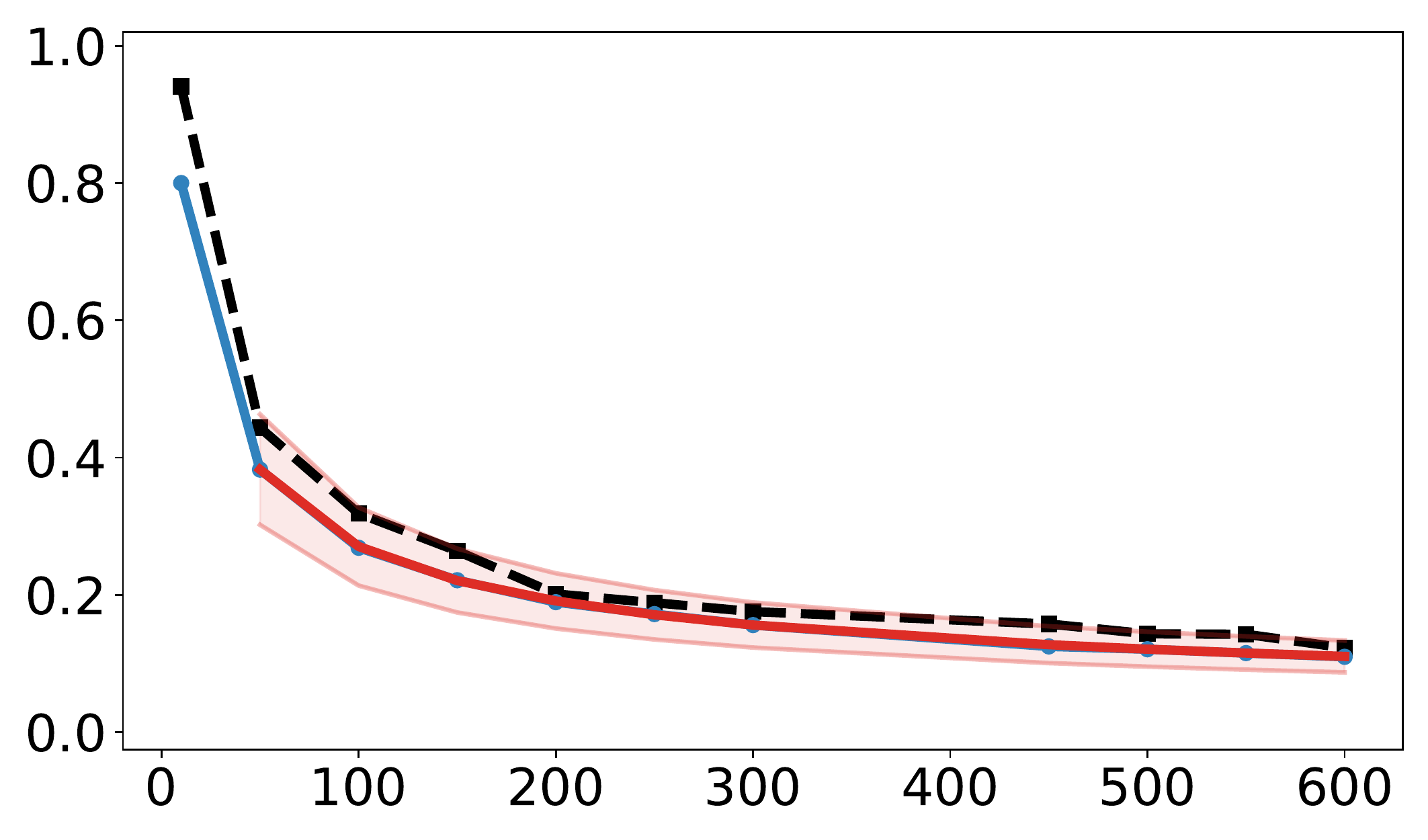} 
	\put(30,60){\color{black}{\small Laplacian kernel}} 
	\put(101,18){\rotatebox{90}{\scriptsize (Buzz-MSD)}}
\end{overpic}\hspace*{-0.2cm}
\end{subfigure}

\begin{subfigure}{1\textwidth}	
\centering
\DeclareGraphicsExtensions{.pdf}

\begin{overpic}[width=0.31\textwidth]{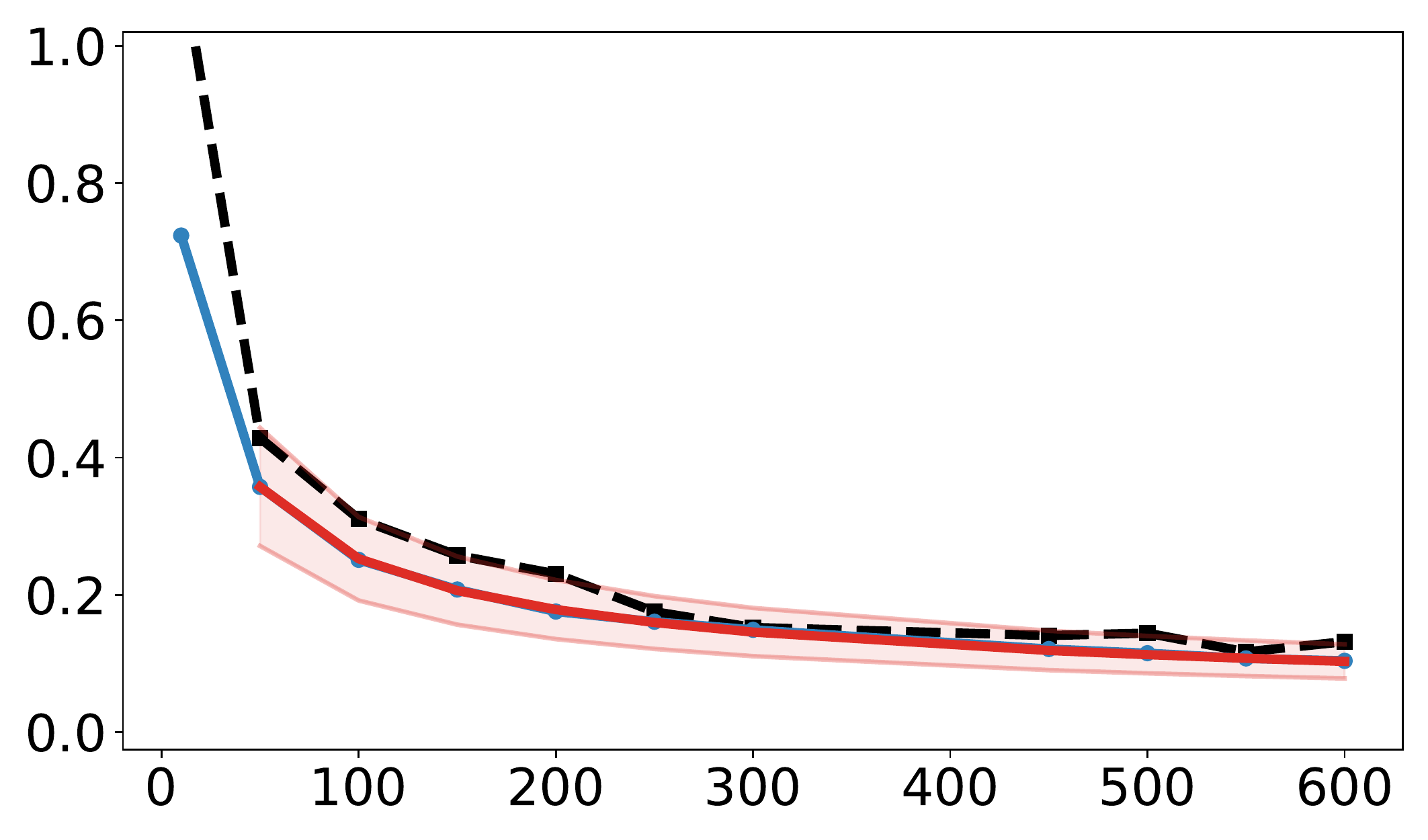} 
	\put(-7,27){\rotatebox{90}{\small  $\delta_{0.99}$}}
				
\end{overpic}\hspace*{-0.2cm}
~
\begin{overpic}[width=0.31\textwidth]{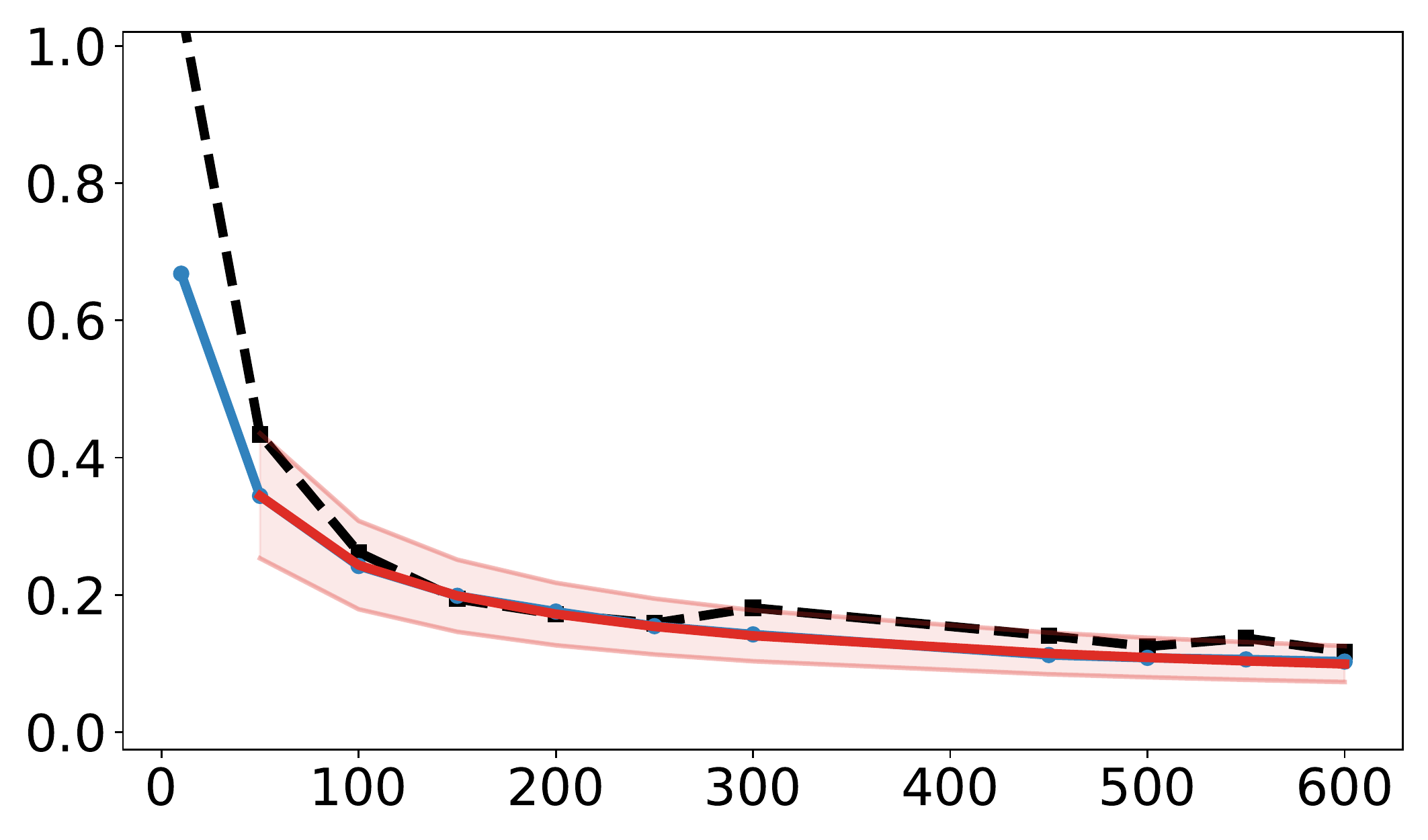} 
\end{overpic}
~
\begin{overpic}[width=0.31\textwidth]{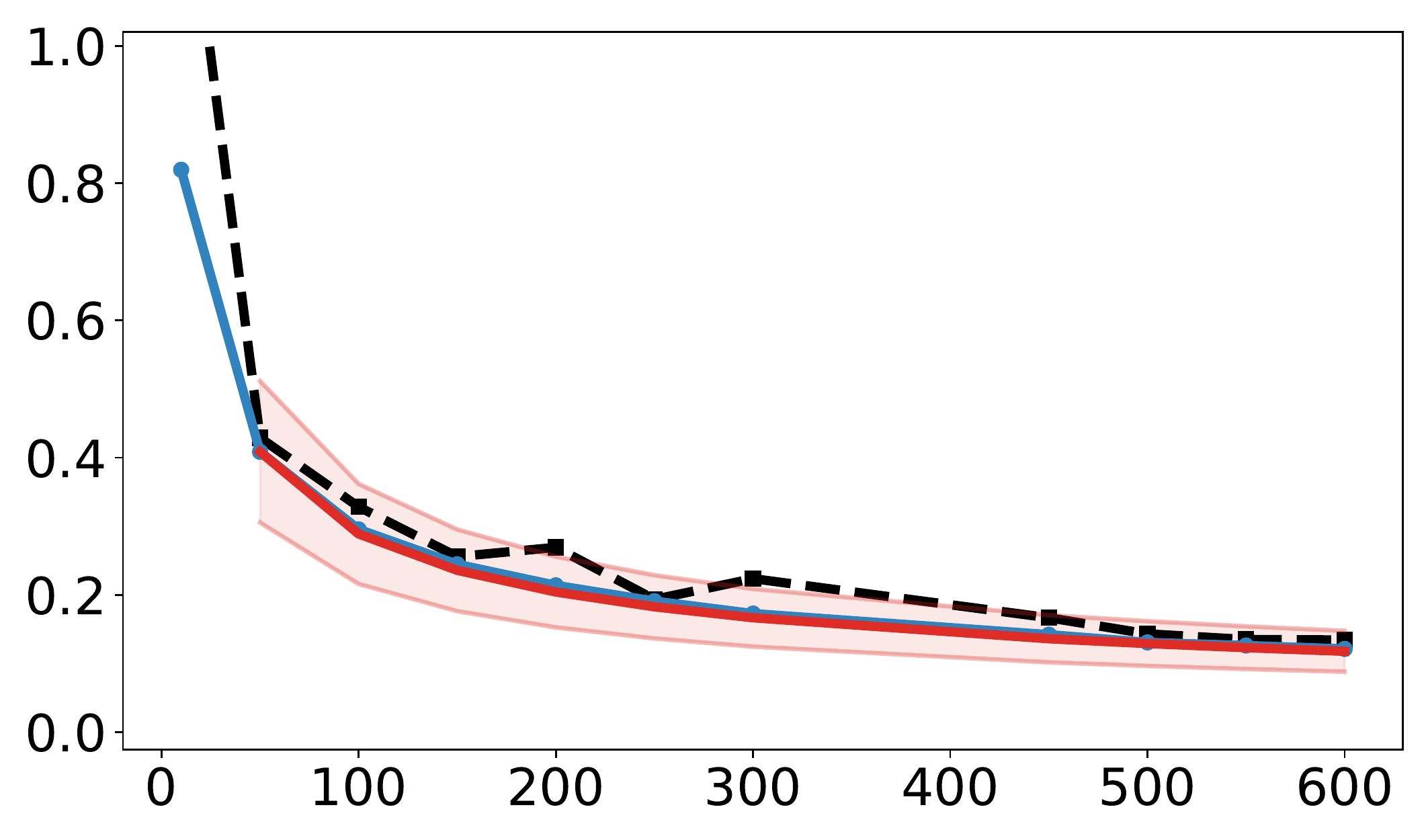} 
	\put(101,15){\rotatebox{90}{\scriptsize (GPU-Emission)}}
\end{overpic}\hspace*{-0.2cm}
	
\end{subfigure}

\begin{subfigure}{1\textwidth}	
\centering
\DeclareGraphicsExtensions{.pdf}

\begin{overpic}[width=0.31\textwidth]{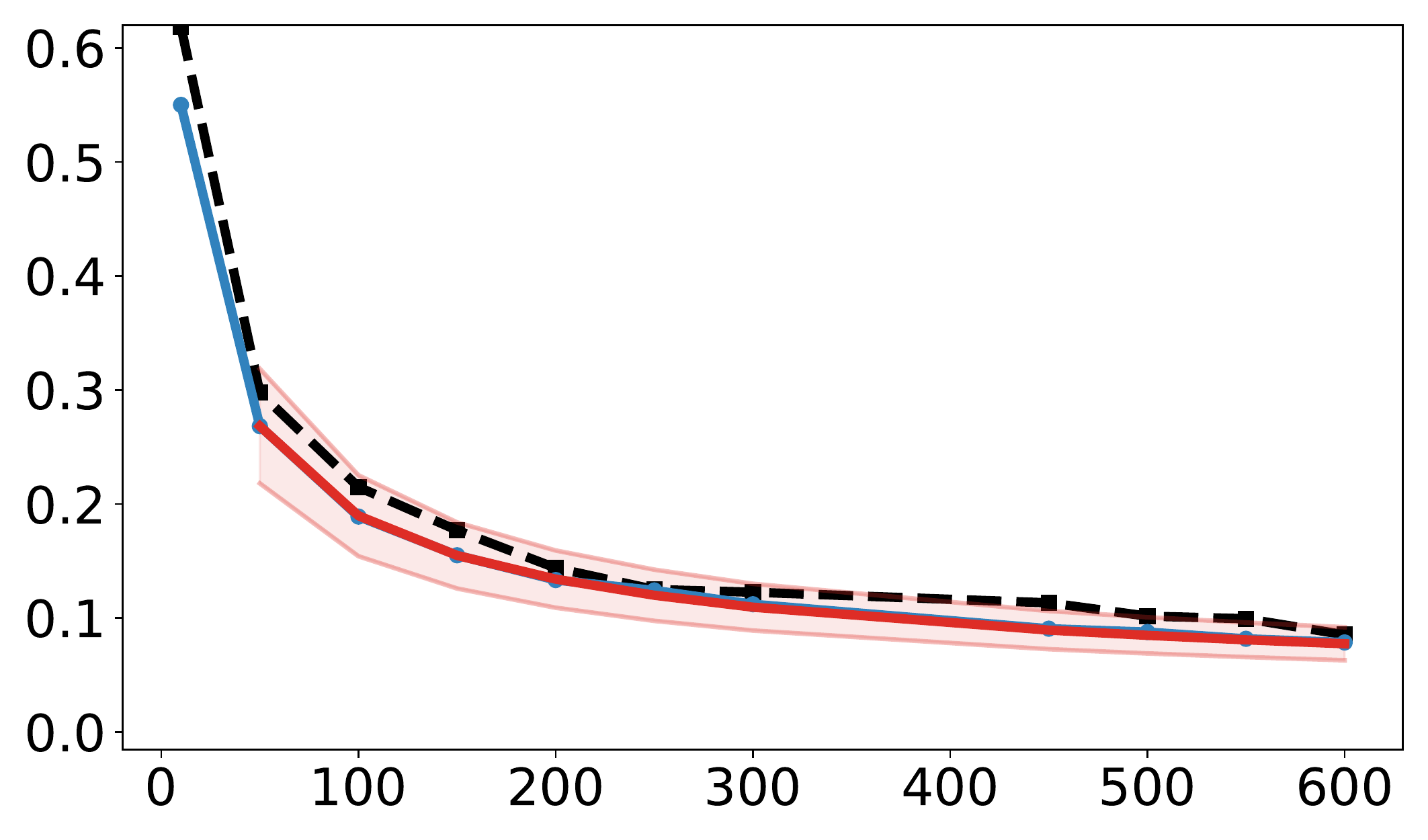} 
	\put(-7,27){\rotatebox{90}{\small  $\delta_{0.99}$}}
	\put(50,-3){\color{black}{\footnotesize $s$}} 
\end{overpic}\hspace*{-0.2cm}
~
\begin{overpic}[width=0.31\textwidth]{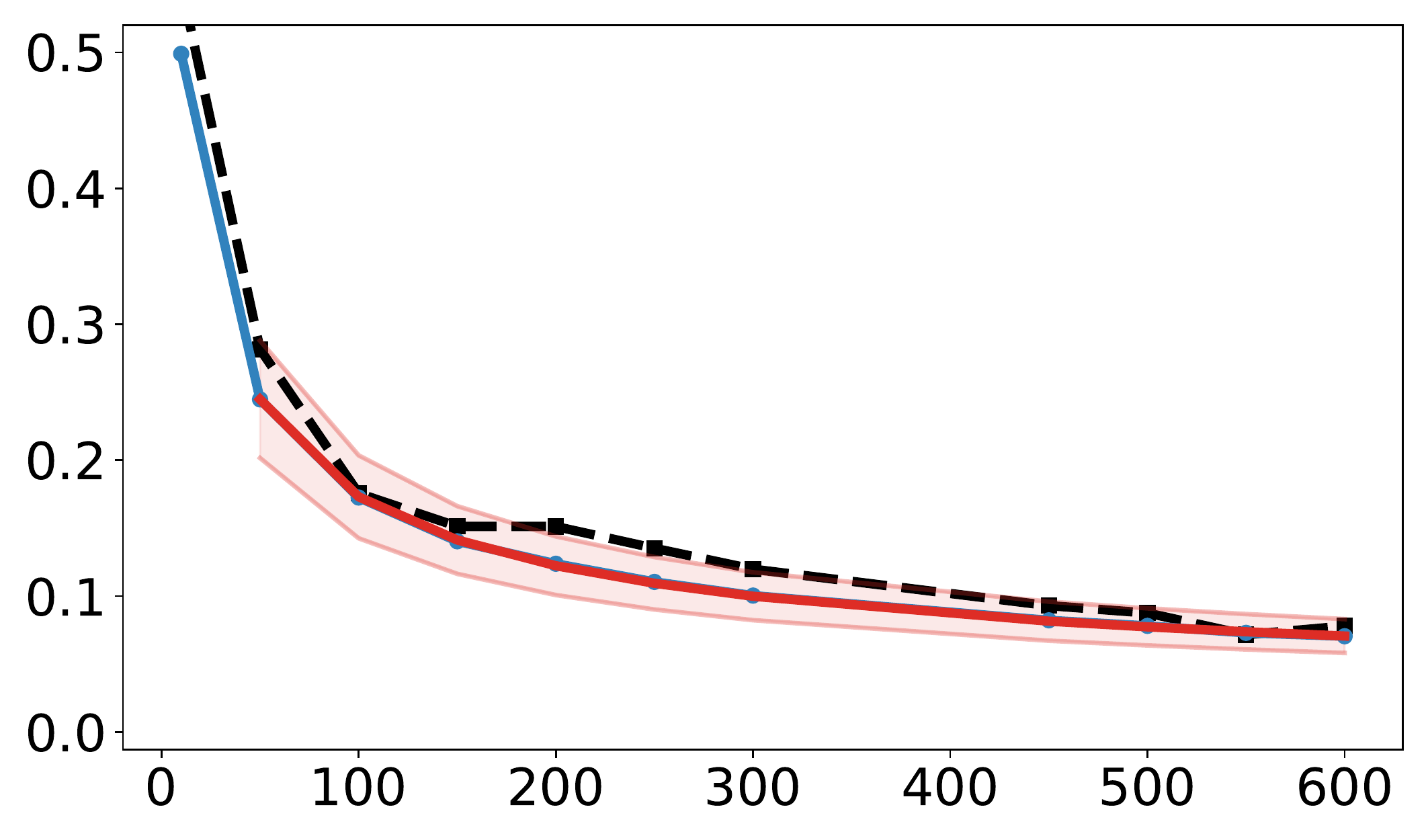} 
	\put(50,-3){\color{black}{\footnotesize $s$}} 
\end{overpic}
~
\begin{overpic}[width=0.31\textwidth]{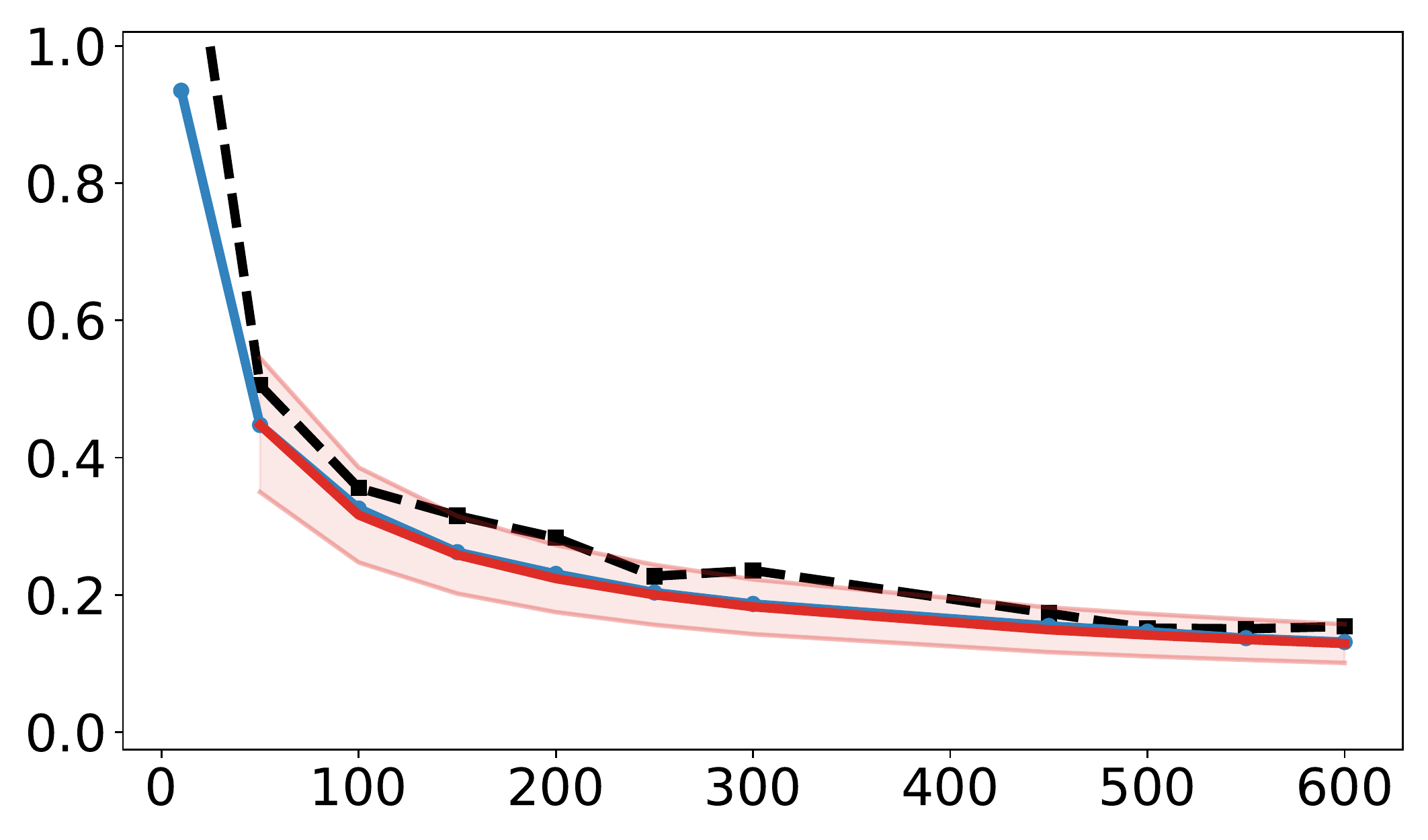} 
	\put(50,-3){\color{black}{\footnotesize $s$}} 
	\put(101,11){\rotatebox{90}{\scriptsize (Synthetic-Synthetic)}}
\end{overpic}\hspace*{-0.2cm}
	
\end{subfigure}

\caption{(Estimation of $\delta_{0.99}$ for $|\psi(\tilde k)-\psi(k)|=|\tilde T-T|$.) The rows correspond to different pairs of datasets, and the columns correspond to different kernels.}  
	
\label{fig:results_mmd_99}
\end{figure*}
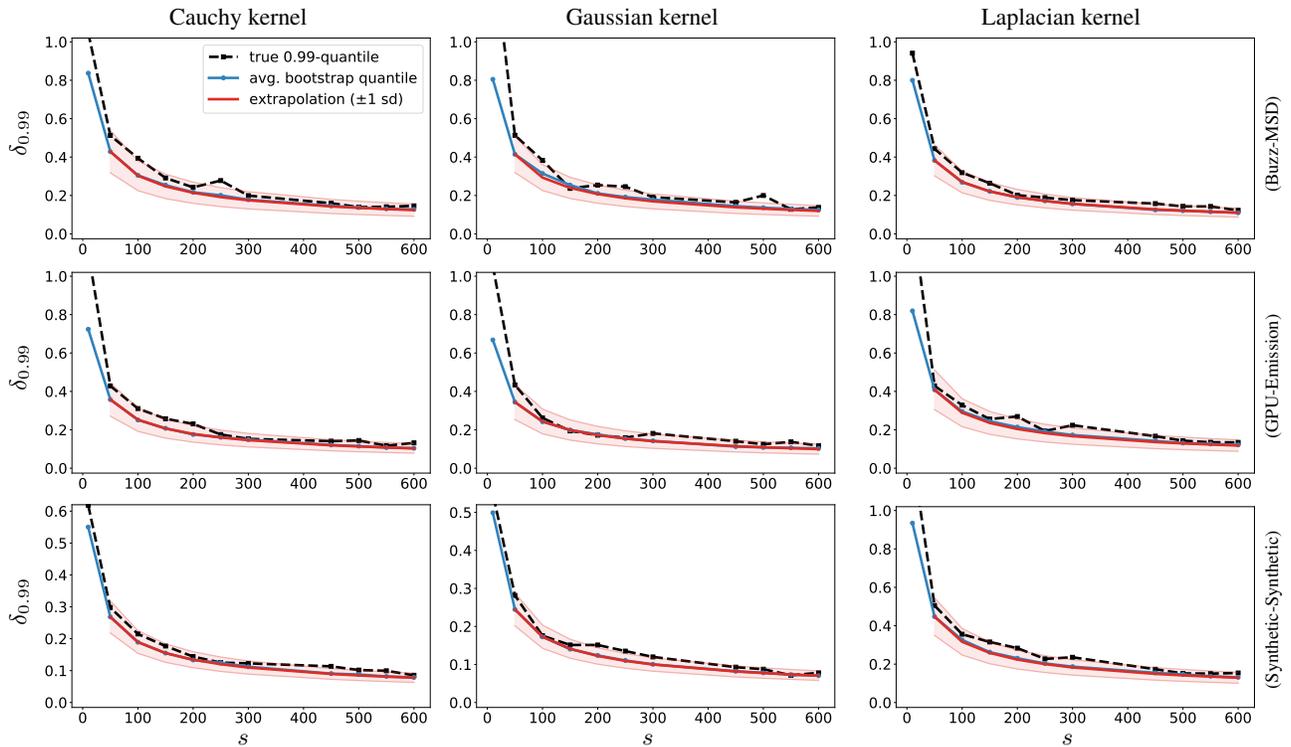

\textbf{Design of experiments.}
Our experiments in this section were organized analogously to those in Section~\ref{sec:ridge}. In particular, for a grid of $s$ values ranging from 30 to 600, we generated 300 realizations of the approximate kernel $\tilde k$, and we applied Algorithm 1 to each such realization with $N=30$ bootstrap iterations. The results for these experiments are displayed in Figures~\ref{fig:results_mmd} and~\ref{fig:results_mmd_99}, where the three colored curves for $(\delta_{\,0.9}, \tilde{\delta}_{0.9}, \tilde{\delta}_{\,0.9}^{\,\textsc{ext}})$ and $(\delta_{\,0.99}, \tilde{\delta}_{0.99}, \tilde{\delta}_{\,0.99}^{\,\textsc{ext}})$ have the same interpretations as the corresponding curves in Figure~\ref{fig:results_kernel_ridge_regrssion}. In the current context, all the curves were multiplied by the relevant value of $1/T$, so that they can be viewed on a more natural scale. Also, the curves for the extrapolated estimates $\tilde{\delta}_{\,0.9}^{\,\textsc{ext}}$ and $\tilde{\delta}_{\,0.99}^{\,\textsc{ext}}$ are based on a starting point of $s_0=50$. Lastly, the experiments were performed with three different kernels: the Gaussian kernel $\exp(-\|x-x'\|_2^2/2)$, the Laplacian kernel $\exp(-\|x-x'\|_1/2)$, and the Cauchy kernel $\prod_{j=1}^{10} 1/(1 + \Delta_j^2/2)$ where $\Delta = x - x'$. 

\textbf{Discussion of results.} Figure~\ref{fig:results_mmd} shows that the estimates  $\tilde{\delta}_{1-\alpha}$ and $\tilde{\delta}_{1-\alpha}^{\,\textsc{ext}}$ agree well with $\delta_{1-\alpha}$ across different choices of kernels and datasets when $1-\alpha=90\%$. The same pattern also appears in Figure~\ref{fig:results_mmd_99} for the case when $1-\alpha=99\%$, which is especially encouraging because the choice of $1-\alpha=99\%$ makes the estimation problem more challenging. Furthermore, it is notable that the same inexpensive choice $s_0=50$ leads to high-quality extrapolations for both choices of $\alpha$.

\vspace{0.5cm}

\vspace{0.5cm}

\clearpage

\section{Additional results on error estimation for RFF in kernel matrix approximation}\label{sec:add_matrix}
This appendix is a continuation of Section 5.1 from the main text, in which we present additional results for data that reside on the well-known 3-dimensional ``Swiss roll'' structure. Specifically, we used code provided by~\cite{marsland2011machine} to generate \smash{$n=20000$} data points. Apart from the choice of the dataset, the experiments here followed the same design and settings as in Section 5.1.

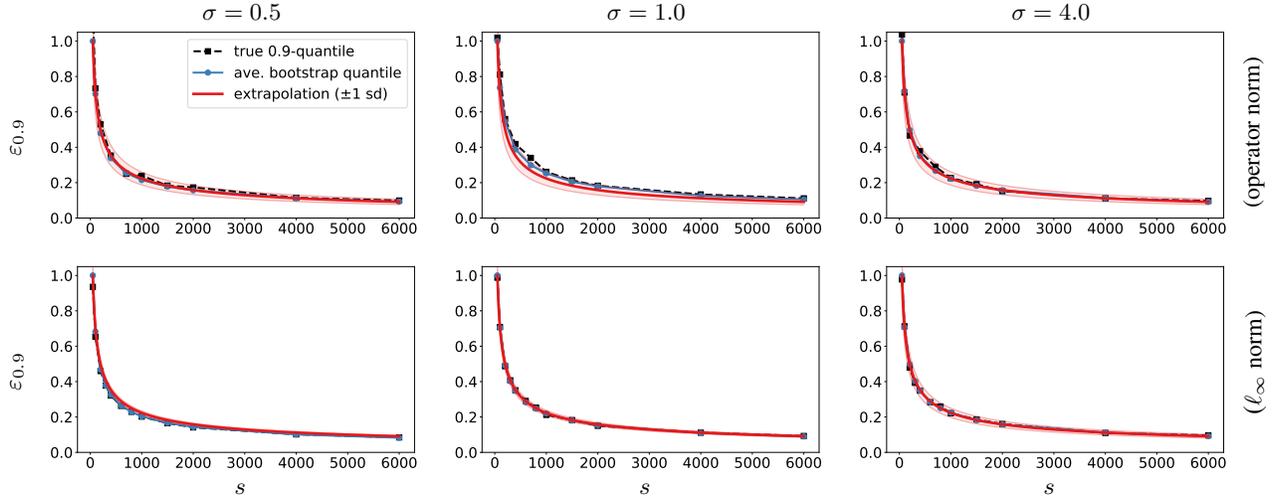
\begin{figure*}[h]
	\centering
	\begin{subfigure}{1\textwidth}	
		\centering
		\DeclareGraphicsExtensions{.pdf}
		\begin{overpic}[width=0.31\textwidth]{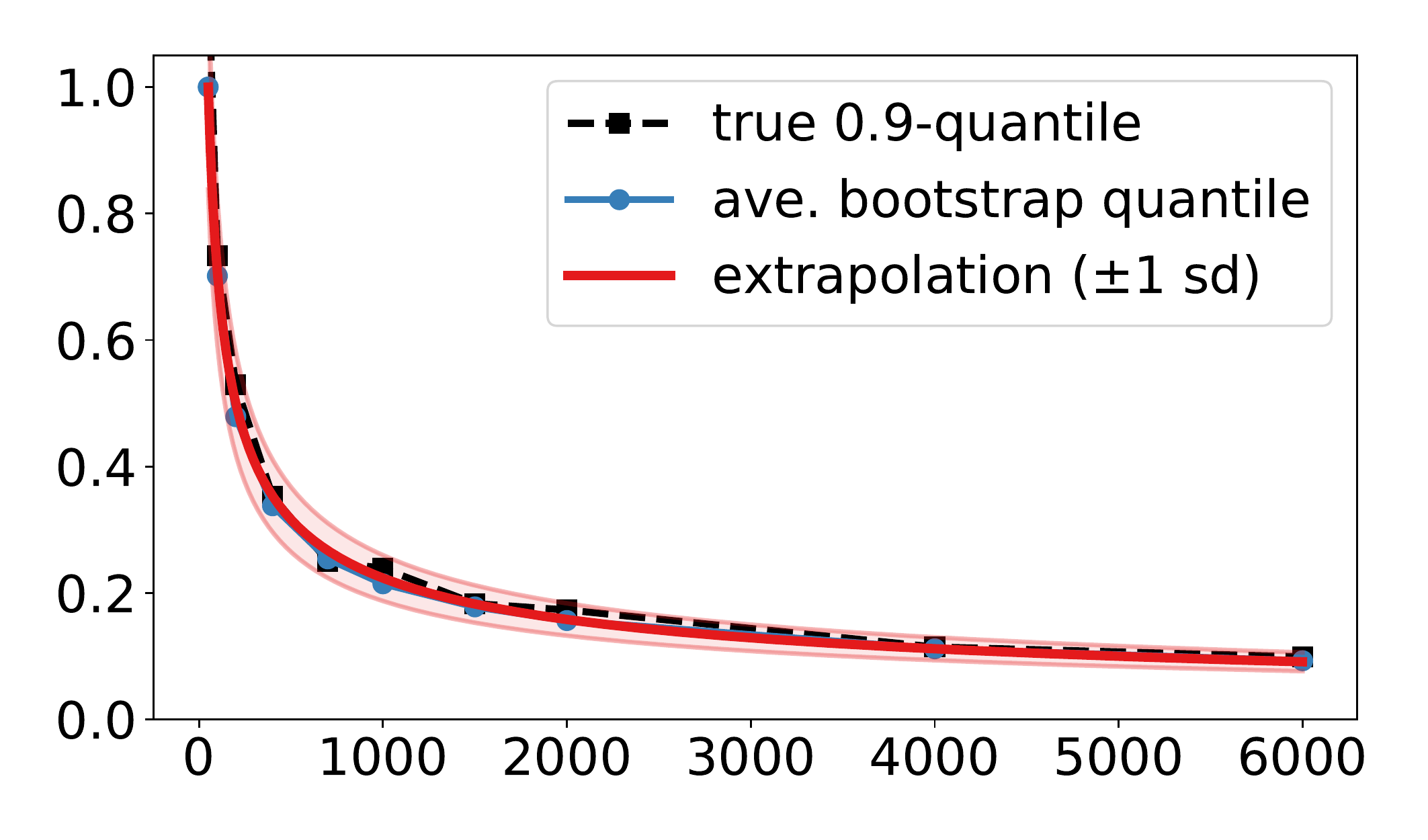} 
			\put(-6,24){\rotatebox{90}{\small $\ve_{0.9}$}} 
			\put(42,58){\color{black}{\small $\sigma=0.5$}} 
		\end{overpic}\hspace*{-0.2cm}
		~
		\begin{overpic}[width=0.31\textwidth]{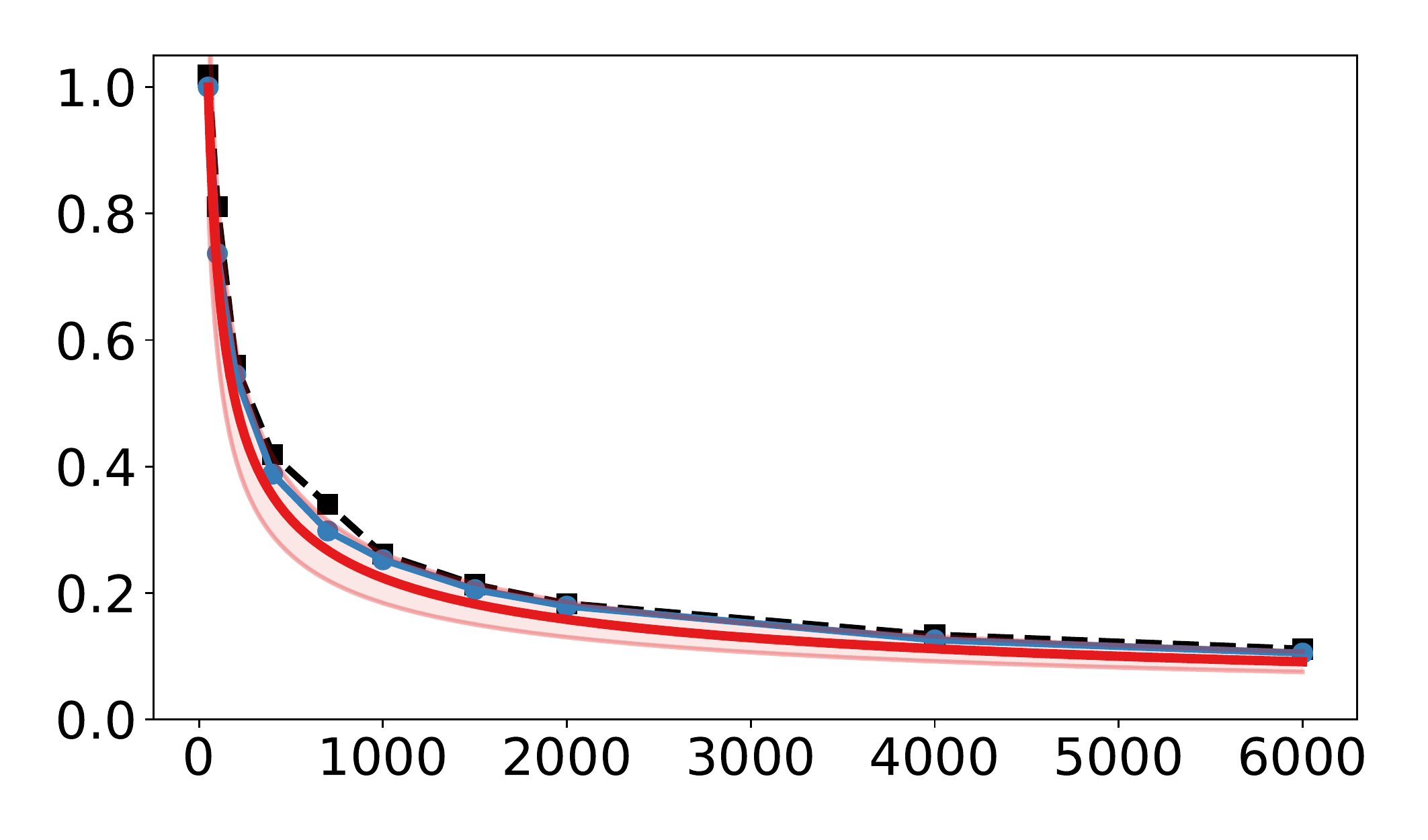} 
			\put(42,58){\color{black}{\small $\sigma=1.0$}} 
		\end{overpic}\hspace*{-0.2cm}
		~
		\begin{overpic}[width=0.31\textwidth]{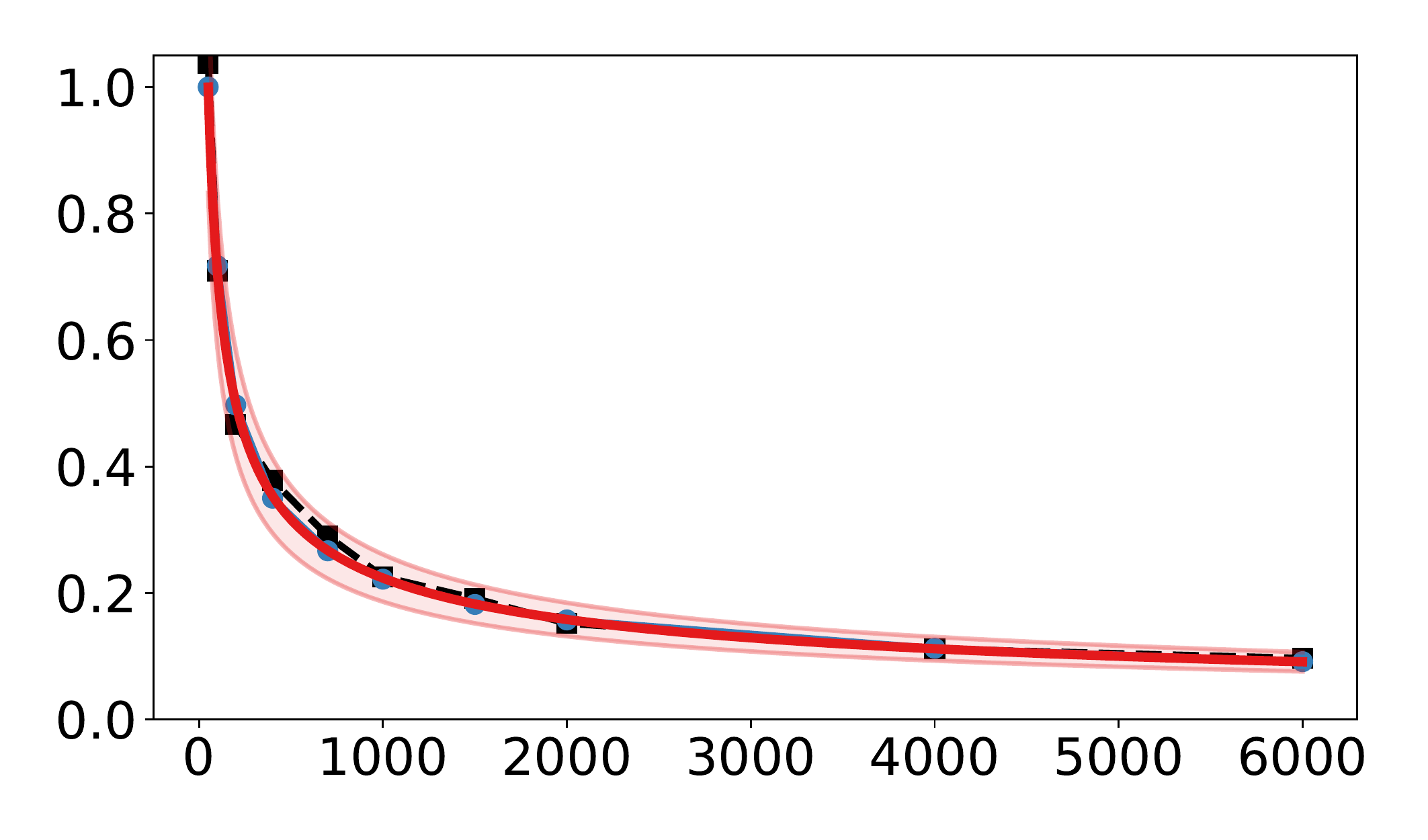} 
			\put(42,58){\color{black}{\small $\sigma=4.0$}} 
			\put(100,11){\rotatebox{90}{\small (operator norm)}}			
		\end{overpic}
	\end{subfigure}

	\begin{subfigure}{1\textwidth}	
		\centering
		\DeclareGraphicsExtensions{.pdf}
		\begin{overpic}[width=0.31\textwidth]{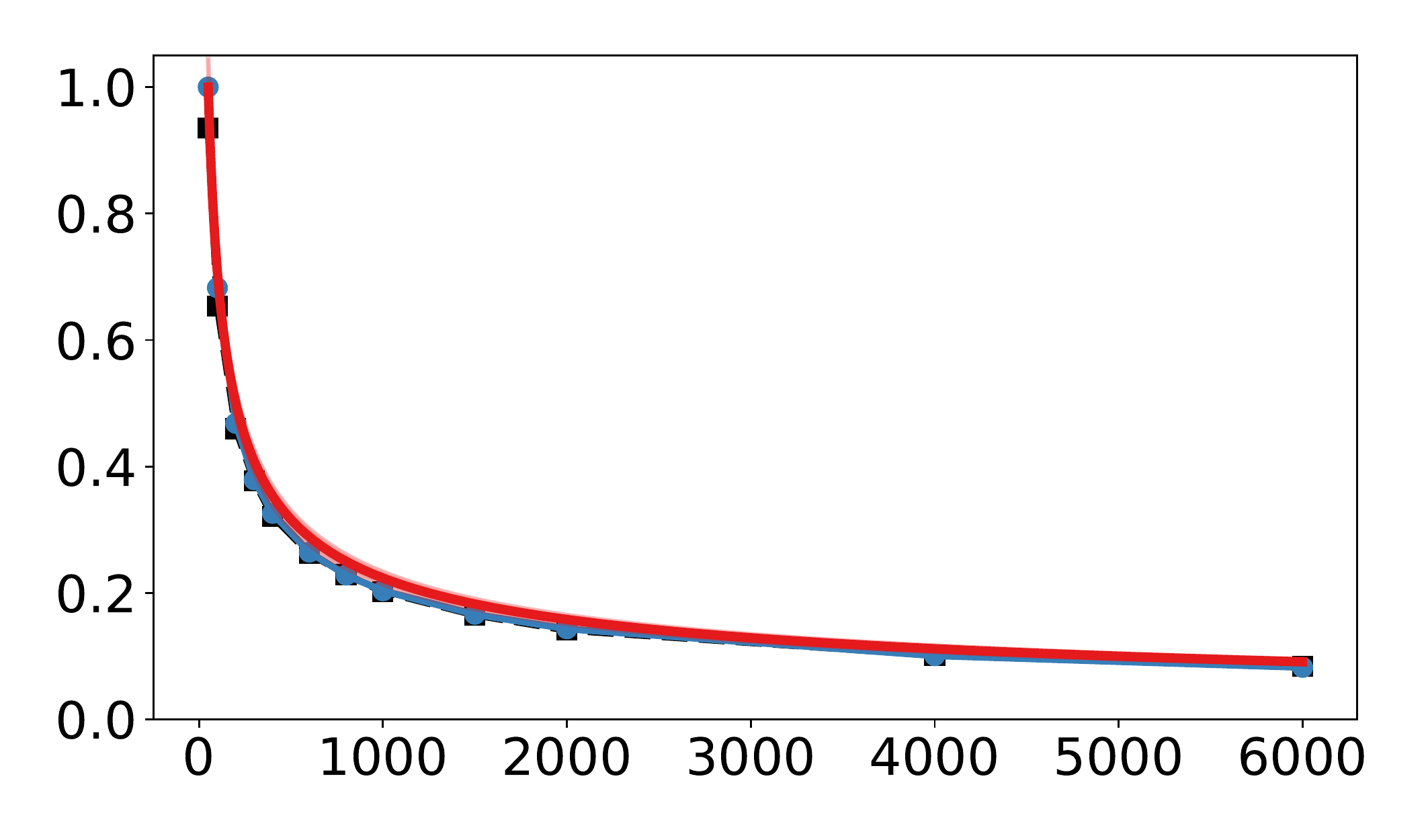} 
			\put(50,-2){\color{black}{\footnotesize $s$}}   
			\put(-6,24){\rotatebox{90}{\small $\ve_{0.9}$}}
		\end{overpic}\hspace*{-0.2cm}
		~
		\begin{overpic}[width=0.31\textwidth]{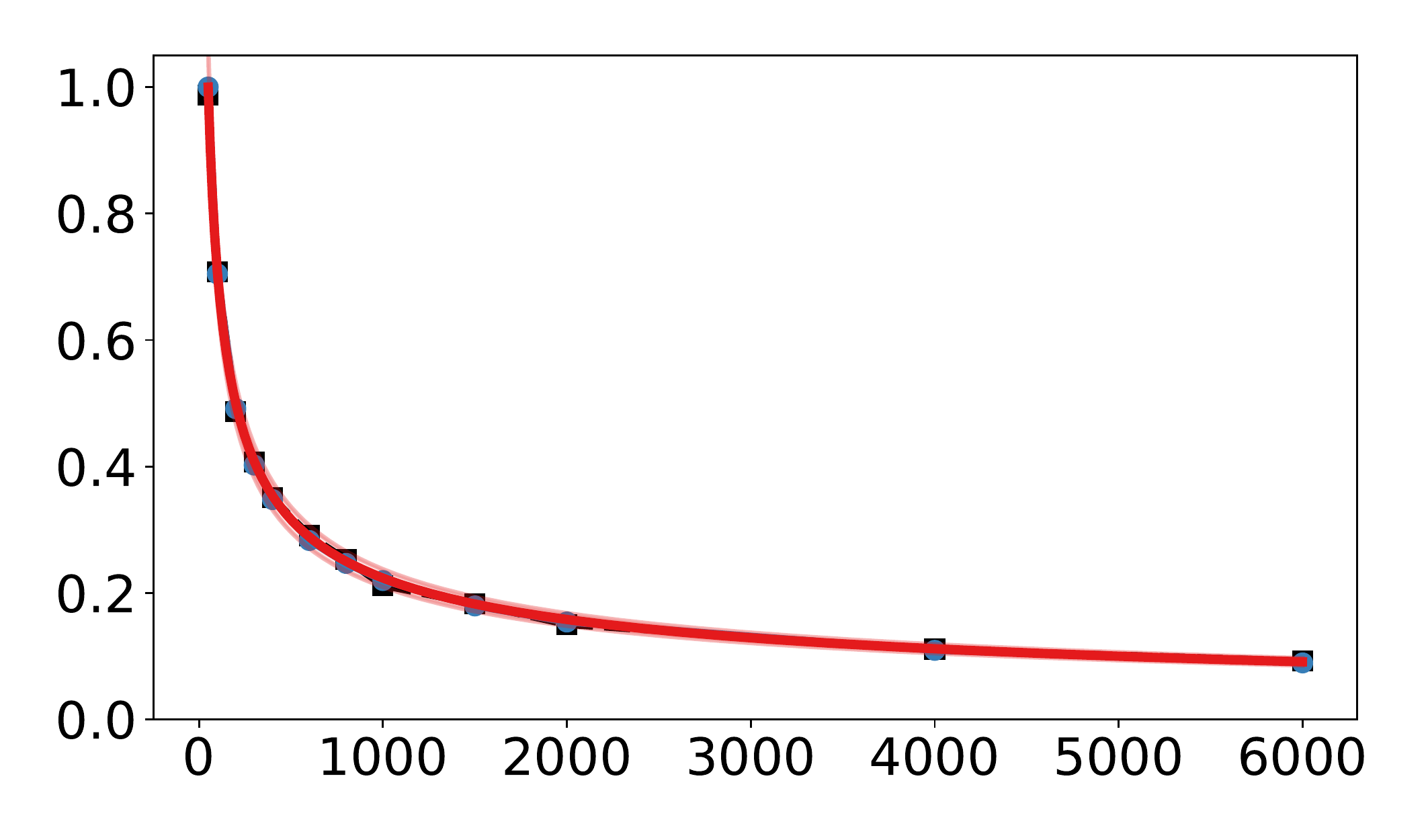} 
			\put(50,-2){\color{black}{\footnotesize $s$}}   
		\end{overpic}\hspace*{-0.2cm}
		~
		\begin{overpic}[width=0.31\textwidth]{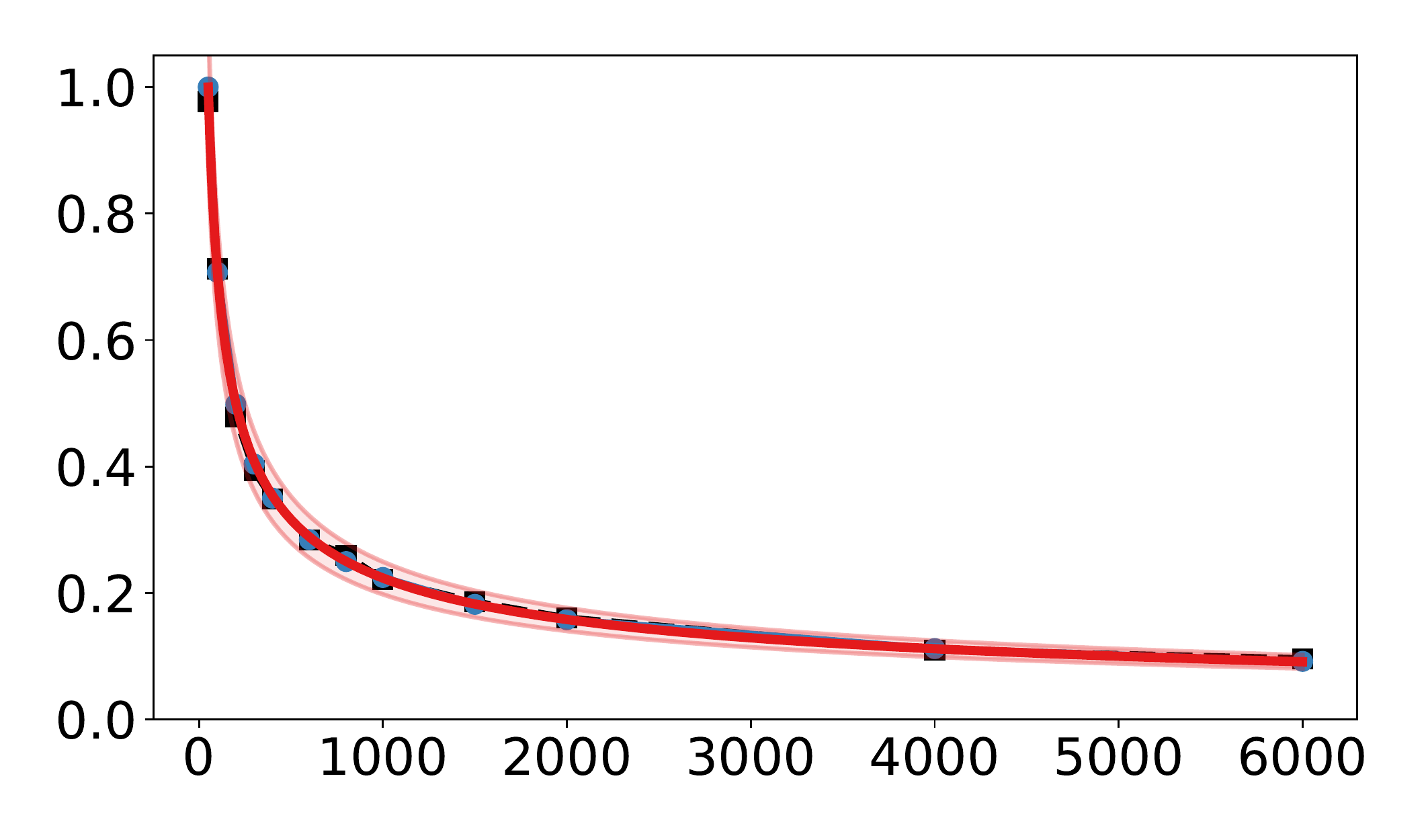} 
			\put(50,-2){\color{black}{\footnotesize $s$}}   
			\put(100,18){\rotatebox{90}{(\small $\ell_{\infty}$ norm)}}			
		\end{overpic}
	\end{subfigure}
%
	%
	\caption{(Estimation of $\ve_{0.9}$ for $\|\tK-\K\|_{\textup{op}}$ and $\|\tK-\K\|_{\infty}$). All plots are based on the Swiss roll dataset. The rows correspond to choices of matrix norm, and the columns correspond to choices of the kernel bandwidth $\sigma$. }

\label{fig:results_swiss}
\end{figure*}

Figure~\ref{fig:results_swiss} displays the performance of $\tilde{\ve}_{0.9}$ and $\tilde{\ve}_{0.9}^{\textsc{ ext}}$ in the task of estimating $\ve_{0.9}$. The top and bottom rows of Figure~\ref{fig:results_swiss} correspond respectively to the cases when matrix approximation error is measured through the operator norm $\|\tK-\K\|_{\textup{op}}$  and the $\ell_{\infty}$-norm $\|\tK-\K\|_{\infty}$. All the plots within Figure~\ref{fig:results_swiss} show that the estimates enjoy the same high degree of accuracy that was observed for the other datasets considered in Section 5.1 of the main text.
%

\nocite{}

\clearpage

\end{document}